\documentclass[10pt]{article} 
\usepackage[accepted]{tmlr}

\usepackage{amsmath,amsfonts,bm}

\def\eqref#1{equation~\ref{#1}}

\def\1{\bm{1}}

\DeclareMathAlphabet{\mathsfit}{\encodingdefault}{\sfdefault}{m}{sl}
\SetMathAlphabet{\mathsfit}{bold}{\encodingdefault}{\sfdefault}{bx}{n}

\usepackage{hyperref}
\usepackage{url}

\usepackage{microtype}
\usepackage{graphicx}
\usepackage{multirow}
\usepackage{subcaption}
\usepackage{booktabs} 

\usepackage{hyperref}

\usepackage[utf8]{inputenc} 
\usepackage[T1]{fontenc}    
\usepackage{hyperref}       
\usepackage{url}            
\usepackage{booktabs}       
\usepackage{amsfonts}       
\usepackage{nicefrac}       
\usepackage{microtype}      
\usepackage{xcolor}         

\usepackage{amsmath}
\usepackage{amssymb}

\usepackage{graphicx}
\usepackage{natbib}

\usepackage{amsthm}
\theoremstyle{plain}
\newtheorem{theorem}{Theorem}[section]
\newtheorem{proposition}[theorem]{Proposition}
\newtheorem{lemma}[theorem]{Lemma}
\newtheorem{corollary}[theorem]{Corollary}
\theoremstyle{definition}

\theoremstyle{remark}

\newcommand{\hsic}{\operatorname{HSIC}}
\newcommand{\cka}{\operatorname{CKA}}
\newcommand{\kcka}{\operatorname{kCKA}}
\newcommand{\mka}{\operatorname{MKA}}
\newcommand{\tr}{\operatorname{trace}}
\newcommand{\rbf}{\operatorname{RBF}}
\newcommand{\lin}{\operatorname{LIN}}
\newcommand{\knn}{\operatorname{kNN}}
\newcommand{\sgn}{\operatorname{sgn}}
\newcommand{\rtd}{\operatorname{RTD}}
\newcommand{\srtd}{\operatorname{sRTD}}
\newcommand{\imd}{\operatorname{IMD}}
\newcommand{\simd}{\operatorname{sIMD}}

\usepackage{amsmath}
\usepackage{amssymb}
\usepackage{mathtools}
\usepackage{amsthm}
\usepackage{bbm}
\usepackage{algorithm}
\usepackage{algpseudocode}

\title{Manifold Approximation leads to Robust Kernel Alignment}

\author{\name Mohammad Tariqul Islam \email mhdtariq@mit.edu \\
      \addr Nano-Cybernetic Biotrek, Media Lab, MIT
      \AND
      \name Du Liu \email liudu@mit.edu \\
      \addr Nano-Cybernetic Biotrek, Media Lab, MIT
      \AND
      \name Deblina Sarkar \email deblina@mit.edu\\
      \addr Nano-Cybernetic Biotrek, Media Lab, MIT}

\def\month{10}  
\def\year{2026} 
\def\openreview{\url{https://openreview.net/forum?id=AP7VqoJcLp}} 

\begin{document}

\maketitle

\begin{abstract}
Centered kernel alignment (CKA) is a popular metric for comparing representations, determining equivalence of networks, and conducting neuroscience research. However, CKA does not account for the underlying manifold and relies on numerous heuristics that cause it to behave differently at different scales of data.  In this work, we propose Manifold-approximated Kernel Alignment (MKA), which incorporates manifold geometry (i.e., local neighborhood) into the alignment task. We derive a theoretical framework for MKA. Our construction removes representation-dependent degree weighting, yields a closed-form alignment score, and gives each row a probabilistic interpretation as a local Markov transition distribution. We further establish kernel- and score-level stability with respect to the neighborhood size. We perform empirical evaluations on synthetic datasets and real-world examples to characterize and compare MKA to its contemporaries. In controlled synthetic experiments, MKA is consistently less sensitive to hyperparameters and captures the selected topology-preserving relationships. On the large-scale benchmark, MKA achieves its strongest aggregate performance in vision, is competitive on graphs, and is weaker than CKA in several NLP evaluations. Our findings suggest that manifold-aware kernel alignment provides a robust foundation for measuring representations and is potentially useful when local neighborhood geometry is important, with possible applications in representation learning.
\end{abstract}

\section{Introduction}

Centered Kernel Alignment (CKA)~\citep{cortes2010twostage,kornblith2019similarity} is a statistical method used to compare the similarity between representations of data, often in the form of feature maps or embeddings. 
It works by aligning kernels, which capture pairwise relationships within datasets, and measuring their agreement. CKA is widely used in studies to compare layers of neural networks, analyze representational similarity, and study how models process information~\citep{ramasesh2020anatomy,nguyen2022origins,ciernik2024training}. 
Its ability to handle datasets of varying sizes and dimensions makes it a powerful tool for understanding complex models and evaluating their performance. However, very few studies have characterized CKA under known representations/topologies. Moreover, the reliability of the CKA measure has been under scrutiny numerous times~\citep{davarireliability,murphy2024correcting}.

Empirically, CKA can be disproportionately influenced by dominant data points~\citep{nguyen2022origins}, outlier samples, and transformations that preserve linear separability or model function~\citep{davarireliability}. The commonly used biased CKA estimator also strongly depends on the feature-to-sample ratio and can produce inflated similarity in low-sample, high-dimensional regimes~\citep{murphy2024correcting}. Theoretically, \citet{davarireliability} showed that even a single outlier can substantially alter the CKA score. Moreover, standard CKA aggregates pairwise relationships globally and does not explicitly encode local neighborhood structure. Restricting kernel comparison to $k$-nearest-neighbor graphs provides a natural way to emphasize local geometry and reduce the influence of distant or anomalous samples. However, this localization introduces dependence on the neighborhood size $k$, and localized CKA variants can exhibit markedly different behavior as $k$ changes.

Motivated by these limitations, particularly the absence of explicit locality and sensitivity to neighborhood scale, we propose Manifold-approximated Kernel Alignment (MKA). 
Manifold approximation is a way to understand and simplify complex data. In many real-world problems, data with many dimensions - like x-rays, medical records, and neuroimaging data - actually lie on a much smaller, curved structure called a ``manifold'' within the high-dimensional space. Known as the ``manifold hypothesis'', this concept is integral to modern statistics and learning algorithms~\citep{fefferman2016testing}. 
Manifold approximation uncovers and represents this underlying structure of the high-dimensional data by exploiting the local relationships among the data points. Manifold approximation is popular in the field of non-linear dimensionality reduction. Methods such as t-distributed Stochastic Neighbor Embedding (t-SNE)~\citep{van2008visualizing} and Uniform Manifold Approximation and Projection (UMAP)~\citep{mcinnes2018umap} are widely used to visualize high-dimensional data in many fields, including representation analysis and single-cell transcriptomics studies.

We use manifold approximation to define a non-linear and non-Mercer kernel. Unlike manifold-learning pipelines that symmetrize local affinities before producing an embedding, MKA intentionally retains directed neighborhood relationships. We combine this directed kernel with a modified alignment operation and impose a constant-row-sum constraint. These choices lead to a closed-form alignment score and allow us to establish element-wise, row-wise, and score-level stability with respect to changes in the neighborhood size. They also yield a probabilistic interpretation in terms of local Markov transition distributions without representation-dependent degree weighting. More specifically, under CKA’s Markov-kernel interpretation, identical row-normalized neighborhood distributions do not necessarily receive maximal alignment when their representation-dependent degree weights differ. MKA removes these degree weights by enforcing equal row mass, so that maximal alignment corresponds directly to agreement of the directed local transition distributions.

We characterize MKA using controlled synthetic datasets involving topology-preserving transformations, perturbations, variations in sample size and dimensionality, and changes in neighborhood size. In these controlled settings, MKA more consistently captures the selected local structural relationships and is less sensitive to neighborhood-scale choices than CKA, localized CKA variants, and several contemporary methods.
We additionally evaluate MKA on the large-scale Representation Similarity (ReSi) benchmark across vision, natural-language, and graph representations. The results are domain-dependent: MKA achieves its strongest aggregate performance in vision, is competitive with other locality-based methods on graphs, and is outperformed by CKA in several NLP evaluations. These findings position MKA as a complementary alignment measure, particularly useful when local neighborhood geometry and robustness to neighborhood-scale selection are important. Initial version of the paper was presented in~\cite{islam2025kernel}. Code is available at \href{https://github.com/tariqul-islam/mka}{https://github.com/tariqul-islam/mka}.

We organize the rest of the paper as follows. Section~\ref{sec:related_works} reviews relevant work on representation similarity and kernel alignment, and then discusses work using manifold approximation for dimensionality reduction and topological data analysis. After that, we provide a brief discussion on CKA and its kernel choices in Section~\ref{sec:cka_intro}. Section~\ref{sec:mka_intro} provides the formal description of MKA and establishes its boundedness, robustness, and implementation. We also compare it to CKA in terms of Markov Kernels in the same section. Section~\ref{sec:experiments} provides extensive experiments on carefully chosen synthetic datasets, a large-scale real-world benchmark (ReSi), and comparison to neural network representations to evaluate MKA. Finally, Section~\ref{sec:conclusions} provides concluding remarks.


\section{Related Works}\label{sec:related_works}

The recent interest in alignment metrics stems from the desire to understand how neural networks work and how the intermediate layers of neural networks are related. To compare learned features, we need metrics that measure alignment between two representations. Earlier studies assessed representational similarity with correlation- and mutual-information–based measures~\citep{li2015convergent} and with linear-classifier probes~\citep{alain2016understanding}. Subsequent progress came from \cite{raghu2017svcca}, who modeled the problem as one of dimensionality reduction and used singular value decomposition (SVD) to remove noise from the representations, followed by canonical correlation for alignment, namely SVCCA. Later, \cite{morcos2018insights} proposed PWCCA, which extends SVCCA by weighting the canonical directions according to their contribution to the original representations, making the similarity measure more robust to noisy or unimportant dimensions. This dimensionality reduction approach has also been followed by a few other studies~\citep{sussillo2013opening,maheswaranathan2019universality}. Other approaches include revisiting classifier probes~\citep{graziani2019interpreting,davari2022probing}, exploring multiple approaches together~\citep{ding2021grounding}, Procrustes analysis~\citep{williams2021generalized}, graphs~\citep{chen2021revisit}, and exploring effect of transformations~\citep{lenc2015understanding}.

However,~\citet{kornblith2019similarity}’s exploration of representations through kernel methods has sparked renewed attention and discoveries in this area. Known as centered kernel alignment (CKA)~\citep{cortes2010twostage,kornblith2019similarity}, this approach compares two different kernel matrices obtained from the representations. The initial studies~\citep{kornblith2019similarity,nguyen2020wide,raghu2021vision} explored the feature similarity of nearby layers (the famous block structure).  Another concurrent observation is that the kernel structure is relatively robust when low-variance components are removed~\citep{ding2021grounding}. Later, \citet{nguyen2022origins} discovered that the block structure is primarily due to a few dominant data points. \citet{davarireliability} formalized these observations theoretically. For a comparison of CKA to its contemporaries (representation similarity analysis and CCA), see \cite{williams2024equivalence}.

Another avenue is to explore the nearest neighbor structure, which, in our opinion, is a natural extension of the CKA philosophy. \citet{huh2024platonic} proposed a mutual nearest neighbor-based extension of CKA, which restricts kernel comparison to neighborhoods supported by both representations. \citet{tsitsulin2019shape} proposed Intrinsic Multi-scale Distance (IMD), which uses the heat kernel to estimate the manifold. Recently, topological data analysis has been applied to propose Representational Topology Divergence (RTD)~\citep{barannikov2021representation,tulchinskiirtd}.

Nearest-neighbor constructions are also central to nonlinear manifold approximation and dimensionality reduction. Methods such as stochastic neighbor embedding~\citep{hinton2002stochastic,van2008visualizing}, Uniform Manifold Approximation and Projection (UMAP)~\citep{mcinnes2018umap}, related variants~\citep{wang2021understanding,damrich2022t,islam2022manifold}, and legacy methods~\citep{belkin2002laplacian,coifman2006diffusion} rely on efficient sampling of the manifold followed by optimization of a low-dimensional embedding. In particular, the use of k-nearest neighbor graphs and parameter-tuned local neighborhoods has proven to be an effective tool for this class of methods. While k-nearest neighbors show usefulness in some recently proposed alignment metrics~\citep{tsitsulin2019shape,huh2024platonic} and topology~\citep{damrich2024persistent}, the kernels arising from manifold approximation have been less explored as the basis of a kernel-alignment measure.

These approaches address complementary aspects of representation comparison. Standard CKA provides a global kernel-based comparison. Nearest-neighbor methods introduce locality, but existing localized CKA variants do not constrain the row masses of their kernels (as scale varies) or provide guarantees about how the resulting score changes with the neighborhood size. IMD and RTD instead compare manifold- or topology-based summaries, rather than directly aligning the conditional neighborhood distributions associated with individual samples. Finally, the local affinity construction used by MKA is closely related to that of UMAP; our contribution is not the affinity rule alone, but its use in a representation-alignment measure that preserves directionality and equal row mass. Together with the modified alignment operation, these properties yield a degree-unweighted Markov interpretation, a closed-form score, maximal alignment for identical directed transition distributions, and stability guarantees with respect to the neighborhood size.

\section{Centered Kernel Alignment (CKA)}\label{sec:cka_intro}
Let $X\in\mathbb{R}^{n\times d_1}$ and $Y\in\mathbb{R}^{n\times d_2}$ be feature sets from $n$ samples each with $d_1$ and $d_2$ features, respectively. The corresponding symmetric kernel matrices are $K$ and $L$ with $K_{ij}=k(x_i,x_j)$ and $L_{ij}=l(y_i,y_j)$, respectively. The CKA measure between the two feature sets is given by
\begin{align}
    \cka(K,L) = \frac{\hsic(K,L)}{\sqrt{\hsic(K,K)\hsic(L,L)}} = \frac{\tr(KHLH)}{\sqrt{\tr(KHKH)\tr(LHLH)}},
\end{align}
where $\hsic(\cdot,\cdot)$ is the Hilbert-Schmidt Independence Criterion given by
$\hsic(K,L) = \frac{1}{(n-1)^2}\tr(KHLH)$. Here, $H=I-\frac{1}{n}\mathbf{1}\mathbf{1}^T$ is a centering matrix that mitigates bias in the kernel. There are other debiasing techniques~\citep{song2007supervised,sucholutsky2023getting}, however, we will consider the simplest and most widely used one in practice. $\hsic$ computes the similarity between the two kernel matrices of the same size, while the $\cka$ measure normalizes this similarity within $[0,1]$.

Various options exist for the kernel. The common ones include the linear kernel (LIN) given by $k(x_i,x_j)=x_i^Tx_j$ and the radial basis function (RBF) kernel given by $k(x_i,x_j)=\exp(-||x_i-x_j||_2^2/(2\sigma^2))$, where $\sigma$ is the bandwidth of the Gaussian. The following theorem establishes an equivalence relation between CKA with linear and RBF kernel:
\begin{theorem}[\cite{alvarez2022gaussian}]\label{theorem:alvarez}
    $\cka(K_{\rbf},L) = \cka(K_{\lin},L)+O(1/\sigma^2)$ as $\sigma\to\infty$. Here, $K_{\rbf}$ is the RBF kernel matrix with bandwidth $\sigma$, $K_{\lin}$ is the linear kernel matrix, and $L$ is any positive definite symmetric kernel matrix.
\end{theorem}
Softly, it states that at higher values of $\sigma$, CKA with linear and RBF kernels behave equivalently. Various studies have reported this in empirical settings (e.g., in~\cite{kornblith2019similarity} and Fig. 4(a) of \cite{davarireliability}). Thus, most researchers use the linear kernel, effectively capturing linear relationships alone. And by Theorem~\ref{theorem:alvarez}, even results with an RBF kernel (without properly tuning the bandwidth, $\sigma$) potentially suffer from the same pitfalls of the linear one. In the rest of the paper, unless explicitly mentioned, we use $\cka$ only with RBF kernel.

\section{Manifold-approximated Kernel Alignment (MKA)}\label{sec:mka_intro}
Manifold approximation is a method for defining a graph that quantifies the pairwise relations within the data. A manifold-approximated kernel is often sparse and typically obtained by the k-nearest neighbor ($\knn$) algorithm. These kernels are generally not positive semidefinite; rather, they are of non-mercer~\citep{ong2004learning} type. Moreover, we will use a kernel function that is non-symmetric (i.e., $k(x_i,x_j)\neq k(x_j,x_i)$). We impose additional conditions for computational advantage, explainability and robustness. Overall, for MKA the design properties are:
\begin{enumerate}
    \item[A1] \textbf{(directed $\knn$ locality; non-symmetric)} $K_{ii}=1$ and for $i\neq j$, $K_{ij}=0$ unless $x_j\in\knn(x_i)$.
    \item[A2] \textbf{(constant row-sum)} $K\mathbf{1}=D\mathbf{1}$, where $D$ is a constant.
    \item[A3] \textbf{(stability)} There exists a parameter sequence $(\theta_k)$ and $(\varepsilon_k)$ such that for all $k$,
\begin{align}
    \max_{i\in[n]}\|K(\theta_{k+1})_{i,:}-K(\theta_k)_{i,:}\|_{1} \le \varepsilon_k.
\end{align}
Here, $\theta_k$ values represent the hyperparameters for each value $k$.
\end{enumerate}
The first property is trivial, while the second one provides a different explainability from that of CKA (in Section~\ref{sec:comparison_to_cka}), and the third ensures that the alignment metric is stable under hyperparameter variation.
With this kernel, we define Manifold-approximated Kernel Alignment (MKA) as:
\begin{align}
    \mka(K,L) = \frac{\langle KH, LH\rangle}{\sqrt{\langle KH, KH\rangle\langle LH, LH\rangle}}. \label{eq:mka_definition}
\end{align}
Despite using non-symmetric kernels, the measure $\mka$ is symmetric ($\mka(K,L)=\mka(L,K)$). 
However, unlike CKA, which performs both row- and column-wise centering, we deliberately use only row-wise centering.
Because each row of the directed kernel represents a source-conditioned local transition distribution, multiplication by $H$ on the right subtracts the uniform distribution from each row while preserving this conditional interpretation (additional details are in Section~\ref{sec:comparison_to_cka}).
In terms of the $\tr$ operator, MKA is given by
\begin{align}
    \mka(K,L) = \frac{\tr(K^TLH)}{\sqrt{\tr(K^TKH)\tr(L^TLH)}}.
\end{align}
Furthermore, exploiting the properties of the kernel matrix, we can simplify and characterize $\mka$ by
\begin{theorem}[Closed-form expression for MKA]\label{thm:mka_simple}
    If the kernels $K$ and $L$ both follow constant row sum property (A2) with $K\mathbf{1} = L\mathbf{1} = D\mathbf{1}$, then $\mka$ reduces to
    \begin{align}
        \mka(K,L) = \frac{\langle K,L\rangle-D^2}{\sqrt{ (\langle K, K \rangle-D^2) (\langle L, L \rangle-D^2) }}. \label{eq:mka_theorem_eq}
    \end{align}
\end{theorem}
\begin{corollary}[Boundedness of MKA]\label{thm:mka_range}
    If the kernels $K$ and $L$ follow properties (A1) and (A2) with $K\mathbf{1} = L\mathbf{1} = D\mathbf{1}$ and $D < \sqrt{n}$, then $0\leq\mka(K,L)\leq1$.
\end{corollary}

\begin{corollary}[Stability of MKA]\label{cor:metric_stability}
If $\exists \eta\in(0,1)$ such that $\max{\{D_{k+1}^2,D_k^2\}}\leq (1-\eta)n$, then
    \begin{align}
 \big|MKA(K(\theta_{k+1}),L)-MKA(K(\theta_k),L)\big| \leq \frac{2\epsilon_k}{\sqrt{\eta}}. 
\end{align}

\end{corollary}

Theorem~\ref{thm:mka_simple} enables fast computation of $\mka$, making it more scalable (especially when combined with approximate nearest neighbor search algorithms). Corollary~\ref{thm:mka_range} shows the boundedness of the MKA values, whereas Corollary~\ref{cor:metric_stability} bounds the change in MKA under adjacent changes in the scale parameter. Few works~\citep{chen2021revisit,huh2024platonic} have considered sparsifying the kernel matrix of CKA by taking the top-k values in rows/columns. However, these works do not consider constraining the rows/columns of the kernel matrix or exploring the robustness of the algorithms.

\subsection{Interpretation of MKA}\label{sec:comparison_to_cka}
To decipher $\mka$ and establish a conceptual relation to $\cka$, we provide a Markov kernel interpretation of the algorithms in the following theorems:
\begin{theorem} [Markov interpretation of nonlinear $\cka$]\label{thm:cka_markov}
$\cka$ with RBF Kernel is the cosine similarity (in the Frobenius inner-product space) between left centered degree-weighted grand-mean centered Markov transition matrices, i.e.,
\begin{align}
\cka(K,L) = \frac{\langle HS(P-U), HT(Q-U)\rangle_F} {||HS(P-U)||_F||HT(Q-U)||_F},
\end{align}
where $S$ and $T$ are diagonal degree matrices with $s_{ii} = \sum_j K_{ij}$ and $t_{ii} = \sum_j L_{ij}$, $P=S^{-1}K$ and $Q=T^{-1}L$ are row-stochastic Markov kernels of the feature sets, and $U=\frac{1}{n}\mathbf{1}\mathbf{1}^T$ is the uniform distribution.
\end{theorem}

\begin{theorem}[Markov interpretation of $\mka$]\label{thm:mka_markov}
$\mka$ is the cosine similarity between grand-mean centered Markov transition matrices, i.e.,
\begin{align}
\mka(K,L) = \frac{\langle P-U, Q-U\rangle_F} {||P-U||_F||Q-U||_F}.
\end{align}
\end{theorem}

Theorem~\ref{thm:mka_markov} is a consequence of the joint design of the kernel and the alignment operation. It removes the degree weighting from the matrices $S$ and $T$. Since $P$ and $Q$ are row-stochastic Markov kernels, each of the rows $P_{i}$ and $Q_{i}$ has a natural probabilistic interpretation as a neighborhood distribution on the manifold; both algorithms measure the distance from the trivial uniform baseline, $U$.
However, $\cka$ is dependent on the row sum of the kernel matrices, which in turn depends on the pairwise distances and the value of $\sigma$. $\mka$, on the other hand, is independent of the row sum (by design) and therefore it compares how the two representations deviate from the uniform distribution in their local transition structure, without requiring the kernels to be symmetric or positive semidefinite. In particular, if the two representations induce identical neighborhood distributions, $P=Q\neq U$, then $\mka(K,L)=1$. Thus, the maximum alignment score corresponds exactly to perfect agreement of the directed neighborhood structure on the manifold.

Using this simple probabilistic interpretation from the Markov transition kernel, we now address the residual bias due to row-only centering. To do this, first, we define a double-centered alignment: $MKA_{DC}=<H(P-U), H(Q-U)>/(||H(P-U)||_F||H(Q-U)||_F)$ that removes this residual bias due to row-only centering. We can also define the centered column sums, $h_p = (P-U)^\top\mathbf{1} = P^\top\mathbf{1}-\mathbf{1}$ and $h_q = (Q-U)^\top\mathbf{1} = Q^\top\mathbf{1}-\mathbf{1}$, that capture effects  arising from the lack of column centering and symmetry (i.e., the hubness~\citep{radovanovic2010hubs}). Additionally, let $\phi_p = \|h_p\|_2^2 / (n\|P-U\|_F^2),$ and $\phi_q=\|h_q\|_2^2 / (n\|Q-U\|_F^2)$, quantities that measure the fraction of the total squared deviation from uniformity attributable to the centered column-sum, or hubness. We have the following:
\begin{proposition}[Effect of row-only centering]
\label{prop:hubness_effect}
$\mka$ and $\mka_{DC}$ are related by,
\begin{align}
\mka = \sqrt{(1-\phi_p)(1-\phi_q)} \mka_{DC} + \sqrt{\phi_p\phi_q} s_{\mathrm{hub}},
\label{eq:mka_mka_dc_relation}
\end{align}
where
$ s_{\mathrm{hub}} = \frac{h_p^{\top}h_q} {\|h_p\|_2\|h_q\|_2}$
measures agreement between the two hubness patterns.
Consequently,
\begin{align}
\left|\mka - \mka_{DC} \right| \le 1-\sqrt{(1-\phi_p)(1-\phi_q)} + \sqrt{\phi_p\phi_q}.    
\label{eq:mka_mka_dc_inequality}
\end{align}

In particular, if $\phi_p\le\epsilon$ and $\phi_q\le\epsilon$ for some $\epsilon\in[0,1]$, then 
\begin{align}
    \left|\mka - \mka_{DC} \right|\le 2\epsilon.
    \label{eq:mka_mka_dc_bound}
\end{align}

\end{proposition}

Proposition~\ref{prop:hubness_effect} establishes a formal relationship between MKA and its double-centered counterpart. Overall, the quantity $\sqrt{(1-\phi_p)(1-\phi_q)}$ denotes the fraction of the double-centered alignment contribution that is captured in MKA, and the quantity $\sqrt{\phi_p\phi_q} s_{\mathrm{hub}}$ gives the portion due to the column sums. As long as the energies $\phi_p$ and $\phi_q$ are small, the difference between the two is also small. We examine these quantities using numerical examples in Appendix~\ref{sec:row_centering_appendix}.

\subsection{Implementation}

We define a manifold-approximated kernel using a pairwise relationship by
\begin{align}
    K_{ij} &= \begin{cases}
            1, &\text{if~~~} i=j\\
            \exp{\left(-\frac{d(x_i,x_j)-\rho_i}{\sigma_i}\right)} & \text{if~~ } x_j\in \knn(x_i,k) \\
            0 & \text{otherwise}
        \end{cases}, \label{eq:UMAP_HIGH_DIM}
\end{align}
where $\knn(x_i,k)$ contains the $k$-nearest neighbors of $x_i$, $d(\cdot,\cdot)$ is a distance metric, $\rho_i = \min_{x_j\in \text{KNN}(x_i,k)} d(x_i,x_j)$ is the minimum distance from the nearest neighbor and $\sigma_i$ is the row-wise bandwidth parameter. $\sigma_i$ is analogous to the $\sigma$ parameter of the RBF kernel; however, $\sigma_i$ operates only on each row of the kernel matrix, while $\sigma$ works globally. This bandwidth is computed such that $\sum_j K_{ij}=1+\log_2(k)$. This constraint fixes the row sum of the kernel matrix to a constant, i.e., (A2), and makes the kernel less sensitive to lone outliers. Additionally, this imposes a rank order within the row. The $\knn$ imposes a stricter constraint on the number of points that are considered related compared to CKA, which allows for a softer, more global measure of similarity. 
The directed affinities in Eq.~(\ref{eq:UMAP_HIGH_DIM}) are adapted from UMAP’s local-connectivity construction. The distinction arises in how they are subsequently used. UMAP symmetrizes the directed affinities through a fuzzy union and uses the resulting graph to optimize an embedding. MKA does not perform either operation. Retaining the directed affinities is primarily a structural choice. Symmetrization would generally alter the row sums and merge the distinct conditional relations, $i\to j$ and $j \to i$, thereby breaking the equal-row-mass property (A2) required for the Markov interpretation, closed-form alignment expression, and stability analysis. Omitting symmetrization also avoids an additional graph-combination operation, although we do not treat this as the principal source of MKA’s computational advantage. The measured computational benefit instead arises from the closed-form score in Eq.~(\ref{eq:mka_theorem_eq}), as evaluated in Appendix~\ref{sec:complexity}.

With this specific manifold-approximated kernel, we can now discuss the stability of MKA in terms of $k$. Overall, $K$ is a graph on the data that depends on only one hyperparameter: $k$ (i.e., $\theta_k=\{k\}$). As $k$ increases, the bandwidth parameter $\sigma_i$ changes slowly, which in turn conforms to property (A3) and thus, MKA becomes a robust alignment measure. To establish this formally, we provide the following theorems:

\begin{theorem}[Kernel stability]
\label{thm:kernel_stability}
Let $K^{(k)}$ denote the kernel in Eq.~(\ref{eq:UMAP_HIGH_DIM}) constructed using $k$ nearest neighbors. Assume $d(x_i,x_{(2)})-\rho_i\ge \delta>0$, where $x_{(2)}$ is the second nearest neighbor of $x_i$ (ordered by increasing distance). Further assume $\sigma_{i,k},\sigma_{i,k+1}\in[\underline{\sigma},\overline{\sigma}]$ for some
$0<\underline{\sigma}\le \overline{\sigma}<\infty$,  where $\sigma_{i,k}$ and $\sigma_{i,k+1}$ are the bandwidths with $k$ and $k+1$ nearest neighbors, respectively.
Then for any $j$,
\begin{align}
 \big|K^{(k+1)}_{ij}-K^{(k)}_{ij}\big| = \mathcal{O}\!\left(\frac{\log_2 k}{k}\right),
\qquad k\ge 3,   
\end{align}
where the implicit constant depends only on $\delta$, $\underline{\sigma}$, and $\overline{\sigma}$.
\end{theorem}

\begin{theorem}[Row-wise kernel stability]
\label{thm:row_stability}
    The kernel in Eq.~(\ref{eq:UMAP_HIGH_DIM}) follows property (A3) with $\varepsilon_k = \mathcal{O}\left(\frac{\log_2 k}{k}\right)$, i.e.,
    \begin{align}
    \max_{i\in[n]} \big\|K^{(k+1)}_{i:}-K^{(k)}_{i:}\big\|_1 = \mathcal{O}\left(\frac{\log_2 k}{k}\right),
\qquad k\geq 3,   
    \end{align}
\end{theorem}

\begin{corollary}[MKA implementation stability]\label{cor:mka_implement_Stability}
    If $\exists \eta \in(0,1)$ such that $D_k^2=(1+\log_2k)^2<D_{k+1}^2=(1+\log_2 (k+1))^2\leq(1-\eta)n$, then,
    \begin{align}
        \big|MKA(K^{(k+1)},L^{(k+1)})-MKA(K^{(k)},L^{(k)})\big| = \mathcal{O}\!\left(\frac{\log_2 k}{k}\right), \qquad k\geq 3. 
    \end{align}
\end{corollary}

Theorem~\ref{thm:kernel_stability} bounds the element-wise kernel change from $k$ to $k+1$, as their element-wise absolute difference is proportional to $ \log_2{(k)}/k$. As $k$ increases, the denominator dominates, making the kernel relatively unchanged. Theorem~\ref{thm:row_stability} shows that the implemented kernel follows property $(A3)$. Finally, Corollary~\ref{cor:mka_implement_Stability} shows that the MKA score is stable under an adjacent change from $k$ to $k+1$, under the assumption $(1+\log_2 (k+1))^2\leq(1-\eta)n$. In Appendix~\ref{sec:eta}, we explore the possible ranges of $\eta$ for this implementation.

\section{Experiments}\label{sec:experiments}

In this section, we empirically characterize MKA using various datasets and benchmarks. We compare MKA with several CKA variants with the RBF kernel: 1) $\cka (\sigma=M)$: $\sigma$ is set to the median, $M$, of the entries of the distance matrix, 2) $\cka (\sigma=\delta M)$: $\sigma$ is set to $\delta M$ for considering local relationships (we mostly use $\delta=0.2$ or $0.45$), and 3) k$\cka$: sparsifying the kernel matrix by considering $k$-nearest neighbors of each sample and setting $\sigma$ to be the median of the considered distances giving us a simple manifold approximation (Note that CKNNA~\citep{huh2024platonic} is a closely related method. For a discussion of the two, see Appendix~\ref{sec:cknna_kcka}). Computationally, kCKA is similar to CKA, but uses a non-Mercer kernel, and thus, we consider it a conceptually intermediate version of MKA (for a runtime evaluation of these methods, see Appendix~\ref{sec:complexity}).

Along with the CKA variants, we consider Representational Topology Divergence (RTD), Intrinsic Multi-scale Distance (IMD), and Support Vector Canonical Correlation Analysis (SVCCA) metrics. 
RTD and IMD provide metrics within $[0,\infty]$, with a lower value corresponding to stronger alignment. We scale these values within $[0,1]$ using the formulae $\srtd=\exp{(-\rtd/\gamma)}$ and $\simd=exp{(-\imd/\gamma)}$ for the respective methods and tune $\gamma$ for each experiment (for additional figures for RTD and IMD for the experiments, see Appendix~\ref{sec:rtdimdraw}). 
In the figures, we explicitly differentiate between IMD (RTD) and sIMD (sRTD), while in the text, we use them interchangeably.

\begin{figure*}[t]
    \centering
    \includegraphics[width=1.0\linewidth]{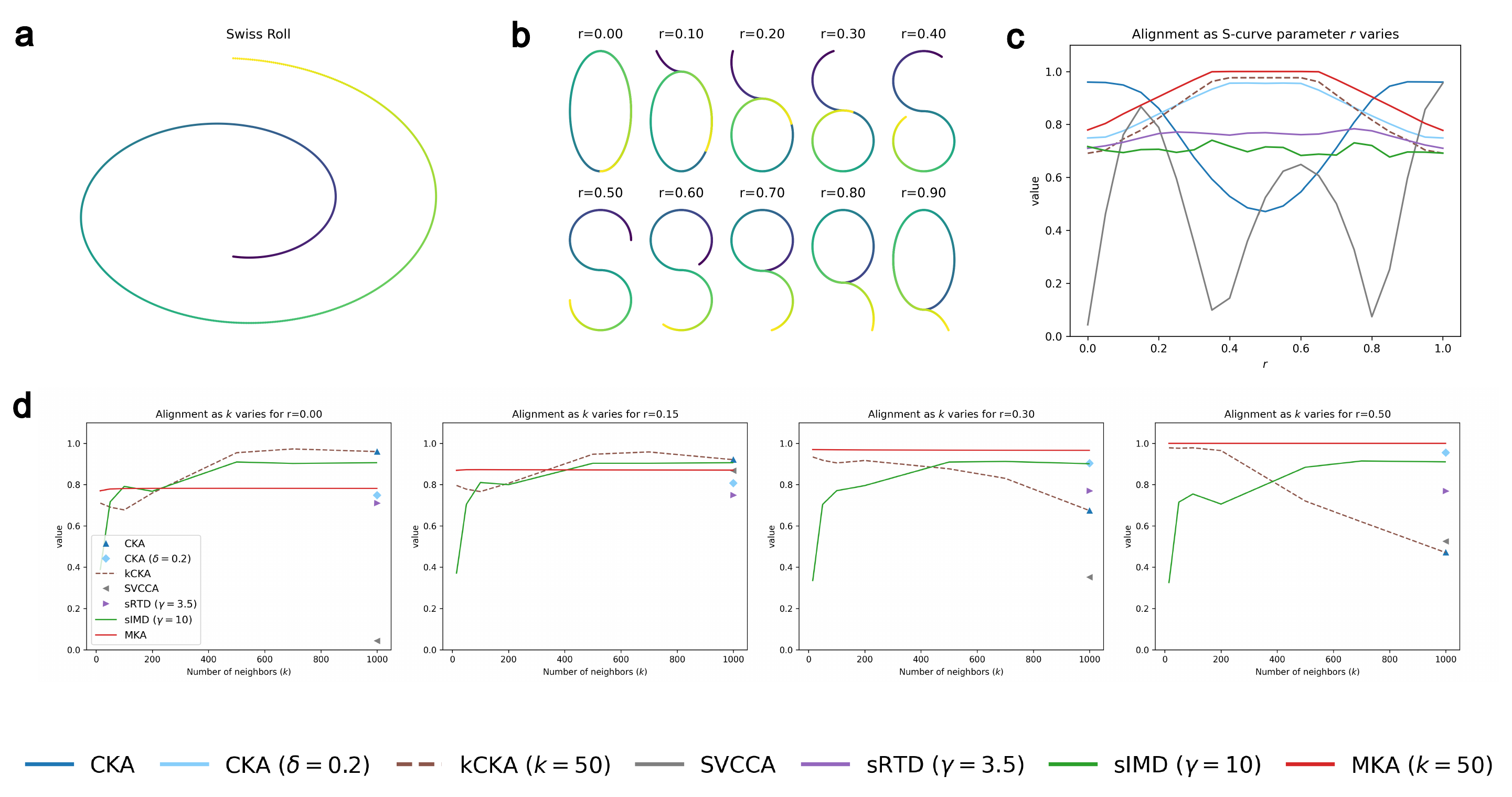}
    \caption{Equivalence of two different shapes with 1-D manifolds. (a) Swiss-roll. (b) S-curve by varying parameter $r$. (c) Alignment for the methods as S-curve parameter, $r$, varies. (d) Alignment for different methods as the number of nearest neighbors, $k$, varies. Note that $\cka$, RTD, and SVCCA do not have any notion of nearest neighbors; thus, we have plotted these values at the end of the x-axis.}
    \label{fig:swiss_s_fig}
\end{figure*}

\subsection{Equivalence of Shapes}

We start the experiments by comparing two classic shapes: Swiss-roll (Fig.~\ref{fig:swiss_s_fig}(a)) and S-curve (Fig.~\ref{fig:swiss_s_fig}(b), $r=0.5$). Although the Swiss roll and the S-curve look drastically different, they are topologically equivalent: both lie on a one-dimensional nonlinear manifold. Furthermore, the parameter $r$ in the S-curve can give it different shapes (Fig.~\ref{fig:swiss_s_fig}(b), for details see Appendix~\ref{sec:moresroll}). A color map shows the correspondence among the shapes. For $r<0.4$ and $r>0.6$, the colors overlap, and the 1-D manifold disappears. For experiments, we sampled 1000 points from each of the shapes and computed the alignment between them.

CKA with $\sigma=M$ fails to align the manifold of the Swiss-roll and S-curve ($r=0.5$), giving a lower value  (Fig.~\ref{fig:swiss_s_fig}(c)). However, for cases where the 1-D  manifold structure is absent (e.g., $r<0.4$ and $r>0.6$), CKA provides a higher value. On the contrary, CKA with $\delta=0.2$, kCKA, and MKA properly capture the alignment of the two shapes. At $r=0.5$, the alignment of the Swiss-roll and S-curve is highest and gets lower as the parameter moves away from this point. RTD and IMD do not show any trends, while SVCCA shows an unrelated oscillatory behavior (from curvature). However, kCKA is more sensitive to the number of nearest neighbors $k$ (Fig.~\ref{fig:swiss_s_fig}(d)), while MKA is very robust to that. 

We identify two mechanisms underlying the greater sensitivity of kCKA to k. First, changing k alters the support of every kernel row by adding or removing neighbors; this effect is shared by both kCKA and MKA. Second, unlike MKA, kCKA does not impose a constraint that controls the resulting row-wise perturbation. In our implementation, kCKA recomputes a global bandwidth $\sigma$ as $k$ changes, so the weights of previously retained neighbors may also change throughout the kernel. Even if $\sigma$ was fixed, the weight of a newly added neighbor is not guaranteed to be close to zero. Therefore, the row-wise kernel difference is not guaranteed to satisfy property (A3). By contrast, MKA calibrates a local bandwidth $\sigma_i$ under the prescribed row-mass constraint $\sum_j K_{ij}=1+\log_2k$, which controls the additional mass introduced as $k$ increases and yields stability guarantees for adjacent neighbor sizes $k$ and $k+1$ in Theorems~\ref{thm:kernel_stability}-\ref{thm:row_stability} and Corollary~\ref{cor:mka_implement_Stability}. The stability observed across the wider range of $k$ in Fig.~\ref{fig:swiss_s_fig}(d) is empirical and is not implied by a uniform theoretical bound over that range.

\subsection{Ranking Structures}

\begin{figure*}[t]
    \centering
    \includegraphics[width=1.0\linewidth]{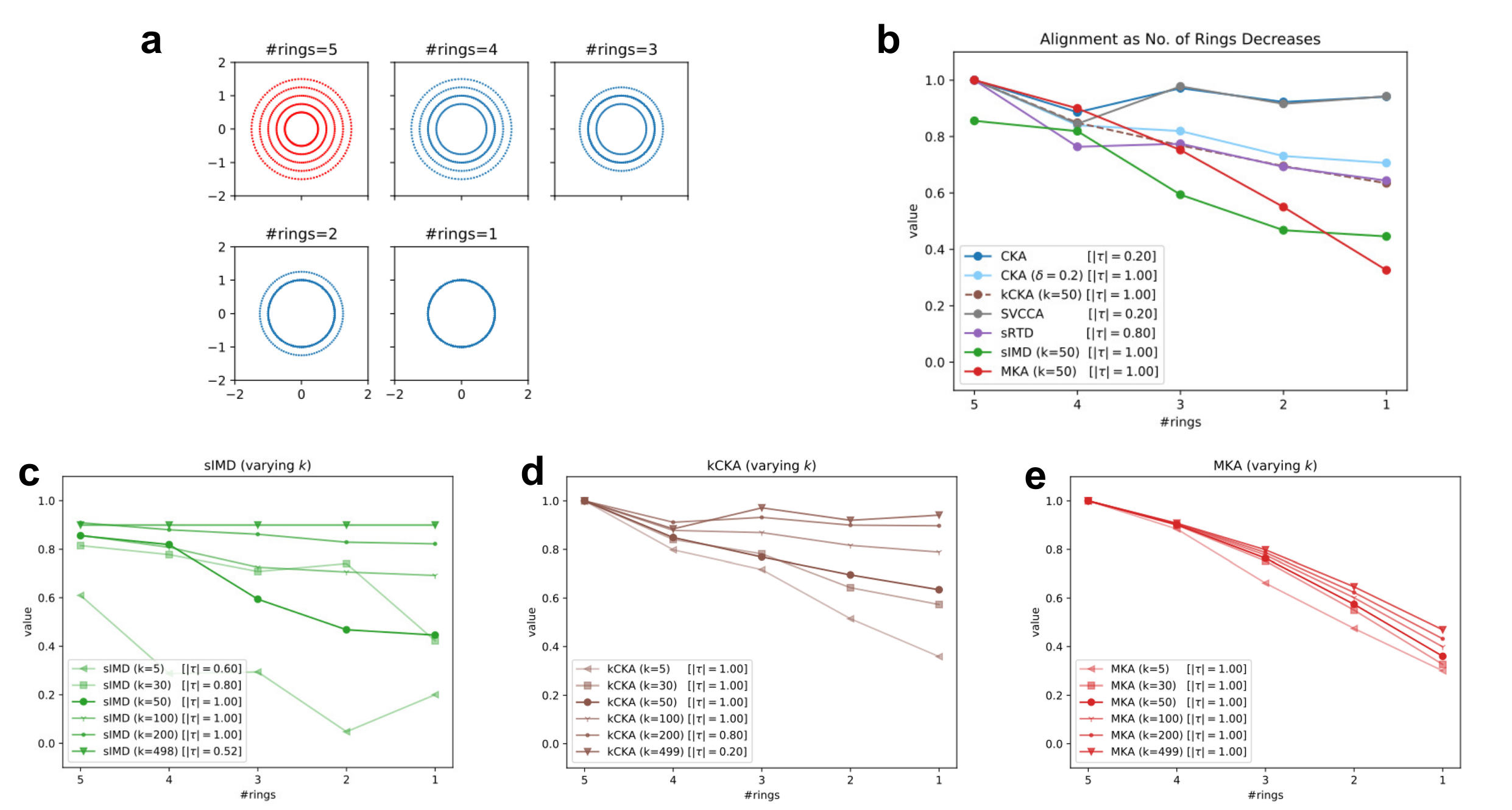}
    \caption{Alignment for the ``rings'' data. (a) Point clouds used in the clusters experiment. (b) Alignment using various methods, along with Kendall's rank correlation ($\tau$, higher is better). (c-e) Alignment by varying nearest neighbors, $k$, in (c) IMD, (d) kCKA, and (e) MKA. MKA shows the greatest robustness to the parameter $k$.}
    \label{fig:ranking_rings}
\end{figure*}

\begin{figure*}[t]
    \centering
    \includegraphics[width=1.0\linewidth]{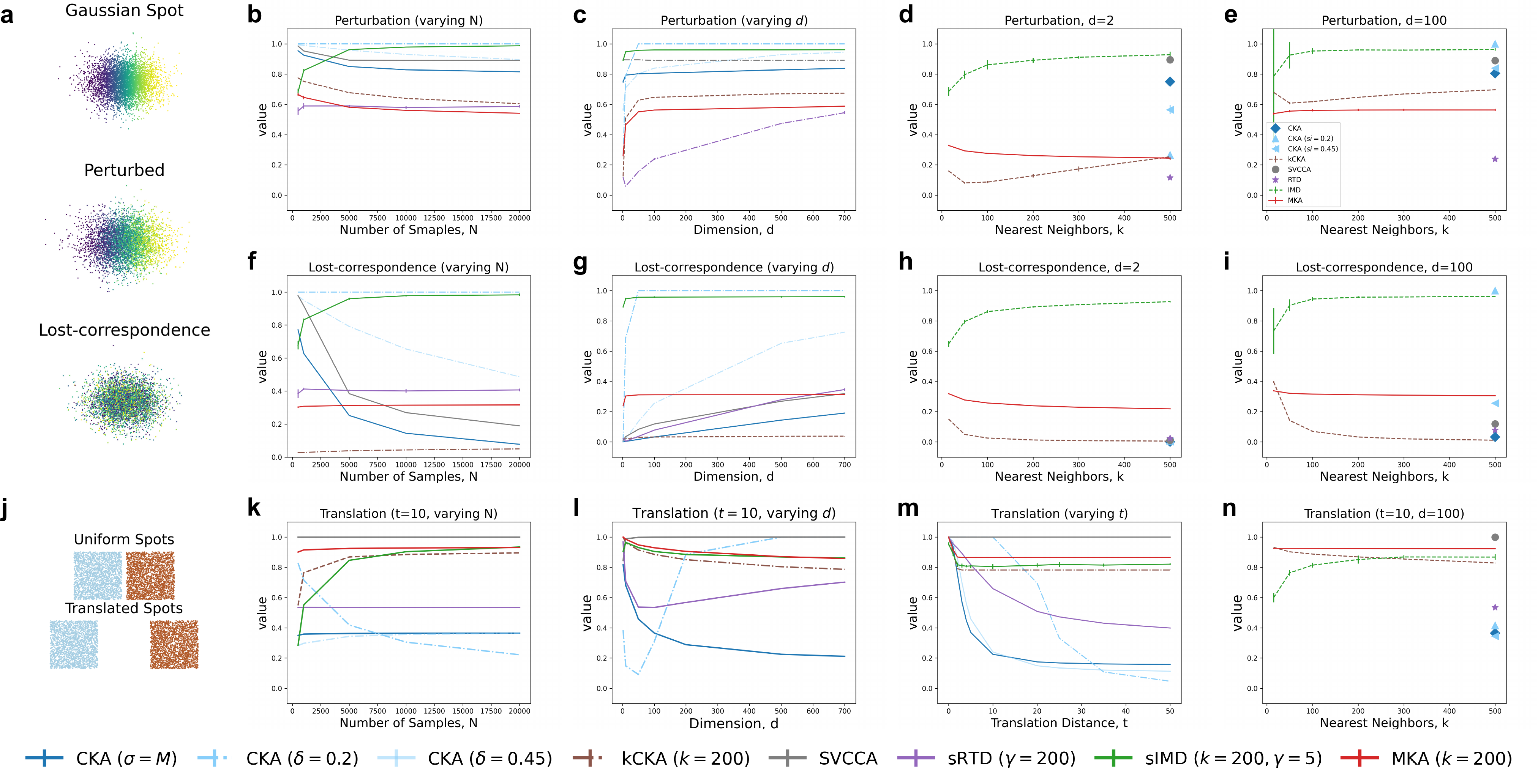}
    \caption{Characterizing MKA using synthetic datasets and comparison to other methods. \textbf{(a)} Top: A Gaussian spot; colors identify the position of the points on the x-axis. Middle: Perturbed Gaussian spot. We added noise to the points of the top figure so that the colors slightly overlap. Bottom: A Gaussian spot with no correspondence to the spot on the top. 
    \textbf{(b-e)} Alignment between a Gaussian spot and when it is perturbed when (b) number of samples, $N$ ($d=1000$), and (c) number of dimensions, $d$ ($N=5000$), varies for various methods, and their performance as the number of nearest neighbors, $k$, varies for (d) $d=2$ and (e) $d=100$ ($N=5000$).
    \textbf{(f-i)} Alignment under lost correspondence when (f) number of samples, $N$ ($d=1000$), and (g) number of dimensions, $d$ ($N=5000$), varies for various methods, and their performance as the number of nearest neighbors, $k$, varies for (h) $d=2$ and (i) $d=100$ ($N=5000$).
    \textbf{(j)} Two uniform spots are located nearby (top) and translated far away (bottom).
    \textbf{(k-n)} Alignment under translation when (k) number of samples, $N$ ($d=1000$), (l) number of dimensions, $d$ ($N=5000$), (m) translation distance, $t$, and (n) number of nearest neighbors, $k$, varies.
    Error bars are drawn up to one standard deviation (5 trials for each experiment).}
    \label{fig:gauss-fig}
\end{figure*}

In the second test, we reproduce the ``rings'' experiments that originally appeared in~\citep{barannikov2021representation}. This dataset consists of 500 points distributed over five concentric rings (radii varying from 0.5 to 1.5). Then, in each iteration, the number of rings decreases (as if one of the rings collapses onto another ring) until it reaches a singular ring (Fig.~\ref{fig:ranking_rings} (a)). 
Then, we use alignment metrics to compare these formations with the original structure (i.e., five rings). The target of the experiment is to check whether the metrics can track the collapsing rings structure. Kendall's rank correlation, $\tau$, can measure this ranking in a statistical sense (the absolute value is sufficient and thus higher is better).

CKA and SVCCA fail to track this collapsing behavior, while RTD closely reflects the changes. CKA ($\delta=0.2$), kCKA, IMD, and MKA capture the ranking quite well (Fig.~\ref{fig:ranking_rings}(b)).  However, varying the nearest neighbor parameter, $k$, causes different behaviors in different methods. IMD shows consistent behavior for $k=50$, $100$, and $200$ (Fig.~\ref{fig:ranking_rings}(c)). kCKA provides correct ranking only for lower values of $k$; at higher values $k \approx 200$ and above the method fails (Fig.~\ref{fig:ranking_rings}(d)). MKA provides correct ranking for all possible values of $k$ (Fig.~\ref{fig:ranking_rings}(e)). To the best of our knowledge, ours is the first paper that achieves correct ranking in this dataset.

We explored an additional dataset of similar interest: ``clusters''. The results are provided in Fig.~\ref{fig:clusters_break}(a) in Appendix~\ref{sec:clusters_data}.


\subsection{Characterizing The Algorithms}\label{sec:characterize}

In this section, we characterize the algorithms using several synthetic datasets inspired by real-world scenarios. First, we consider the alignment between a d-dimensional Gaussian spot ($x_i\sim\mathcal{N}(\mathbf{0},I_d)$, Fig.~\ref{fig:gauss-fig}(a) top) and its perturbed version ($y_i=x_i+0.5\mathcal{N}(\mathbf{0},I_d)$, Fig.~\ref{fig:gauss-fig}(a) middle). 
Such a scenario may occur when a representation learning algorithm runs repeatedly. This results in altered orders of the points in the point cloud (seen as colors slightly overlapping). 
As the number of samples in the spots increases ($d=1000$, Fig.~\ref{fig:gauss-fig}(e)), their alignment values using different methods decrease slightly (notable exceptions are IMD, which increases and then stabilizes, and CKA ($\delta=0.2$), which saturates). 
This is expected, as the denser the spot gets, the higher the chance of orders within the point cloud. 
However, the dimensionality ($d$) of the data affects the values differently ($N=5000$, Fig.~\ref{fig:gauss-fig}(c)). 
All methods, except CKA with $\delta=0.2, 0.45$ and RTD, are fairly consistent as $d$ increases. 
CKA with $\delta=0.2$ saturates rapidly, while with $\delta=0.45$ it approaches saturation as $d$ increases. RTD starts with a lower value, and it increases with $d$. 
Additionally, $\kcka$ shows inconsistent behavior as the number of nearest neighbors ($k$) increases and sIMD shows high variance, while $\mka$ values remain consistent across a wide range (Fig.~\ref{fig:gauss-fig}(d,e)). 
Overall, $\mka$ is more restrictive to perturbations in the features than other methods.

We can take this scenario to the extreme and make the colors completely overlap each other (Fig.~\ref{fig:gauss-fig}(a), bottom). 
The orderings (based on some criterion) of both the Gaussian spots will not correspond to each other at all, and thus, we call it a lost-correspondence scenario. 
The $\cka$ (and $\delta=0.45$) and SVCCA  measures are sensitive to the number of samples, while $\kcka$, RTD, and $\mka$ are fairly consistent ($d=1000$, Fig.~\ref{fig:gauss-fig}(f)). 
The CKA ($\delta=0.45$ as well) measure tends to increase with higher data dimensionality, reflecting the effect of the curse of dimensionality ($N=5000$, Fig.~\ref{fig:gauss-fig}(g)). 
SVCCA and RTD also behave similarly. 
$\kcka$, IMD, and $\mka$, on the other hand, are fairly robust and less affected by the curse.  
However, like before, kCKA is highly sensitive to the number of nearest neighbors ($k$), which gets resolved at a higher value of $k\geq200$ (Fig.~\ref{fig:gauss-fig}(h,i)). 
Like previously, MKA is consistent for a wide range of $k$, even for values smaller than $200$. 
Overall, $\mka$ is more consistent with varying hyperparameters than other methods.

Finally, we consider two uniform spots separated by a small distance (Fig.\ref{fig:gauss-fig}(j); this scenario is inspired by~\cite {davarireliability}). 
Both spots ($N=2500$ each) are drawn from a uniform distribution as $x_i\sim\mathcal{U}(-0.5,0.5)$ and $y_i\sim p+\mathcal{U}(-0.5,0.5)$ with $p=[1.1+t, 0, 0, \dots, 0]$, where the translation distance, $t (>0)$, controls the separation of the two spots. 
Regardless of the translation distance, the topology of the data remains the same, and alignment should be high. 
Surprisingly, most methods are consistent as the number of samples increases (except CKA with $\delta=0.2$). 
CKA gives a low alignment score between the two representations, while kCKA and IMD stabilize as the number of samples increases. 
We get a more diverse result as the number of dimensions, $d$, (Fig.~\ref{fig:gauss-fig}(l)) and the translation distance, $t$, (Fig.~\ref{fig:gauss-fig}(m)) vary.
$\cka$ fails to capture this phenomenon. As $t$ increases, $\cka$ value decreases; even using a smaller bandwidth $\delta=0.2$ fails. 
Surprisingly, RTD also joins CKA and fails to capture the invariance of topology (RTD fuses individual graphs of the two representations into a single graph. In the fused graph, the large distances dominate and thus RTD follows CKA). SVCCA shows maximum alignment between the two representations under all circumstances. 
In contrast, kCKA, IMD, and MKA settle to a constant and higher number as $d$ and $t$ increase. 
As $k$ increases, the pattern mirrors the earlier experiments; by $k\simeq100$ most methods stabilize, whereas MKA is already consistent at small $k$ (Fig.~\ref{fig:gauss-fig}(n)).

\subsection{Evaluation using Representation Similarity (ReSi) Benchmark}

\begin{figure*}[t]

    \centering
    \includegraphics[width=1.0\linewidth]{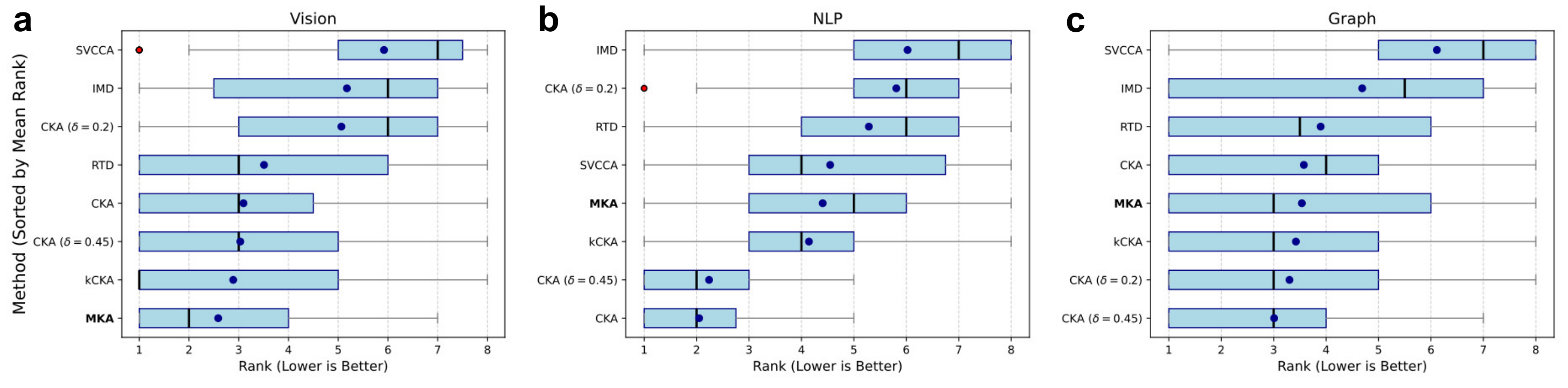}
    \caption{Aggregated ranks of alignment measures using the ReSi benchmark across different models and tests, separated by domains: (a) vision, (b) natural language processing, and (c) graph. Boxplots indicate quartiles of rank distributions; the whiskers extend up to 1.5 times the interquartile range. The black dots indicate the mean rank.}
    \label{fig:resi}
\end{figure*}

Representation Similarity (ReSi) Benchmark~\citep{klabunde2024resi} is a collection of six different tests to assess the performance of representational similarity or alignment metrics. 
The tests are Correlation to Accuracy Difference (correlates the alignment score of a pair of models with the absolute difference in their accuracies), Correlation to Output Difference (correlates alignment metrics with the instance-wise disagreement and Jensen-Shannon divergence of the predictions), Label Randomization (evaluates whether alignment metrics can separate models trained with varying levels of label corruption), Shortcut Affinity (evaluates whether alignment metrics can distinguish models trained with spurious shortcut features at different shortcut–label correlation strengths), Augmentation (evaluates whether alignment metrics can stratify models trained with varying augmentation strengths, when all are tested on the same clean, non-augmented set), and Layer Monotonicity (evaluates whether alignment score decreases as the distance between layers increases within the same model). We used the ReSi tests on vision, natural language processing (NLP), and graph domain tasks. For the vision task, we used the ImageNet-100 dataset and seven representative networks from three different architectures: Residual Networks (ResNet-18, ResNet-34, ResNet-101)~\citep{he2016deep}, Visual Geometry Group networks (VGG-11, VGG-19)~\citep{simonyan2014very}, and Vision Transformers (ViT B32, ViT L32)~\citep{dosovitskiy2020image}. For the language task, we used the MNLI dataset~\citep{williams2017broad} and two language models: BERT~\citep{devlin2019bert} and ALBERT~\citep{lan2019albert}. For the graph data, we explored three different datasets: Cora \citep{kipf2016semi}, Flickr~\citep{hamilton2017inductive}, and OGBN-Arxiv~\citep{velivckovic2017graph}, and four different graph networks: Graph Convolutional Network (GCN)~\citep{yang2016revisiting}, Graph Sample and Aggregate (SAGE)~\citep{zeng2019graphsaint}, Graph Attention Network (GAT)~\citep{hu2020open}, and Position-aware Graph Neural Networks (PGNN)~\citep{you2019position}. The original ReSi benchmark concluded that no method consistently outperforms others across domains. We expect to find a similar result here as well.

Figure~\ref{fig:resi} summarizes mean-rank distributions per domain (lower is better). We used $k=100$ to compute the nearest neighbor graphs. 
In the vision domain, MKA attains the best central tendency with the tightest spread, edging out kCKA and clearly outperforming CKA and RTD (Fig.~\ref{fig:resi}(a)). 
By contrast, in the NLP domain, CKA and CKA with $\delta=0.45$ achieve lower mean and median ranks than MKA (Fig.~\ref{fig:resi}(b)).
Finally, for graphs, the methods that focus on local geometry, i.e., MKA, kCKA, and CKA ($\delta=0.2,0.45$), cluster together (same median for all of them and the mean is within $\pm1$). 
Overall, MKA performs best in aggregate in vision, is comparable to the strongest locality-based methods on graphs, and is weaker than CKA in several NLP evaluations.

One possible interpretation is that MKA performs well when benchmark-relevant information is reflected in local neighborhood structure, whereas CKA can be preferable when global pairwise relationships are more informative. Establishing whether this geometric distinction explains the observed differences between domains requires further study. For graphs, the methods perform comparably. Note that the ReSi benchmark does not probe dimensionality and sample numbers (as we have done in Section~\ref{sec:characterize}).

\subsection{Neural Network Representations}
\begin{figure*}
    \centering
    \includegraphics[width=\linewidth]{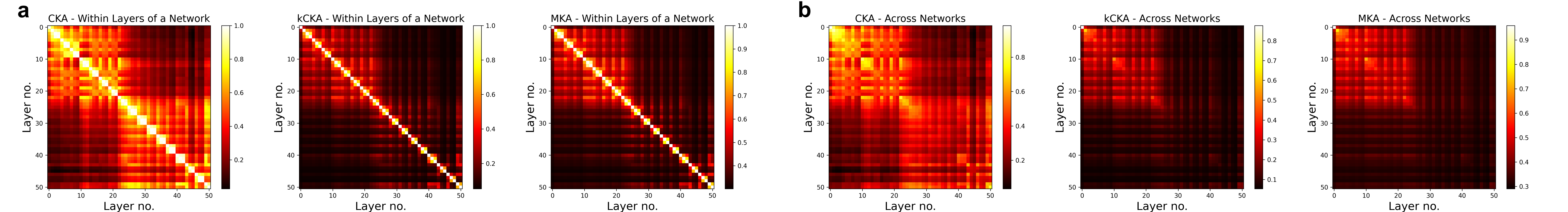}
    \caption{Alignment between features from different layers of ResNet-50 trained on the CIFAR-10 dataset. (a) Alignment between layers of a network using (left) $\cka$, (middle) kCKA, and (right) $\mka$. (b) Alignment between layers across different networks using (left) $\cka$, (middle) kCKA, and (right) $\mka$. The results are an average of 10 instances of ResNet-50 trained on CIFAR-10, each initialized randomly and using a subset of $10000$ samples from the test set.}
    \label{fig:resnet50}
\end{figure*}
In this section, we explore the representational similarity using ResNet-50 models trained on the CIFAR-10 dataset. 
First, we compute alignment between feature representations extracted from different layers (after activation) of the network to investigate how representational structure evolves across the depth of the model (Fig.~\ref{fig:resnet50}(a)). 
We considered only CKA, kCKA, and MKA for this experiment (as other methods have been explored elsewhere) and highlight how these three competing methods process information. 
Using CKA, we can reproduce the famous block structure~\citep{kornblith2019similarity,nguyen2022origins}. 
As we said previously, dominant clusters cause the block structure~\citep{nguyen2022origins}. 
However, when a k-nearest neighbor graph constrains the kernel, this block structure disappears. For kCKA, block structure appears in the early layers, but it is less pronounced in the later layers. MKA exhibits a similar behavior; however, for adjacent layers ($\pm1$ around the diagonal), it shows higher alignment than kCKA, suggesting equivalence of the directed neighborhood graph, whereas kCKA additionally reflects within-neighborhood distances.
Overall, CKA is sensitive to dominant high-density regions of large distances in the distance matrix compared to kCKA, and MKA is even less so. When we compare features of the networks to each other, the block structure is still present for $\cka$, less pronounced for kCKA and $\mka$ (Fig.~\ref{fig:resnet50}(b)). 
This suggests the hypothesis that the same architecture, under different random initializations, can converge to distinct internal orientations, i.e., manifold-level perturbations of the learned representation, despite similar test accuracy; this effect is often less apparent under CKA.

\section{Discussion and Conclusions}\label{sec:conclusions}

In this paper, we introduced Manifold-approximated Kernel Alignment (MKA), provided theoretical arguments to explain the algorithm, characterized it using several datasets, and compared it to different algorithms. Mechanistically, MKA's local weighting rule is closely related to that used in UMAP. The methodological contribution is to preserve the directed and equal-row-mass structure of these affinities when constructing a representation-alignment measure. This differs from manifold-learning pipelines that symmetrize the graph for embedding and from sparse CKA variants that do not control the row masses. Preserving these properties supports a closed-form alignment score (Theorem~\ref{thm:mka_simple}), a transition-probability interpretation without the degree weighting (Theorem~\ref{thm:mka_markov}), and permits the sensitivity to $k$ to be analyzed for the chosen kernel (Theorems~\ref{thm:kernel_stability}~and~\ref{thm:row_stability} and Corollary~\ref{cor:mka_implement_Stability}).

Experimentally, we found that methods using k-NN graphs are suitable for comparing the selected topology-motivated synthetic tests, including the Swiss-roll/S-curve, rings, and clusters experiments (MKA and kCKA in Figs.~\ref{fig:swiss_s_fig},\ref{fig:ranking_rings}, and \ref{fig:clusters_break}) and sometimes perform even better than their topological counterparts. By analyzing Gaussian distributions and their perturbations (Fig.~\ref{fig:gauss-fig}), we showed that methods that rely on local neighborhoods show less sensitivity to intrinsic parameters of datasets (number of samples and dimensionality). However, most methods require hyperparameter tuning, while MKA is consistent with varying hyperparameters. When tested with uniform spots and their translation, we found MKA to be robust, even compared to other topological methods (Fig.~\ref{fig:gauss-fig}(j-m)). We then show that on the ReSi benchmark, MKA achieves its strongest aggregate performance in vision and performs comparably to other locality-based measures on graphs, while CKA remains stronger in several NLP evaluations (Fig.~\ref{fig:resi}). By analyzing representations of neural networks, we conclude that $\mka$ perceives the neural network representations differently than CKA, with kCKA working as a conceptually intermediate step.

These observations clarify the distinction among the three kernel-based measures. CKA is globally density-weighted: a single high-density region of large distances in the kernel matrix can dominate the score. kCKA mitigates this by restricting interactions to local k-NN neighborhoods, making it less susceptible to interactions from large distances. MKA goes further by ordering neighbors within each neighborhood and assigning weights that depend on rank and local density. 
In essence, vanilla CKA ignores ranks and depends solely on pairwise distances. kCKA, on the other hand, dichotomizes pairs into ``within-k'' vs ``outside-k'', reducing the contribution of distant pairs, but it retains globally bandwidth-scaled affinities and unconstrained row masses.
RTD, a topology-based approach, sometimes aligns with the underlying manifold structure, but in other cases exhibits behavior closer to CKA. We attribute this to RTD’s graph-fusion step, which can overweight long-range edges, i.e., large distances tend to persist across filtration scales and thus dominate the resulting barcodes. Accordingly, RTD primarily reflects multi-scale persistence, whereas kCKA and MKA are single-scale methods based on k-NN neighborhoods.
At a fixed k, the k-NN graph is either faithful or not; persistence, by averaging over scales, can smooth away local structure, occasionally drifting toward density-driven behavior.

Several limitations remain for exploration in the future. 
MKA depends on the chosen distance metric and neighborhood size. 
The stability results control changes from $k$ to $k+1$ under the stated assumptions, but do not provide a uniform stability guarantee over a wide range of $k$, nor imply that every value of $k$ is equally informative. 
Its local construction may also discard global relationships that are useful in some settings, consistent with the stronger performance of CKA in several NLP evaluations. 
Moreover, row-only centering retains an incoming-mass component. Proposition~\ref{prop:hubness_effect} quantifies its contribution and shows when MKA approaches its double-centered counterpart, but does not eliminate it. 
Finally, exact nearest-neighbor and distance computations can dominate the end-to-end runtime, although sparse storage and approximate-neighbor methods may improve scaling.

In future, one could explore other kernel functions, e.g., effective resistance~\citep{doyle1984random}, diffusion distance~\citep{coifman2006diffusion}, and $\alpha$-decaying kernels~\citep{moon2019visualizing}. Further work could also investigate principled strategies for selecting the neighborhood size and distance metric, determine when local and global alignment criteria are most informative, and evaluate MKA in applications such as neural decoding, brain-representation analysis, model comparison, and graph or protein-representation learning. Overall, the results support MKA as a theoretically interpretable and empirically stable complementary alignment measure, particularly when agreement of local neighborhood geometry is central to the comparison.

\section*{Acknowledgment}

Mohammad Tariqul Islam is supported by MIT-Novo Nordisk Artificial Intelligence Fellowship. Special thanks Baju C. Joy and Pengrui Zhang for the discussion.

\section*{Software and Data}

To install $\mka$ on you computer visit: \href{https://github.com/tariqul-islam/mka}{https://github.com/tariqul-islam/mka}

Codes to reproduce the results are available at: \href{https://github.com/tariqul-islam/mka_paper_code}{https://github.com/tariqul-islam/mka\_paper\_code}

All the data used in this paper are publicly available.


\bibliography{thesis,kernel_alignment,du_references}
\bibliographystyle{tmlr}

\newpage
\appendix
\onecolumn
\section*{Appendix}

In the supplementary material, we provide some additional details and results. Appendix~\ref{sec:proofs} provides the proofs of the theorems. Appendix~\ref{sec:eta} discusses the range of $\eta$ required for the validity of the statements. 
In Appendix~\ref{sec:algorithm} we provide the pseudocode for computing MKA (along with $\sigma_i$ computations) and $k$CKA metric. 
Appendix~\ref{sec:row_centering_appendix} provides numerical results comparing MKA and its double-centered counterparts. 
In Appendix~\ref{sec:more_cka}, we discuss CKA with manifold approximation. 
In Appendix~\ref{sec:linear_kernel}, we discuss the linear kernel of kCKA. 
In Appendix~\ref{sec:complexity}, we provide a short comment on the computational complexity of the algorithms and provide timing decompositions of CKA, $k$CKA, and MKA. 
Appendix~\ref{sec:clusters_data} provides details of the ``clusters'' data experiment.
In Appendix~\ref{sec:alternate_row_sum}, we discuss a few other row-sum constraints. Appendix~\ref{sec:cknna_kcka} discusses CKNNA, another variant of CKA that uses a mutual k-NN kernel.
Appendix~\ref{sec:moresroll} gives details of the Swiss-roll and S-curve. 
Appendix~\ref{sec:rtdimdraw} gives supplementary figures for the experiments in the main text. 
We provide implementation details in Appendix~\ref{sec:implementaiton_Details}. 
Finally, we follow it by detailing the ReSi benchmark in Appendix~\ref{sec:resi_details} and corresponding supplementary results in Section~\ref{sec:resi_scores}.

\section{Proofs}\label{sec:proofs}

\begin{proof}[(Proof of Theorem~\ref{thm:mka_simple})]
Let $KH=\bar{K}$ and $LH=\bar{L}$. Using the constant row sum property we show:
\begin{align}
    \bar{K}_{ij} &= K_{ij} - \frac{1}{N} \sum_{j} K_{ij}  \nonumber \\
            &= K_{ij} - \frac{1}{N} D.
\end{align}
Now, we can compute the inner product,
\begin{align}
    \langle \bar{K}, \bar{K} \rangle &= \sum_{i,j} (K_{ij}-\frac{1}{N} D)^2 \nonumber \\
    &= \sum_{i,j} \left( K_{ij}^2 - \frac{2}{N}D K_{ij} + \frac{1}{N^2} D^2 \right) \nonumber \\
    &= \sum_{i,j} K_{ij}^2 - \frac{2}{N}D \sum_{i,j} K_{ij} + \frac{1}{N^2} D^2 \sum_{i,j} 1 \nonumber \\
    &= \sum_{i,j} K_{ij}^2 - D^2 \nonumber \\
    &= \langle K, K \rangle  - D^2
\end{align}
We used the fact that $\sum_{i,j}K_{ij}=ND$ and $\sum_{i,j}1=N^2$. Similarly, $\bar{L}_{ij}=L_{ij} - \frac{1}{N} D$ and $\langle \bar{L}, \bar{L} \rangle = \langle L, L \rangle - D^2$. Finally,
\begin{align}
    \langle \bar{K}, \bar{L} \rangle &= \sum_{i,j} (K_{ij} - \frac{1}{N} D) (L_{ij} - \frac{1}{N} D) \nonumber \\
    &= \sum_{i,j} K_{ij} L_{ij} - \frac{1}{N} D (K_{ij}+L_{ij}) - \frac{1}{N^2} D^2 \nonumber \\
    &= \sum_{i,j} K_{ij} L_{ij} - D^2 \nonumber \\
    &= \langle K, L \rangle - D^2
\end{align}
\end{proof}

\begin{proof}[(Proof of Corollary~\ref{thm:mka_range})]
We start from the inner products,
\begin{align}
    \langle K, K \rangle - D^2 &= \sum_{i,j} K_{ij}^2 -D^2 \nonumber \\
    &= \sum_{i,i} 1 + \sum_{i,j, i\neq j} K_{ij}^2 - D^2 \nonumber \\
    &= N - D^2 + \sum_{i,j, i\neq j} K_{ij}^2.
\end{align}
Similarly, 
\begin{align}
    \langle L, L \rangle - D^2 &= N - D^2 + \sum_{i,j, i\neq j} L_{ij}^2
\end{align}
And finally, 
\begin{align}
    \langle K, L \rangle - D^2 &= N - D^2 + \sum_{i,j, i\neq j} K_{ij} L_{ij}
\end{align}
The value $\sum_{i,j, i\neq j} K_{ij} L_{ij}$ can be zero if the nearest neighbors in the kernels do not overlap. Otherwise, this value is positive. Thus, the lower bound is guaranteed when $N>D^2$. The upper bound is due to Cauchy--Schwarz inequality.
\end{proof}

\begin{proof}[Proof of Corollary~\ref{cor:metric_stability}]
    We can write MKA as a cosine similarity using Frobenius norm:
    \begin{align}
        MKA(K(\theta_k),L) = <\hat A_{(k)},\hat B>_F
    \end{align}
where $A_{k}=K(\theta_k)H$, $\hat A=\frac{A}{||A||_F}$, $B=LH$ and $\hat B=\frac{B}{||B||_F}$. Then, using Cauchy--Schwarz and Massera-Sch{\"a}ffer~\citep{massera1958linear,maligranda2006simple} inequalities:
\begin{align}
    |MKA(K(\theta_{k+1}),L)-MKA(K(\theta_k),L)| &= |<\hat A_{k+1}-\hat A_{k},\hat B>_F| \\
    & \leq ||\hat A_{k+1}-\hat A_{k}||_F ||\hat B||_F \\
    & = ||\hat A_{k+1}- \hat A_{k}||_F \\
    & \leq \frac{2 ||A_{k+1}- A_{k}||_F}{\max{\{||A_{k+1}||_F,||A_k||_F\}}} \label{eq:MKA_inequality_massera}
\end{align}
    We treat the numerator first. Let $\max_{i\in[n]} ||K(\theta_{k+1})_{i,:}-K(\theta_{k})_{i,:}||_1 \leq \epsilon_k$ (Property (A3)). Then, we have:
    \begin{align}
        ||A_{k+1}- A_{k}||_F^2 &= ||(K(\theta_{k+1})-K(\theta_{k}))H||_F^2 \\
        & \leq ||K(\theta_{k+1})-K(\theta_{k})||_F^2 \\
        &= \sum_i ||K_{i,:}(\theta_{k+1})-K_{i,:}(\theta_{k})||_2^2 \\
        &\leq \sum_i ||K_{i,:}(\theta_{k+1})-K_{i,:}(\theta_{k})||_1^2 \\
        &\leq n \epsilon^2 \label{eq:cor_4_3_num}
    \end{align}
    Now, we focus on the denominator. Using the results from Theorem~\ref{thm:mka_simple} and Corollary~\ref{thm:mka_range}:
    \begin{align}
        ||A_{k}||_F^2 &= ||K(\theta_{k})||_F^2 - D_k^2 \\
        &\geq n-D_k^2.
    \end{align}
    Similarly, $||A_{k+1}||_F^2\geq n-D_{k+1}^2$. Using the assumption $\max{\{D_{k+1}^2,D_k^2\}}\leq(1-\eta)n$, we have $\max{\{||A_{k+1}||_F,||A_k||_F\}}\geq \sqrt{\eta n}$, which together with Eq.~(\ref{eq:cor_4_3_num}) gives the desired result.
\end{proof}

\begin{proof}[Proof of Theorem~\ref{thm:cka_markov}]
    A simple calculation shows that
    \begin{align}
        (KH)_{ij} &= K_{ij} - \frac{1}{n}\sum_j K_{ij} \\
        &= K_{ij} - \frac{s_i}{n} \\
        &= s_i\Big(\frac{K_{ij}}{s_i} - \frac{1}{n}\Big) \\
        &= s_i\big(P_{ij} - U_{ij}\big),
    \end{align}
    Thus, in matrix form it becomes, $KH = S(P-U)$, and  $LH = T(Q-U)$. Here, $P-U$ and $Q-U$ are the grand-mean-centered Markov kernels, as defined in~\cite{prothero2021new}. The derivation of the matrices $P$ and $Q$ follows from standard graph to Markov notations~\citep{lovasz1993random,coifman2006diffusion}.
    This makes
    \begin{align}
        HKH = HS(P-U) \text{~~~and~~~} 
        HLH = HT(Q-U).
    \end{align}
\end{proof}

\begin{proof}[Proof of Theorem~\ref{thm:mka_markov}]
    For manifold-approximated kernels following property (A2), $\sum_j K_{ij} = \sum_j L_{ij} = D$.
\end{proof}

\begin{proof}[Proof of Proposition~\ref{prop:hubness_effect}]
Let $A=P-U$, and $B=Q-U$. Then $\mka=\frac{\langle A,B\rangle_F}{\|A\|_F\|B\|_F}$, and $\mka_{DC}=\frac{\langle HA,HB\rangle_F}{\|HA\|_F\|HB\|_F}$. Since $I=H+U$, we can decompose $A$ as 
\begin{align}
    A=HA+UA=HA + \frac{1}{n} \mathbf{1}h_{p}^\top. \label{eq:orthogonal_eq}
\end{align} 

The two components in Eq.~(\ref{eq:orthogonal_eq}) are orthogonal. Similarly, we obtain $B=HB + \frac{1}{n} \mathbf{1}h_{q}^\top$. Interestingly, $HA$ and $\frac{1}{n} \mathbf{1}h_{q}^\top$ are orthogonal to each other (and consequently, $HB$ and $\frac{1}{n} \mathbf{1}h_{p}^\top$ are orthogonal to each other) as well. Consequently, 
\begin{align}
\langle A,B\rangle_F &= \langle HA,HB\rangle_F + \frac{1}{n}h_p^\top h_q. \label{eq:hubness_inner_product} \\
\|A\|_F^2 &= \|HA\|_F^2 + \frac{1}{n}\|h_p\|_2^2, \\
\|B\|_F^2 &= \|HB\|_F^2 + \frac{1}{n}\|h_q\|_2^2,
\end{align}

Putting these identities in the definition of $\mka$, and using the definitions of $\phi_p$, $\phi_q$ and $s_{\text{hub}}$, we obtain the result in Eq.~(\ref{eq:mka_mka_dc_relation}). Subtracting $\mka_{DC}$ from both sides of this relation gives:
\begin{align}
\mka-\mka_{DC} ={}& \left[\sqrt{(1-\phi_p)(1-\phi_q)}-1\right]\mka_{DC} + \sqrt{\phi_p\phi_q}\,s_{\mathrm{hub}}.
\end{align}
Since $|\mka_{DC}|\leq1$ and $|s_{\mathrm{hub}}|\leq1$ (both are cosine similarities), we reach the inequality~(\ref{eq:mka_mka_dc_inequality}). Under the assumptions $\phi_p\leq\epsilon$ and $\phi_q\leq\epsilon$, this inequality directly yields the bound in Eq.~(\ref{eq:mka_mka_dc_bound}).
\end{proof}

\begin{lemma}[Stability of the bandwidth under $k\to k+1$] \label{thm:sigma_lemma}

    Fix a point $x_i$, let the neighbor distances be
    \begin{align}
        d_1\leq d_2 \leq \dots, d_j = d(x_i, x_{(j)}) 
    \end{align}
    where $x_{(j)}$ is the $j^{th}$ nearest neighbor of $x_i$ and $\rho=d_1$, and define 
    \begin{align}
        \Delta_j = d_j-\rho.
    \end{align}
    Let $\sigma_{k}$ and $\sigma_{k+1}$ be two bandwidth parameters for two kernels obtained from Eq.~(\ref{eq:UMAP_HIGH_DIM}) for $k$ and $k+1$ nearest neighbors. Assume the following non-degeneracy bounds hold for the bandwidths:
    \begin{align}
        \Delta_2 \geq \delta > 0, \text{~and~} \sigma_k, \sigma_{k+1} \in [\underline{\sigma},\overline{\sigma}] \text{~for some~} 0<\underline{\sigma}<\overline{\sigma}<\infty. \label{eq:lemma_assumption}
    \end{align}
    Then $\forall$ $k\geq3$,
    \begin{align}
        |\sigma_{k+1}-\sigma_k| \leq \frac{\overline{\sigma}^2}{\delta (\log_2 3 -1)}  \left( \log_2\left(1+\frac{1}{k}\right) + \frac{\log_2 k - 1}{k-1} \right) = \mathcal{O}\left(\frac{\log_2 k}{k}\right)
    \end{align}
\end{lemma}
\begin{proof}
    Let
    \begin{align}
        F_k(\sigma) = \sum_{j=2}^k \exp(-\Delta_j/\sigma)
    \end{align}
    Then, $F_k(\sigma_k)=\log_2(k)-1$, $F_{k+1}(\sigma_{k+1})=\log_2(k+1)-1$ and $F_{k+1}(\sigma)=F_k(\sigma)+\exp(-\Delta_{k+1}/\sigma)$.
    By applying the mean value theorem, $\exists$ $\varepsilon$ between $\sigma_{k+1}$ and $\sigma_{k}$ such that,
    \begin{align}
        F_{k+1}(\sigma_{k+1}) - F_{k+1}(\sigma_k) = F'_{k+1}(\varepsilon) (\sigma_{k+1} - \sigma_k).
    \end{align}
    Thus, we obtain:
    \begin{align}
        |\sigma_{k+1} - \sigma_k| &= \frac{|F_{k+1}(\sigma_{k+1})-F_{k+1}(\sigma_k)|}{F'_{k+1}(\varepsilon)} \label{eq:start_ineq}
    \end{align}
    Let's focus on the numerator of the equation (\ref{eq:start_ineq}) first:
    \begin{align}
        |F_{k+1}(\sigma_{k+1})-F_{k+1}(\sigma_k)| &= \left|\log_2\left(1+\frac{1}{k}\right)-\exp\left(-\frac{\Delta_{k+1}}{\sigma_k}\right)\right| \\
        &\leq \log_2\left(1+\frac{1}{k}\right)+\exp\left(-\frac{\Delta_{k+1}}{\sigma_k}\right) \label{eq:fkfkineq_1}
    \end{align}
    To shape the exponential into a more desirable form, we notice that,
    \begin{align}
        \exp\left(-\frac{\Delta_{k+1}}{\sigma_k}\right) &\leq \exp\left(-\frac{\Delta_{k}}{\sigma_k}\right) \label{eq:logic_eq_start}\\
        &= \min_{2\leq j \leq k} \exp\left(-\frac{\Delta_{j}}{\sigma_k}\right) \\
        &\leq \frac{1}{k-1} F_k(\sigma_k)\\
        &= \frac{\log_2 k - 1}{k-1} \label{eq:logic_eq_end}
    \end{align}
    Thus, inequality~(\ref{eq:fkfkineq_1}) becomes:
    \begin{align}
        |F_{k+1}(\sigma_{k+1})-F_{k+1}(\sigma_k)| \leq \log_2\left(1+\frac{1}{k}\right) + \frac{\log_2 k -1}{k-1} \label{eq:fkfkineq_57}
    \end{align}
    Now, we focus on the denominator of the equation (\ref{eq:start_ineq}),
    \begin{align}
        F'_{k+1}(\varepsilon) &= \sum_{j=2}^{k+1} \frac{\Delta_j}{\varepsilon^2} \exp\left( -\frac{\Delta_j}{\varepsilon} \right) \\ 
                         & \geq \frac{\Delta_2}{\varepsilon^2} \sum_{j=2}^{k+1} \exp\left( -\frac{\Delta_j}{\varepsilon} \right) \\
                         & = \frac{\Delta_2}{\varepsilon^2} F_{k+1}(\varepsilon).
    \end{align}
    Since $F_{k+1}$ is monotonically increasing in $\sigma$, and $\varepsilon$ lies between $\sigma_{k+1}$ and $\sigma_k$, we have
    \begin{align}
        F_{k+1}(\varepsilon) &\geq \min{\{F_{k+1}(\sigma_{k+1}),F_{k+1}(\sigma_k)\}} \\
        &\geq \log_2 k - 1,
    \end{align}    
    where the second inequality follows from $F_{k+1}(\sigma_{k+1})=\log_2(k+1)-1$ and $F_{k+1}(\sigma_k)\ge F_k(\sigma_k)=\log_2 k-1$.
    For $k\geq3$,
    \begin{align}
        F_{k+1}(\varepsilon) \geq \log_2 3 - 1. \label{F_kp1log3}
    \end{align}
    Since $\sigma_k,\sigma_{k+1}\in[\underline{\sigma},\overline{\sigma}]$, and $\varepsilon$ lies between them, $\varepsilon\leq\overline{\sigma}$. Using this, the result in (\ref{F_kp1log3}), and the assumption $\Delta_2\geq\delta>0$, we have
    \begin{align}
       F'_{k+1}(\varepsilon) \geq \frac{\delta}{\overline{\sigma}^2} (\log_2 3 -1 ) > 0.\label{eq:deno_ineq}
    \end{align}
    Plugging in (\ref{eq:fkfkineq_57}) and (\ref{eq:deno_ineq}) into (\ref{eq:start_ineq}), we obtain the desired result. 
\end{proof}

\begin{proof}[Proof of Theorem~\ref{thm:kernel_stability}]
    The diagonal and the first nearest neighbor stay unchanged. $K_{ii}=1$ and $K_{i(1)}=\exp(0)=1$ for all $k$. Entries that are zero in both kernels also remain unchanged. 
    
    Now, let's consider the neighbors that exist in both graphs (ranks $r\leq k$). If $x_j=x_{(r)}$ with $2\leq r \leq k$, then
    \begin{align}
        K_{i,(r)}^{(k)} = \exp\left(-\frac{\Delta_r}{\sigma_k}\right) \text{~and~} K_{i,(r)}^{(k+1)} = \exp\left(-\frac{\Delta_r}{\sigma_{k+1}}\right)
    \end{align}
    These are the same functions with variables $\sigma_k$ and $\sigma_{k+1}$. Let $w(\sigma)=\exp(-\Delta_r/\sigma)$. Then using the mean value theorem, we have
    \begin{align}
        |w(\sigma_{k+1}) - w(\sigma_k)| \leq \sup_{\sigma>0} |w'(\sigma)| \cdot |\sigma_{k+1}-\sigma_{k}|
    \end{align}
    The supremum is attained at $\sigma=\Delta_r/2$. This results in
    \begin{align}
        |K_{i,(r)}^{(k+1} - K_{i,(r)}^{(k)}| \leq \frac{4}{e^2\Delta_r} |\sigma_{k+1}-\sigma_k|.
    \end{align}
    By assumption, $ \Delta_r \geq \Delta_2 \geq \delta > 0$ and using the result from Lemma~\ref{thm:sigma_lemma}, we have
    \begin{align}
        \left| K_{i,(r)}^{(k+1)} - K_{i,(r)}^{(k)} \right| \leq \frac{4}{e^2\delta} |\sigma_{k+1}-\sigma_k| = \mathcal{O}\left( \frac{\log_2 k}{k} \right)
    \end{align}
    Now we consider the new edge $(k+1)$-th neighbor:
    \begin{align}
        K_{i,(k+1)}^{(k)} = 0, \text{~~~} K_{i,(k+1)}^{(k+1)} = \exp\left( - \frac{\Delta_{k+1}}{\sigma_{k+1}} \right)
    \end{align}
    Following the same logic from Eqs.~(\ref{eq:logic_eq_start})-(\ref{eq:logic_eq_end}), we can show that
    \begin{align}
        \left|  K_{i,(k+1)}^{(k+1)} - K_{i,(k+1)}^{(k)} \right| = \exp\left( - \frac{\Delta_{k+1}}{\sigma_{k+1}} \right) \leq \frac{\log_2(k+1)-1}{k} = \mathcal{O}\left( \frac{\log_2 k}{k} \right)
    \end{align}
    
\end{proof}

\begin{proof}[Proof of Theorem~\ref{thm:row_stability}]
    For a fixed row $i$, let $a_i=K_{i,(k+1)}^{(k+1)}$ be the weight of the newly added neighbor and for the other common neighbors, i.e., $r=2,3,\dots,k$, we define, $c_{ir}=K_{i,(r)}^{(k+1)}-K_{i,(r)}^{(k)}$. Let $D_k = 1+ \log_2 k$, and thus we have
    \begin{align}
        \delta_k = D_{k+1}-D_{k} = \sum_{r=2}^{k}c_{ir} + a_{i}.
    \end{align}
    Consequently, 
    \begin{align}
        ||K_{i:}^{(k+1)}-K_{i:}^{(k)}||_1 &=  \sum_{r=2}^{k} |c_{ir}| + a_{i} \label{eq:kdiff_eq}
    \end{align}
    Now, for every common neighbor $r=2,\dots,k$,
    \begin{align}
        c_{ir} = K_{i,(r)}^{(k+1)}-K_{i,(r)}^{(k)} = \exp\left(-\frac{\Delta_r}{\sigma_{k+1}}\right) - \exp\left(-\frac{\Delta_r}{\sigma_k}\right).
    \end{align}
    Let $w_r(\sigma)=\exp\left(-\frac{\Delta_r}{\sigma}\right)$, then
    \begin{align}
        w_r'(\sigma) = \frac{\Delta_r}{\sigma^2} \exp\left(-\frac{\Delta_r}{\sigma}\right)\geq0.
    \end{align} 
    So, $w_r(\sigma)$ is non-decreasing in $\sigma$. Therefore, if \(\sigma_{k+1}\ge\sigma_k\), then \(c_{ir}\ge0\) for every common neighbor \(r\); if \(\sigma_{k+1}\le\sigma_k\), then \(c_{ir}\le0\) for every common neighbor. Hence all \(c_{ir}\) have the same weak sign. Consequently, 
    \begin{align}
        \sum_{r=2}^k |c_{ir}| = \left|\sum_{r=2}^k c_{ir}\right| = |\delta_k-a_i|.
    \end{align}
    Thus, Eq. (\ref{eq:kdiff_eq}) becomes,
    \begin{align}
        ||K_{i:}^{(k+1)}-K_{i:}^{(k)}||_1 &= |\delta_k-a_i| + a_i \\
                                          &\leq \delta_k + 2a_i. \label{eq:leqkp1mk}
    \end{align}
    The upper bound of $a_i$ is
    \begin{align}
        a_i \leq \frac{\log_2(k+1)-1}{k}.
    \end{align}
    This converts the inequality~(\ref{eq:leqkp1mk}) to
    \begin{align}
        ||K_{i:}^{(k+1)}-K_{i:}^{(k)}||_1 &\leq \log_2\left({1+\frac{1}{k}}\right) + 2 \frac{\log_2(k+1)-1}{k} \\
                                            &= \mathcal{O}\left(\frac{\log_2 k}{k}\right)
    \end{align}
\end{proof}

\begin{proof}[Proof of Corollary~\ref{cor:mka_implement_Stability}]
    By following Corollary~\ref{cor:metric_stability} and Theorem~\ref{thm:row_stability} we can achieve the one-sided result,
    \begin{align}
        |\mka(K^{(k+1)},L)-\mka(K^{(k)},L)|=O\left(\frac{\log_2 k}{k}\right),\qquad k\ge3.
    \end{align}
    This one-sided result is independent of the row-sum of the kernel $L$ (i.e., the value of row-sum can differ between $K$ and $L$, as the derivation in Corollary~\ref{cor:metric_stability} uses the general definition from Eq.~(\ref{eq:mka_definition})). Now, the two-sided result is obtained using the triangle inequality:
    \begin{align}
        &|\mka(K^{(k+1)},L^{(k+1)})-\mka(K^{(k)},L^{(k)})|\nonumber\\
        &\le|\mka(K^{(k+1)},L^{(k+1)})-\mka(K^{(k)},L^{(k+1)})|\nonumber\\ 
        &\quad+|\mka(K^{(k)},L^{(k+1)})-\mka(K^{(k)},L^{(k)})|. \label{eq:cor4_9_ineq}
    \end{align}
    Using the symmetry of MKA, we can write
    \begin{align}
    |\mka(K^{(k)},L^{(k+1)})-\mka(K^{(k)},L^{(k)})|=
    |\mka(L^{(k+1)},K^{(k)})-\mka(L^{(k)},K^{(k)})|,
    \end{align}
    making the terms on the right side of the inequality (\ref{eq:cor4_9_ineq}) independently bounded by $O\left(\frac{\log_2k}{k}\right)$, which gives the desired result.
\end{proof}

\section{Discussion on the Assumption: $D^2\leq(1-\eta)n$}\label{sec:eta}

\begin{figure}[t]
    \centering
    \includegraphics[width=0.4\linewidth]{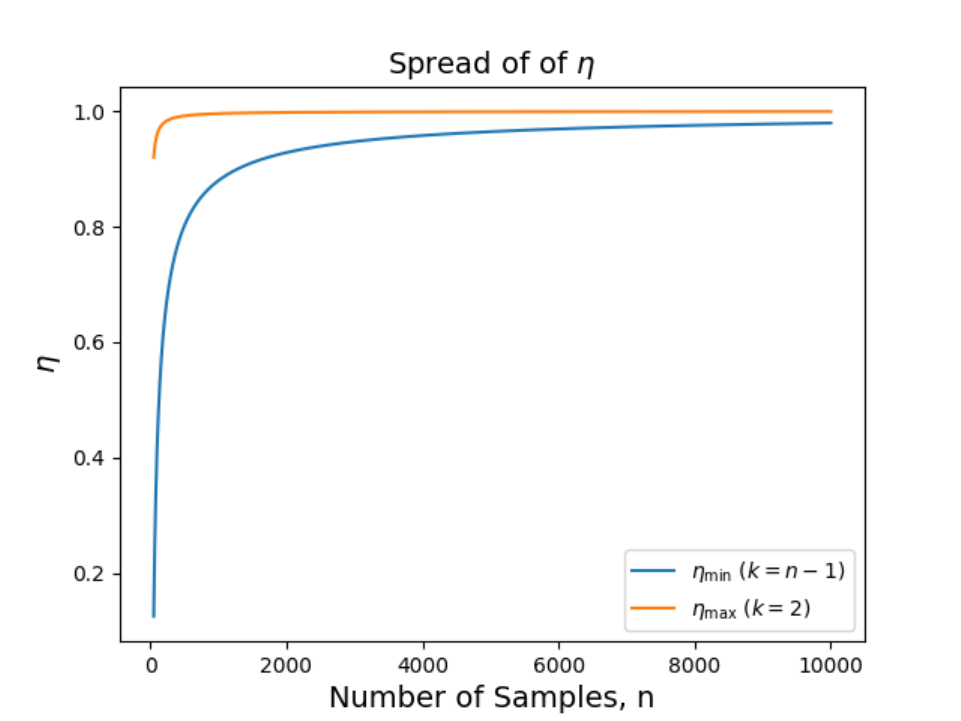}
    \caption{Validity of the assumption: $D^2=(1+\log_2k)^2=(1-\eta)n$. Any value below $\eta_{\min}$ will satisfy the assumption. For a reasonable number of samples, $N$, $\eta_{\min}$ is high and approaches $1$ as $N$ increases.}
    \label{fig:eta}
\end{figure}

To prove Corollary~\ref{cor:metric_stability} and~\ref{cor:mka_implement_Stability}, we assumed $D^2\leq(1-\eta)n$. Particularly, for the equality $D=1+\log_2k$, we can compute $\eta_{\min}$ and $\eta_{\max}$. 
We achieve this by setting $k=2$ and $k=n-1$ in the equality $D^2=(1+\log_2k)^2=(1-\eta)n$ for $\eta_{\max}$ and $\eta_{\min}$, respectively (Fig.~~\ref{fig:eta}). The value of $\eta_{\min}$ becomes stable as the number of samples increases. Any value below $\eta_{\min}$ will satisfy the condition of the statements for the respective number of samples.

\clearpage
\section{Algorithmic Details}\label{sec:algorithm}

In this section, we provide the algorithmic details of MKA and kCKA. Algorithm~\ref{alg:mka_bandwidth} provides the details of computing bandwidth parameter of MKA using binary search. The MKA algorithm is described in Algorithm~\ref{alg:mka}. The kCKA algorithm and its bandwidth computation are available in Algorithm~\ref{alg:kcka}.

\begin{algorithm}[h]
\caption{Computation of the row-wise bandwidth $\sigma_i$ using binary search}
\label{alg:mka_bandwidth}
\begin{algorithmic}[1]
\Require Sorted distances to the $k(\geq3)$ nearest neighbors (excluding itself)
\begin{align*}
    d_{i(1)} \leq d_{i(2)} \leq \cdots \leq d_{i(k)},
\end{align*}
tolerance $\varepsilon$, maximum iterations $T$, and bandwidth range $[\sigma_{\min},\sigma_{\max}]$
\Ensure Row-wise bandwidth $\sigma_i$

\State Set the local-connectivity parameter: $\rho_i \gets d_{i(1)}$

\State Compute the shifted distances: $ \Delta_{ir} \gets d_{i(r)}-\rho_i, \qquad r=1,\ldots,k $

\State Define: $F_i(\sigma) \gets \sum_{r=1}^{k} \exp\left(-\frac{\Delta_{ir}}{\sigma}\right) $

\For {$t = 1, \dots, T$}:
\State Set $\sigma=\frac{\sigma_{\min}+\sigma_{\max}}{2}$

\If{$| F_i(\sigma) - \log_2 k| < \varepsilon$}
    \State \Return $\sigma$
\EndIf

\If{$F_i(\sigma) < \log_2 k$}
    \State $\sigma_{\min} \gets \sigma$
    \Comment{Target lies below the supplied interval}
\Else
    \State $\sigma_{max} \gets \sigma $
    \Comment{Target lies above the supplied interval}
\EndIf

\EndFor
\State \Return $\sigma$ \Comment{Maximum number of iterations reached}
\end{algorithmic}
\end{algorithm}

\begin{algorithm}[t]
\caption{Manifold-approximated Kernel Alignment (MKA)}
\label{alg:mka}
\begin{algorithmic}[1]
\Require Representations
$\mathbf{X}\in\mathbb{R}^{n\times d_1}$ and
$\mathbf{Y}\in\mathbb{R}^{n\times d_2}$,
nearest neighbor parameter $k$,
distance function $d(\cdot,\cdot)$.
\Ensure Alignment score
$\mka(\mathbf{X},\mathbf{Y})$

\Function{ConstructMKA-Kernel}{$\mathbf{Z},k$}

    \State Let
    $\mathbf{Z}=[\mathbf{z}_1,\ldots,\mathbf{z}_n]^\top$

    \State Initialize $\mathbf{G}\gets \mathbf{0}_{n\times n}$

    \For{ each row $i=1,\ldots,n$}

        \State Compute the non-self distances $d_{ij}\gets d(\mathbf{z}_i,\mathbf{z}_j), \qquad j\neq i$

        \State Sort the pairs $(d_{ij},j)$ first by increasing distance and then by increasing sample index

        \State Let $j_1,\ldots,j_k$ be the indices of the first
        $k$ non-self neighbors

        \State Set $d_{i(r)} \gets d(\mathbf{z}_i,\mathbf{z}_{j_r}), \qquad r=1,\ldots,k$

        \State Compute $\sigma_i$ from $d_{i(1)},\ldots,d_{i(k)}$ using Algorithm~\ref{alg:mka_bandwidth}

        \State Set the local-connectivity parameter $\rho_i\gets d_{i(1)}$

        \State Set the diagonal entry
        $G_{ii}\gets 1$

        \For{$r=1,\ldots,k$}
            \State Set $G_{i j_r} \gets \exp\left( -\frac{d_{i(r)}-\rho_i}{\sigma_i} \right)$
        \EndFor

    \EndFor

    \State \Return $\mathbf{G}$

\EndFunction

\State Construct the directed manifold kernels
$$\mathbf{K} \gets \Call{ConstructMKA-Kernel}{\mathbf{X},k}$$
$$\mathbf{L}\gets\Call{ConstructMKA-Kernel}{\mathbf{Y},k}$$
\State Set $D \gets 1+\log_2k$

\State Compute alignment score $s$ using Eq.~\ref{eq:mka_theorem_eq}.

\State \Return $s$

\end{algorithmic}
\end{algorithm}

\begin{algorithm}[t]
\caption{$k$-nearest-neighbor Centered Kernel Alignment ($k$CKA)}
\label{alg:kcka}
\begin{algorithmic}[1]

\Require Representations
$\mathbf{X}\in\mathbb{R}^{n\times d_1}$ and
$\mathbf{Y}\in\mathbb{R}^{n\times d_2}$,
number of non-self neighbors $k$, and distance function
$d(\cdot,\cdot)$

\Ensure Similarity score $\operatorname{kCKA}(\mathbf{X},\mathbf{Y})$

\Function{KNNKernel}{$\mathbf{Z},k$}

    \State Let
    $\mathbf{Z}=[\mathbf{z}_1,\ldots,\mathbf{z}_n]^{\top}$

    \State Initialize
    $\mathbf{G}\gets\mathbf{0}_{n\times n}$

    \State Initialize $\mathcal{D}\gets\varnothing$
    
    \For{$i=1,\ldots,n$}
        \State Compute
        $d_{ij}\gets d(\mathbf{z}_i,\mathbf{z}_j)$
        for every $j\neq i$
        \State Sort the pairs $(d_{ij},j)$ first by increasing
        distance and then by increasing sample index
        \State Let $j_{i,1},\ldots,j_{i,k}$ denote the indices
        of the first $k$ non-self neighbors
        \For{$r=1,\ldots,k$}
            \State Set
            $d_{i(r)}\gets
            d(\mathbf{z}_i,\mathbf{z}_{j_{i,r}})$
            \State Add $d_{i(r)}$ to $\mathcal{D}$
        \EndFor
    \EndFor
    \State Set the global bandwidth
    $\sigma_{\mathbf{Z}}\gets\operatorname{median}(\mathcal{D})$

    \For{$i=1,\ldots,n$}
        \State Set $G_{ii}\gets 1$
        \For{$r=1,\ldots,k$}
            \State Set
            $G_{i,j_{i,r}}\gets\exp\left(-\dfrac{d_{i(r)}^2}{2\sigma_{\mathbf{Z}}^2}\right)$
        \EndFor
    \EndFor
    \State \Return $\mathbf{G}$
\EndFunction

\State Construct kCKA kernels:
$$\mathbf{K}\gets \Call{KNNKernel}{\mathbf{X},k}$$
$$\mathbf{L}\gets \Call{KNNKernel}{\mathbf{Y},k}$$
\State Compute similarity score: $s \gets \frac{\hsic{(K,L)}}{\sqrt{\hsic{(K,K)}\hsic{(L,L)}}}$
\State \Return $s$
\end{algorithmic}
\end{algorithm}

\clearpage
\section{Effect of Row-only Centering}\label{sec:row_centering_appendix}

\begin{figure}[h]
    \centering
    \includegraphics[width=\linewidth]{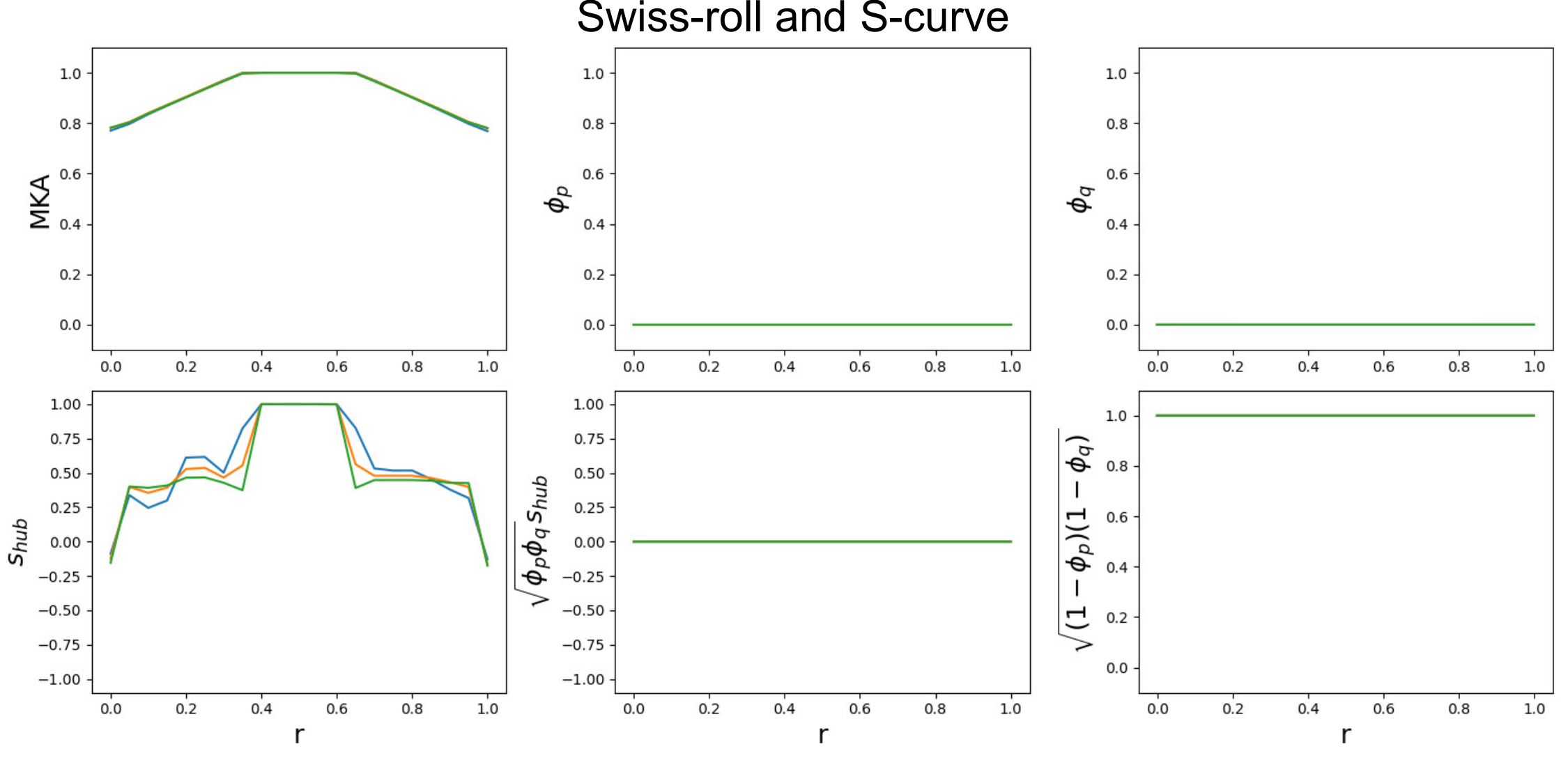}
    \caption{Effect of row-only centering in the alignment of Swiss-roll and S-curve data. (Top) MKA values, $\phi_p$, $\phi_q$, (Bottom) $s_{\text{hub}}$, $\sqrt{\phi_p,\phi_q} s_{\text{hub}}$ and $\sqrt{(1-\phi_p)(1-\phi_q)}$ for several values of $k$.}
    \label{fig:hubness_swiss}
\end{figure}

\begin{figure}
    \centering
    \includegraphics[width=\linewidth]{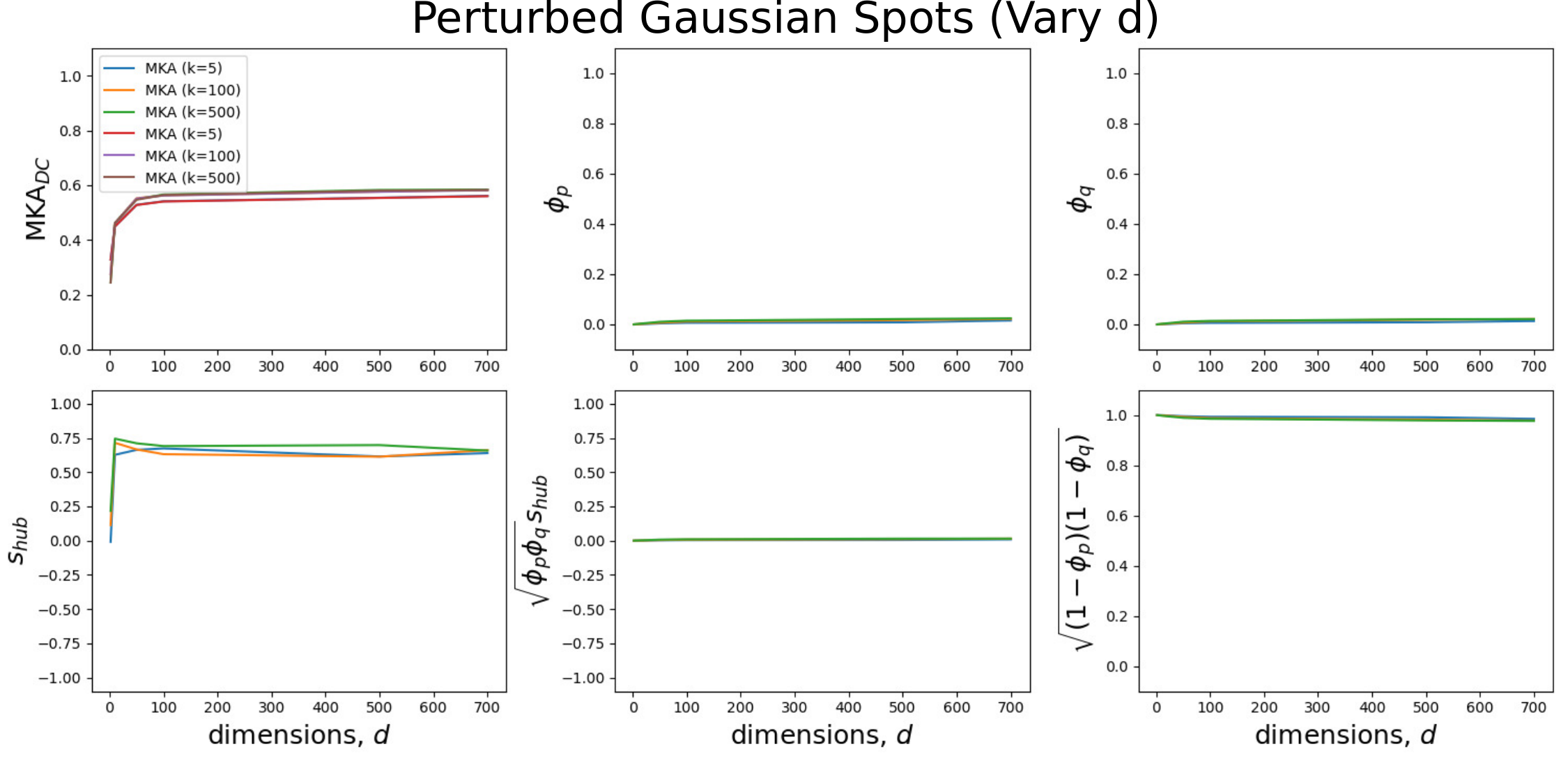}
    \caption{Effect of row-only centering in Perturbed Gaussian Spot dataset as number of dimensions varies for a fixed number of samples ($N=5,000$). (Top) MKA values, $\phi_p$, $\phi_q$, (Bottom) $s_{\text{hub}}$, $\sqrt{\phi_p,\phi_q} s_{\text{hub}}$ and $\sqrt{(1-\phi_p)(1-\phi_q)}$  for several values of $k$.}
    \label{fig:hubness_gaussian_perturb}
\end{figure}

\begin{figure}
    \centering
    \includegraphics[width=\linewidth]{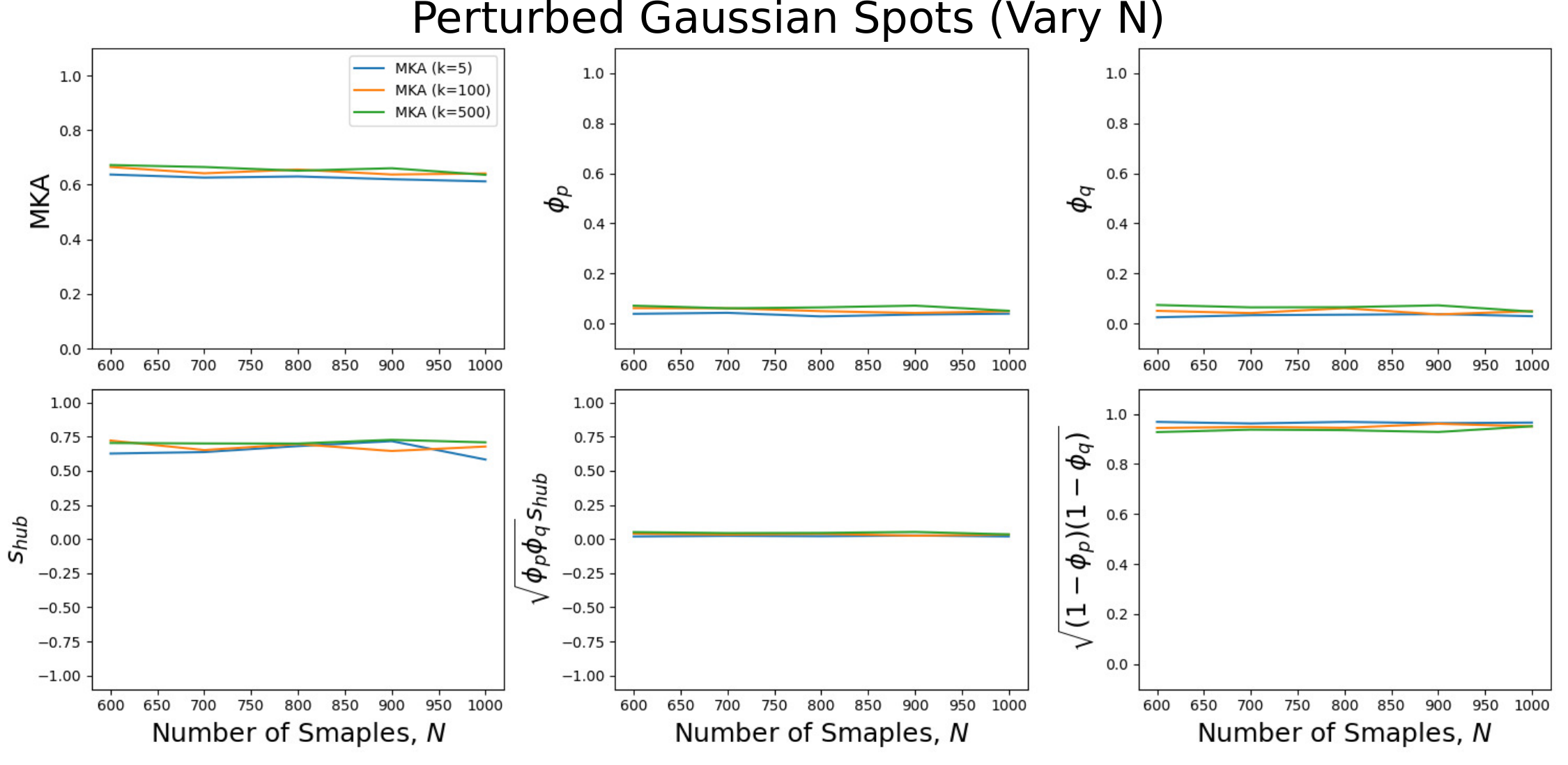}
    \caption{Effect of row-only centering in Perturbed Gaussian Spot datasets as number of samples varies for a fixed dimensionality ($d=1,000$). (Top) MKA values, $\phi_p$, $\phi_q$, (Bottom) $s_{\text{hub}}$, $\sqrt{\phi_p,\phi_q} s_{\text{hub}}$ and $\sqrt{(1-\phi_p)(1-\phi_q)}$  for several values of $k$.}
    \label{fig:hubness_gaussian_perturb_vary_N}
\end{figure}

In Section~\ref{sec:comparison_to_cka}, we have provided a simple Markov kernel interpretation of MKA, and in Proposition~\ref{prop:hubness_effect}, we have separated the effect of row-only centering. In this section, we evaluate the quantities $\phi_p$, $\phi_q$ and $s_{\text{hub}}$ for several experiments. We focus on Swiss-roll/S-curve and the perturbed Gaussian spots datasets. Overall, we see that for the Swiss-roll and S-curve (Fig.~\ref{fig:hubness_swiss}), the effect of one-sided centering is minimal. On the other hand, the Gaussian spots show effects from two different angles: varying dimensions (d) and number of samples (N). When the number of samples ($N=5000$) is reasonably high compared to the value of $k$, the row-only centering has minimal effect (Fig.~\ref{fig:hubness_gaussian_perturb}). But when the value of $k$ is close to the number of samples ($k=500$ for $N=600$, green curve in Fig.~\ref{fig:hubness_gaussian_perturb_vary_N}), the quantity $\sqrt{(1-\phi_p)(1-\phi_q)}$ is around $0.95$, while the quantity $\sqrt{\phi_p \phi_q}s_{\text{hub}}$ is still close to zero. Thus, MKA captures $95\%$ of the value of the double-centered counterpart. However, as the number of samples increases, $\sqrt{(1-\phi_p)(1-\phi_q)}$ approaches $1$.

\clearpage

\section{CKA with Manifold Approximation}\label{sec:more_cka}

\begin{figure*}[h]
    \centering
    \includegraphics[width=\linewidth]{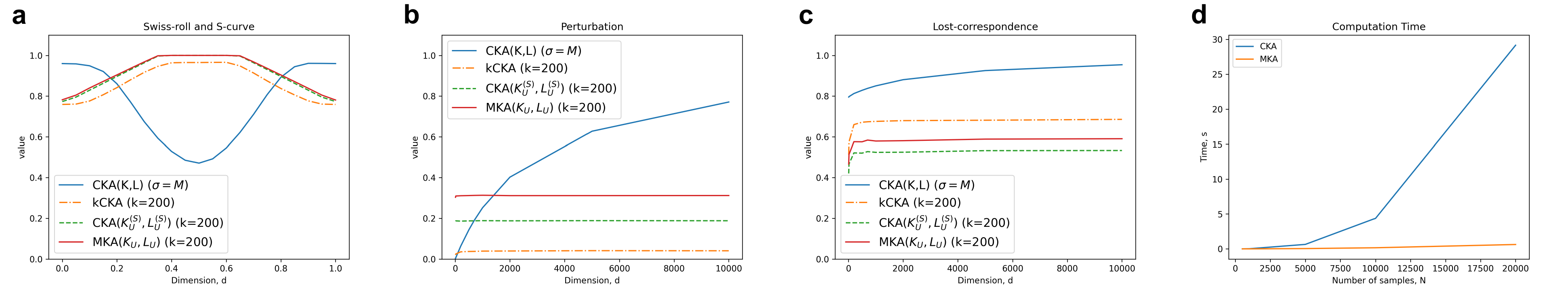}
    \caption{Effect of Kernel Approximation on the $\cka$ algorithm. (a) Alignment between Swiss-roll and S-curve. (b,c) Gaussian spots under (b) perturbation and (c) lost-correspondence. $\cka$ with manifold approximation ($\cka(K_U^{(S)},K_L^{(S)})$ behaves similarly to $\mka$, but with less bias. (d) Computation time for $\cka$ and $\mka$. $\mka$ requires much less time than $\cka$ (average of 5 runs). Note that we have excluded the computation time for the kernel matrix.}
    \label{fig:cka_time}
\end{figure*}

We can symmetrize the manifold-approximated kernel matrix, $K_U$ (we define it as $K_U$ here to differentiate from the RBF kernel matrix, $K$), using the probabilistic t-conorm given by 
\begin{align}
    K_U^{(S)}=K_U+K_U^T - K_U \circ K_U^T,
\end{align}
where $\circ$ denotes element-wise multiplication. This operation does not guarantee a positive semidefinite kernel. However, we can now directly apply CKA on the approximated kernels $K_U^{(S)}$ and $L_U^{(S)}$. The $\cka$ results obtained from this kernel matrix behave similarly to those of $\mka$ but with less bias (Fig.~\ref{fig:cka_time}(b-c)). However, computing $\mka$ requires much less time compared to $\cka$ (Fig.~\ref{fig:cka_time}(d), using \texttt{NumPy}~\citep{harris2020array}).

\section{Linear vs Non-linear CKA, kCKA}\label{sec:linear_kernel}

Linear and non-linear CKA in their default forms provide similar values, as has been known empirically~\citep{kornblith2019similarity,davarireliability} and more recently, theoretically (Theorem~\ref{theorem:alvarez}; \cite{alvarez2022gaussian}). Following this, we can claim the following for linear and non-linear kCKA:

\begin{corollary}[Linear vs. Non-linear kCKA]\label{theorem:kcka}
    $\kcka(K_{\rbf},L) = \kcka(K_{\lin},L)+O(1/\sigma^2)$ as $\sigma\to\infty$. Here, $K_{\rbf}$ is the RBF kernel matrix with bandwidth $\sigma$, $K_{\lin}$ is the linear kernel matrix, and $L$ is any positive definite symmetric kernel matrix.
\end{corollary}

In our implementation of kCKA, we constrained $\sigma$ to be the median of the distances within the k-NN set, which is often small compared to its CKA counterpart. As a result, while linear CKA and non-linear CKA can be equivalent by default implementation, it is hardly the case for kCKA. To make them equivalent, one has to arbitrarily set a large $\sigma$, which we consider an uncommon scenario. However, kCKA is sensitive to the value of $k$, and as $k$ is increased it approaches CKA.

\clearpage

\section{Computational Cost}\label{sec:complexity}

\begin{figure}[h]
    \centering
    \includegraphics[width=\linewidth]{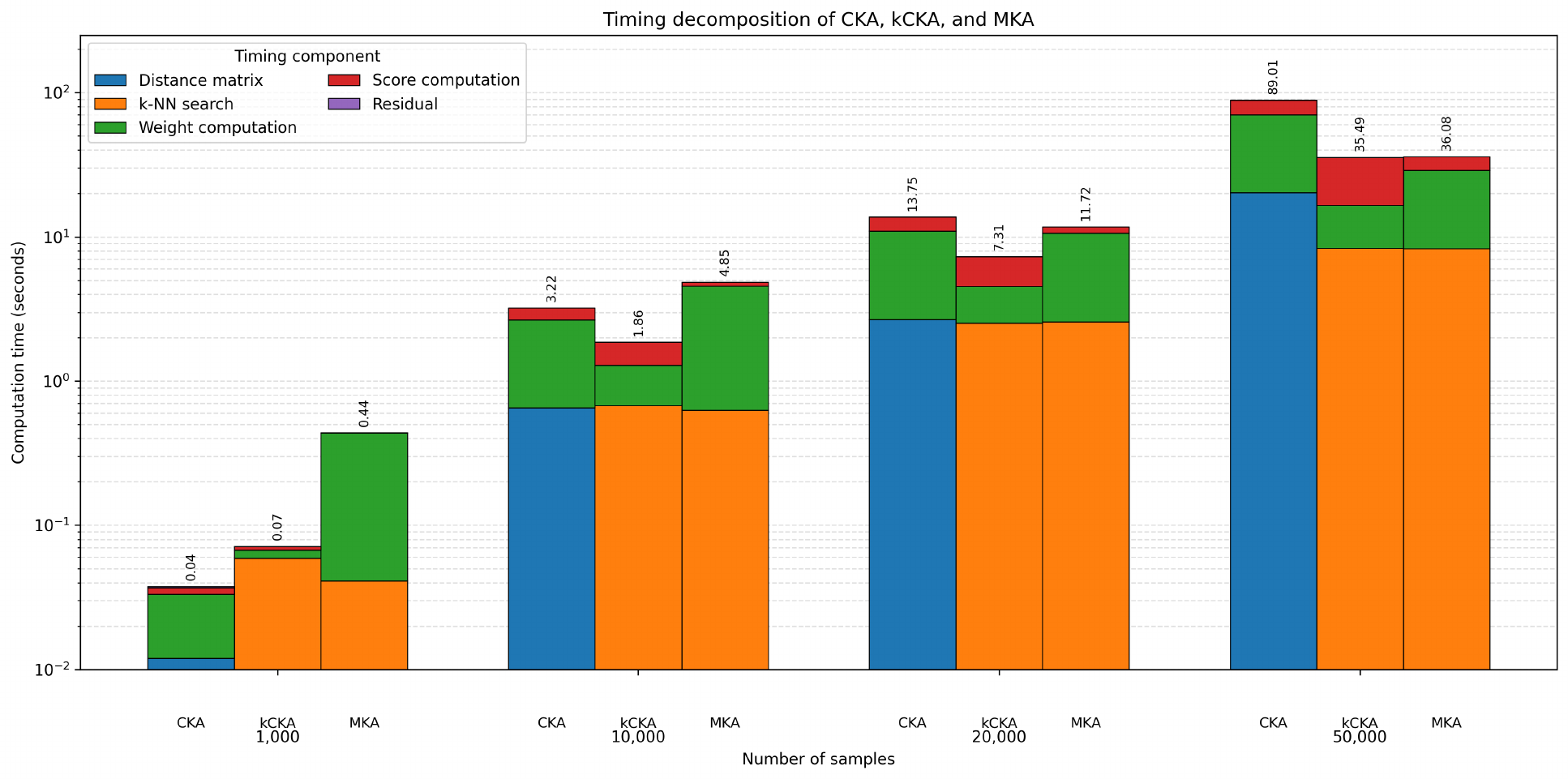}
    \caption{Timing decomposition for CKA, kCKA (k=100), and MKA (k=100) for individual components of the algorithm. `Residual' denotes the difference between total time and sum of  the components present in the graph.}
    \label{fig:timing}
\end{figure}

In this section we briefly discuss the computational complexity of the discussed algorithms. For CKA, kCKA, and MKA, we provide timing decomposition of kernel computation and score computation.

Let's assume the two representations have n samples each with $d_1$ and $d_2$ dimensions, respectively. Most algorithms discussed in the paper rely on nearest neighbor search and matrix multiplications. Particularly, constructing the $k$-nearest neighbor graphs ($O(n^2(d+\log k))$) is the costliest operation within many of them. Additionally, MKA relies on bisection method to compute the $\sigma_i$ values (Eq.~\ref{eq:UMAP_HIGH_DIM}) with a complexity $O(nk\log(\Delta/\epsilon))$, where $\Delta$ is the search range and $\epsilon$ is the tolerance. For MKA, $\log(\Delta/\epsilon)=\log(1000/10^{-12})\simeq50$ is a constant, which we ignore.  Overall, the complexity of the MKA is $O(n^2(1+d_1+d_2+\log k+nk))$. The complexities of the other algorithms are: kCKA - $O(n^3+n^2(d_1+d_2+\log k))$, CKA - $O(n^3+n^2(d_1+d_2))$. Thus, all these methods have cubic complexity in n. The complexity of SVCCA is $O(nd_1\min(n,d_1)+nd_2\min(n,d_2))$ (dominated by the singular value decomposition). RTD complexity depends on two factors: computing the distance matrix, which is the same as others, and computing the topological barcode, which is cubic in the number of simplexes~\citep{barannikov2021representation}. IMD is dominated by constructing the k-NN graph and performing m-steps of stochastic Lanczos quadrature algorithm with $n_v$ starting vectors ($O(n_v(m\log m+knm)$), giving an overall complexity of $O(n^2(d_1+d_2+\log k)+n_v(m\log m+knm))$~\citep{tsitsulin2019shape}. On the other hand, the space complexity is roughly the same for all the algorithms, primarily to store the kernel matrices, and thus it is dominated by the $O(n^2)$ term (or $O(nk)$ if only k-NN graphs are stored). 

We further investigated the computational cost of the individual steps in CKA, kCKA, and MKA (Fig.~\ref{fig:timing} and Table~\ref{tab:timing}). For this analysis, we implemented optimized versions of the three methods using 
\texttt{NumPy} and \texttt{scikit-learn} routines, including \texttt{euclidean\_distances}, \texttt{einsum}, and \texttt{NearestNeighbors}, to the best of our abilities. To compute the HSIC term efficiently, rather than explicitly evaluating \texttt{trace(K@H@L@H)}, we perform: \\ 
{ \texttt{KH = K - np.mean(K, axis=1, keepdims=True), \\LH = L - np.mean(L, axis=1, keepdims=True),\\ numerator=float(np.einsum("ij,ji->", KH, LH, optimize=True))}}.\\ 
For the nearest-neighbor computations, we used the \texttt{brute} algorithm with \texttt{n\_neighbors=k=100} and \texttt{n\_jobs=-1} on a 24-core CPU (\texttt{threadpool\_limit} was set to 48 for all routines). The MKA routine for computing the row-wise bandwidths $\sigma_i$ is our own implementation and uses standard Python code without parallelization. The numerator computation routine of the MKA is as follows:\\
\texttt{D2 = (1.0 + np.log2(k))**2 \\ numerator = np.sum(K * L) - D2}. \\
The terms in the denominator follow a similar structure.
We have used dense matrices for the score computation in all the algorithms. For each case, we sampled the representations from a 1000-dimensional Gaussian distribution ($\mathcal{N}(0,\mathbf{I}_{1000})$). We ran three trials and reported the mean.

Figure~\ref{fig:timing} presents the timing breakdown as stacked bar plots. Overall, kCKA is the fastest method, particularly at smaller sample sizes. As the number of samples increases, however, the runtimes of kCKA and MKA become comparable. Because the figure uses a logarithmic scale, the dominant component is not always immediately apparent from the bar heights alone; Table~\ref{tab:timing} therefore reports the corresponding timings in seconds. For CKA, the total runtime is dominated by kernel construction, while score computation remains relatively inexpensive. For kCKA, kernel construction dominates at smaller sample sizes, whereas score computation becomes the main bottleneck as the sample size increases. For MKA, weights (and the computation of the row-wise bandwidths $\sigma_i$) account for most of the runtime, while score computation is cheaper due to the computational advantage from Eq.~\ref{eq:mka_theorem_eq}. Across the tested sample sizes, MKA’s score computation is approximately $1.9-2.7$ times faster.

\begin{table}[t]
    \caption{Timing decomposition (in seconds) for CKA, kCKA ($k=100$), and MKA ($k=100$). `Residual' denotes the difference between total time and sum of the individual components reported in the table.}
    \label{tab:timing}
    \centering
    \begin{tabular}{c|c|c|c|c|c|c}
    \toprule
    & Method & Distance Matrix & kNN & Kernel/Weight & Score Computation & Residual \\
    \hline
    \multirow{3}{*}{\rotatebox[origin=c]{90}{\small{$1,000$}}} &
        CKA & 0.0120 & -- & 0.0214 & 0.0036 & 0.0007 \\
& kCKA & -- & 0.0592 & 0.0081 & 0.0037 & 0.0007 \\
& MKA & -- & 0.0412 & 0.3950 & 0.0016 & 0.0008 \\
       \hline
    \multirow{3}{*}{\rotatebox[origin=c]{90}{\small{$10,000$}}} &
        CKA & 0.6522 & -- & 1.9994 & 0.5585 & 0.0116 \\
& kCKA & -- & 0.6778 & 0.6142 & 0.5657 & 0.0069 \\
& MKA & -- & 0.6275 & 3.9232 & 0.2890 & 0.0070 \\
       \hline
    \multirow{3}{*}{\rotatebox[origin=c]{90}{\small{$20,000$}}} &
        CKA & 2.6684 & -- & 8.3188 & 2.7383 & 0.0294 \\
& kCKA & -- & 2.5241 & 2.0101 & 2.7617 & 0.0157 \\
& MKA & -- & 2.5672 & 7.9989 & 1.1421 & 0.0136 \\
       \hline
    \multirow{3}{*}{\rotatebox[origin=c]{90}{\small{$50,000$}}} &
        CKA & 20.3605 & -- & 49.9062 & 18.6183 & 0.1292 \\
& kCKA & -- & 8.3277 & 8.2126 & 18.9045 & 0.0411 \\
& MKA & -- & 8.2720 & 20.7067 & 7.0574 & 0.0391 \\
       \bottomrule
    \end{tabular}
\end{table}

\clearpage

\section{Clusters Data}\label{sec:clusters_data}
\begin{figure}[ht]
    \centering
    \includegraphics[width=1.0\linewidth]{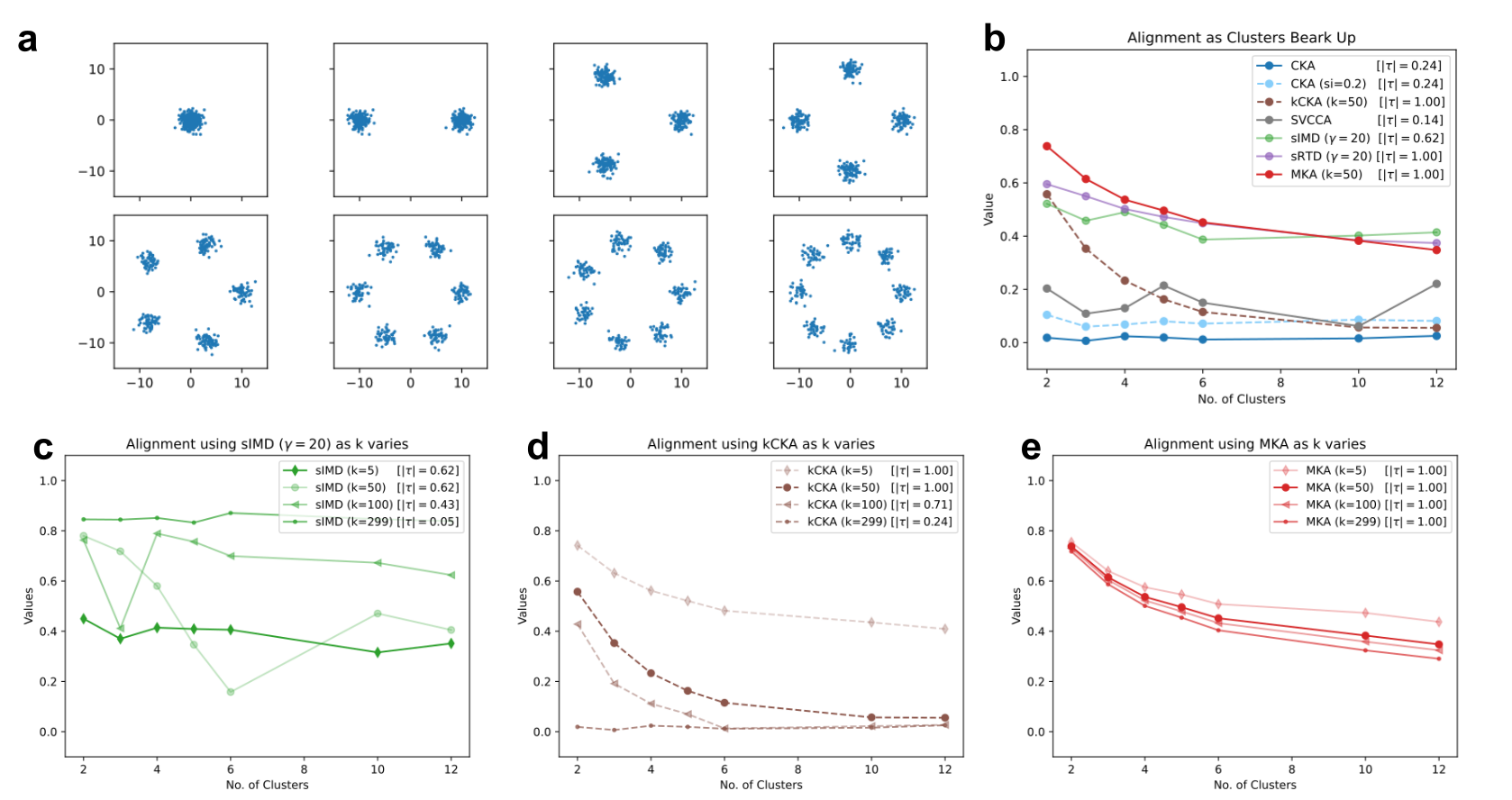}
    \caption{Alignment for the ``clusters'' data. (a) Point clouds used in the clusters experiment. (b) Alignment using various methods, along with Kendall's rank correlation (higher is better). (c-e) Alignment by varying nearest neighbors, $k$, in (c) IMD, (d) kCKA, and (e) MKA. MKA shows the greatest robustness to parameters.}
    \label{fig:clusters_break}
\end{figure}

Similar to the ``rings'' data, the ``clusters'' data was also compiled by~\cite{barannikov2021representation}. The ``clusters'' set consists of 300 points sampled from a 2D normal distribution (mean=$(0,0)$). Then the points are split into $2$, $3$, $\dots$, $12$ by moving them into a circle of radius $10$ (Fig.~\ref{fig:clusters_break}). 
Then, we use alignment metrics to compare these formations with the original structure (i.e., one cluster). The target of the experiment is to check whether the metrics can track how the data breaks into multiple clusters. Kendall's rank correlation, $\tau$, can measure this in a statistical sense (for our case, the absolute value is sufficient and thus higher is better).

CKA, SVCCA, and IMD fail to track this clustering behavior, while kCKA, RTD, and MKA capture the ranking quite well (Fig.~\ref{fig:clusters_break}(b)).  However, varying the nearest neighbor parameter, $k$, causes different behaviors in different methods. IMD shows inconsistent behavior (Fig.~\ref{fig:clusters_break}(c)). kCKA provides correct ranking only for lower values of $k$; at higher values $k \approx 100$ and above the method fails (Fig.~\ref{fig:clusters_break}(d)). MKA provides correct ranking for all possible values of $k$ (Fig.~\ref{fig:clusters_break}(e)).

\clearpage

\section{Exploring Alternate Row Sum Constraints}\label{sec:alternate_row_sum}

\begin{figure}[h]
    \centering
    \includegraphics[width=\linewidth]{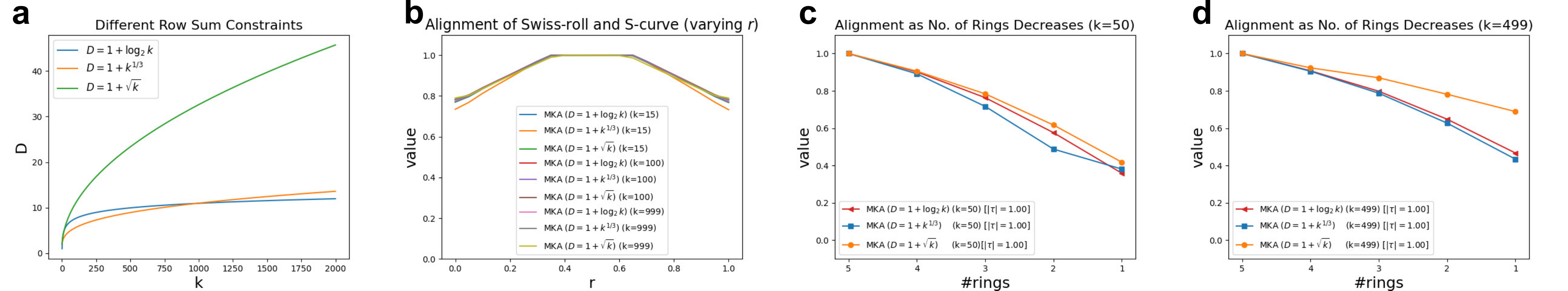}
    \caption{Exploring alternate row sum constraints in MKA. (a) Characteristics of the constraints. (b) Alignment of Swiss-roll and S-curve as $r$ varies. Alignment as the number of rings decreases for (c) $k=50$ and (d) $k=499$.}
    \label{fig:mka_different_D}
\end{figure}

In the main text, we used $D=1+\log_2(k)$ as the row sum constraint. Here, we explore two additional row sum constraints: 1. $D=1+k^{1/3}$ and 2. $D=1+\sqrt(k)$. The first one follows closely that of $D=1+\log_2(k)$, while the second one ($D=1+\sqrt(k)$) diverges earlier (Fig.~\ref{fig:mka_different_D}~(a)). However, the metric outputs and trends remain consistent for several experiments (Fig.~\ref{fig:mka_different_D}~(b-d)). Notably, $1+\sqrt{k}$, which grows faster than $1+\log_2k$, shows less sensitivity in the metric as $k$ increases (Fig. ~\ref{fig:mka_different_D}~(d), using k=499). Thus, the constant-row-sum property can control the sensitivity.

\section{Discussion on CKNNA and kCKA}\label{sec:cknna_kcka}
\begin{figure}[h]
    \centering
    \includegraphics[width=\linewidth]{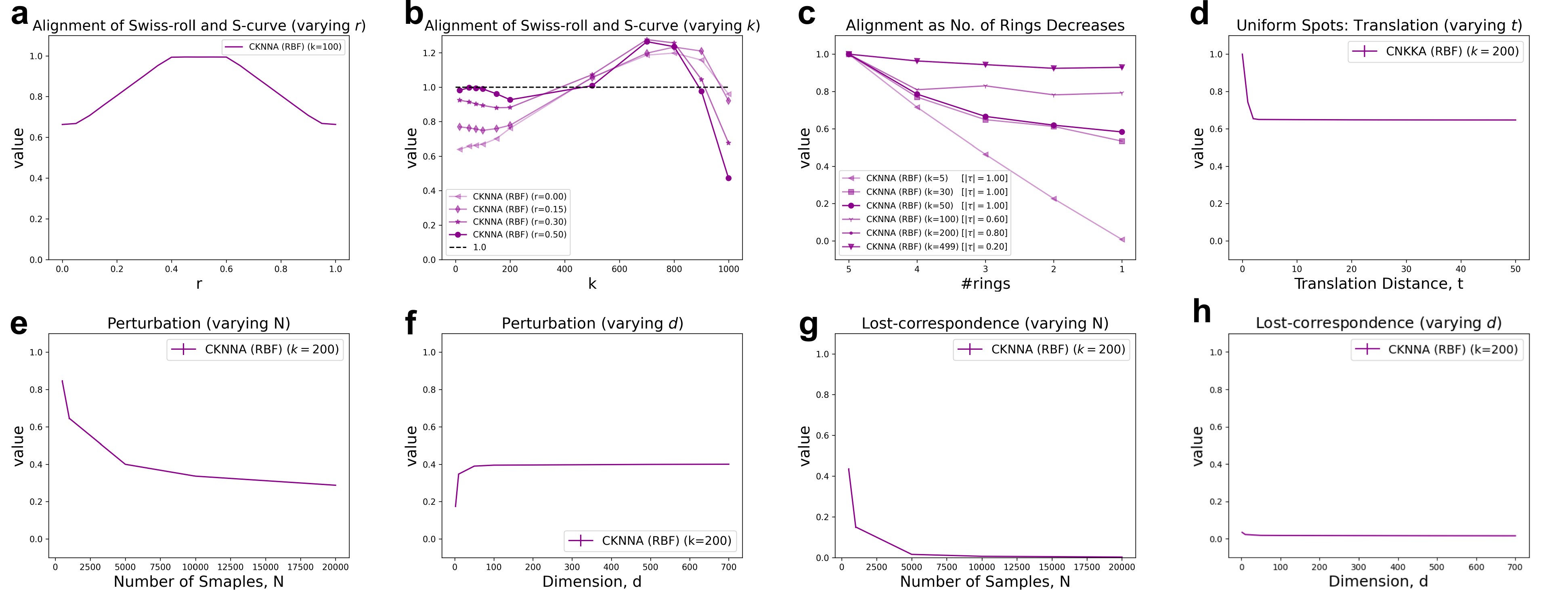}
    \caption{CKNNA scores on different experiments. Alignment between Swiss-roll and S-curve as (a) $r$ and (b) $k$ vary. (c) Results on the Rings dataset for various values of $k$. (d) Behavior on the Uniform Spots dataset as translation distance is increased. Alignment between a Gaussian spot and when it is perturbed when (e) the number of samples and (f) the number of dimensions vary. Alignment under lost correspondence when (g) the number of samples and (h) the number of dimensions vary.}
    \label{fig:cknna_results}
\end{figure}
In this section, we briefly consider Centered Kernel Nearest-Neighbor Alignment (CKNNA), introduced by \cite{huh2024platonic} as a nearest-neighbor-restricted variant of CKA. There are some discrepancies between the mathematical description in the paper and how the metric was implemented in the code. We describe the metric based on the code implementation available here: \hyperlink{https://github.com/minyoungg/platonic-rep/blob/main/metrics.py}{https://github.com/minyoungg/platonic-rep/blob/main/metrics.py}. 

CKNNA defines an alignment metric between two kernels $K$ and $L$:
\begin{equation}
\operatorname{Align}_{k}(K,L) =  \tr((\alpha^{KL}\circ K)H (\alpha^{KL} \circ L)H),
\end{equation}
where $\circ$ is the elementwise multiplication operator and $\alpha^{KL}$ is the mutual nearest neighbor mask obtained from the $k$-NN structure of the underlying data ($X$ and $Y$ for $K$ and $L$, respectively):
\begin{align}
\alpha^{KL}_{ij} = 
\begin{cases}
    1 & \text{if } x_j \in \operatorname{kNN}(x_i,k), y_j \in \operatorname{kNN}(y_i,k), \text{ and } i \neq j \\
    0 & \text{otherwise}
\end{cases}
\end{align}
The normalized CKNNA score is then
\begin{align}
\operatorname{CKNNA}(K,L) &= \frac{ \operatorname{Align}_{k}(K,L) }{ \sqrt{ \operatorname{Align}_{k}(K,K) \operatorname{Align}_{k}(L,L)}} \\
&=\frac{\tr((\alpha^{KL}\circ K)H (\alpha^{KL} \circ L)H)}{\sqrt{\tr((\alpha^{KK}\circ K)H (\alpha^{KK} \circ K)H)\tr((\alpha^{LL}\circ L)H (\alpha^{LL} \circ L)H)}}.
\end{align}

However, in kCKA each kernel retains its own kNN support:
\begin{align}
    \operatorname{kCKA}(K,L) =\frac{\tr((\alpha^{K}\circ K)H (\alpha^{L} \circ L)H)}{\sqrt{\tr((\alpha^{K}\circ K)H (\alpha^{K} \circ K)H)\tr((\alpha^{L}\circ L)H (\alpha^{L} \circ L)H)}},
\end{align}
where
\begin{align}
    \alpha^{K}_{ij} = 
\begin{cases}
    1 & \text{if } x_j \in \operatorname{kNN}(x_i,k) \text{ or } i = j \\
    0 & \text{otherwise}
\end{cases}.
\end{align}

This makes CKNNA and kCKA conceptually similar. An obvious discrepancy is that CKNNA uses a mutual nearest neighbor mask in the numerator. Moreover, the diagonal of the kernels is zero-filled in CKNNA, whereas it's one-filled in kCKA (to match the kernel property of standard CKA and MKA). Making the diagonal zero causes the self-similarity baseline to be removed from the metric~\citep{murphy2024correcting}. Additionally, for the RBF kernel, the bandwidth computation differs: kCKA computes the bandwidth based on the median distances of the kNN pairs, whereas CKNNA retains the bandwidth of the original CKA kernel.  

Since CKNNA uses different masks in its cross-alignment numerator and its two self-alignment normalization terms, the standard Cauchy--Schwarz argument for a normalized cosine score does not directly apply. In our experiments, CKNNA occasionally produced values above $1$. We did not observe $k$CKA values outside $[0,1]$ in the evaluated settings. 

Overall, as long as the individual kNN supports have strong overlap, kCKA and CKNNA show similar trends. Figure~\ref{fig:cknna_results} shows CKNNA results (for the RBF kernel) for different experiments. Overall, CKNNA shows good alignment for reasonable values of $k$.

\section{Details of Swiss-roll and S-Curve}\label{sec:moresroll}
Swiss-roll and S-curve are parameterized by variable $t\in[0,1]$. S-curve contains an additional control parameter $r\in[0,1]$ that determines the shape. $r=0.5$ gives the familiar S-curve used in many studies. We only consider 2-D shapes in this study. 
\begin{align}
\textbf{Swiss-Roll:} \nonumber \\
    z   &= \frac{3\pi}{2} (1+2t) \\
    x_1 &= z \cos(z) \\
    x_2 &= z \sin(z) \\
\textbf{S-Curve:} \nonumber \\
    z   &= 3 \pi (t-r) \\
    y_1 &= \sin(z) \\
    y_2 &= \sgn(z) (\cos(z)-1)
\end{align}

\section{Additional Details of the Experiments}\label{sec:rtdimdraw}

\begin{figure}[h]
    \centering
    \includegraphics[width=1\linewidth]{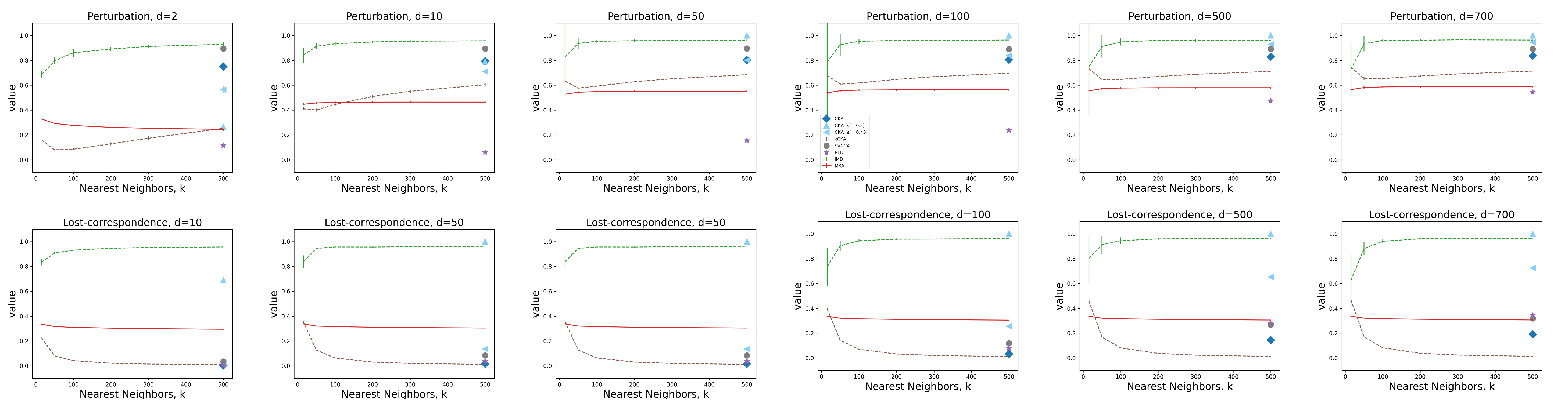}
    \caption{Dependency of the algorithms on nearest neighbor parameter $k$ for various algorithms.}
    \label{fig:all_dim_data}
\end{figure}

\begin{figure}[h]
    \centering
    \includegraphics[width=0.8\linewidth]{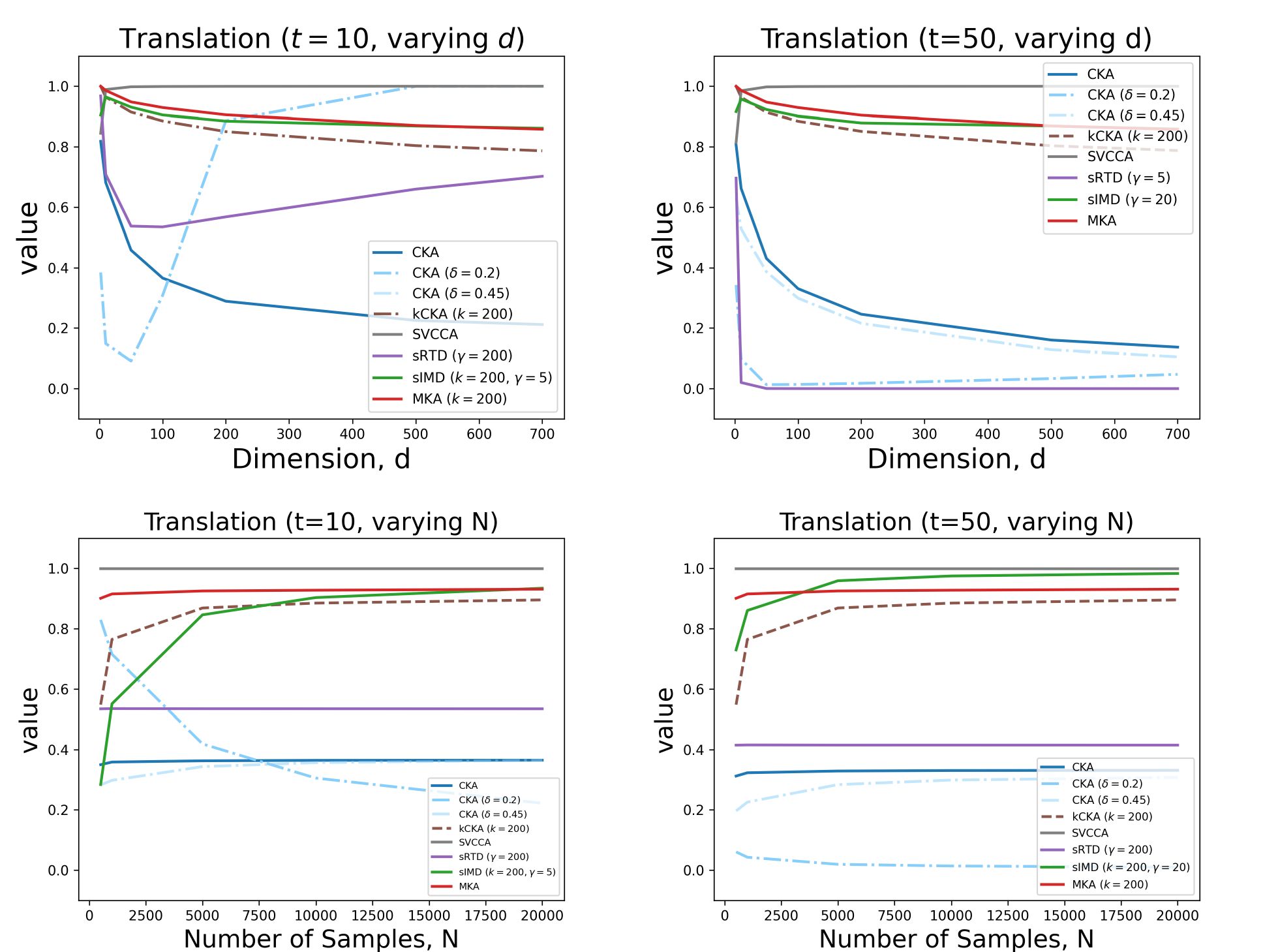}
    \caption{Additional Data for the Uniform Spots experiment. (left column) reproducing data from the main text for $t=10$. (right column) data for $t=50$.}
    \label{fig:all_t_data}
\end{figure}

\begin{figure}[t]
    \centering
    \includegraphics[width=0.8\linewidth]{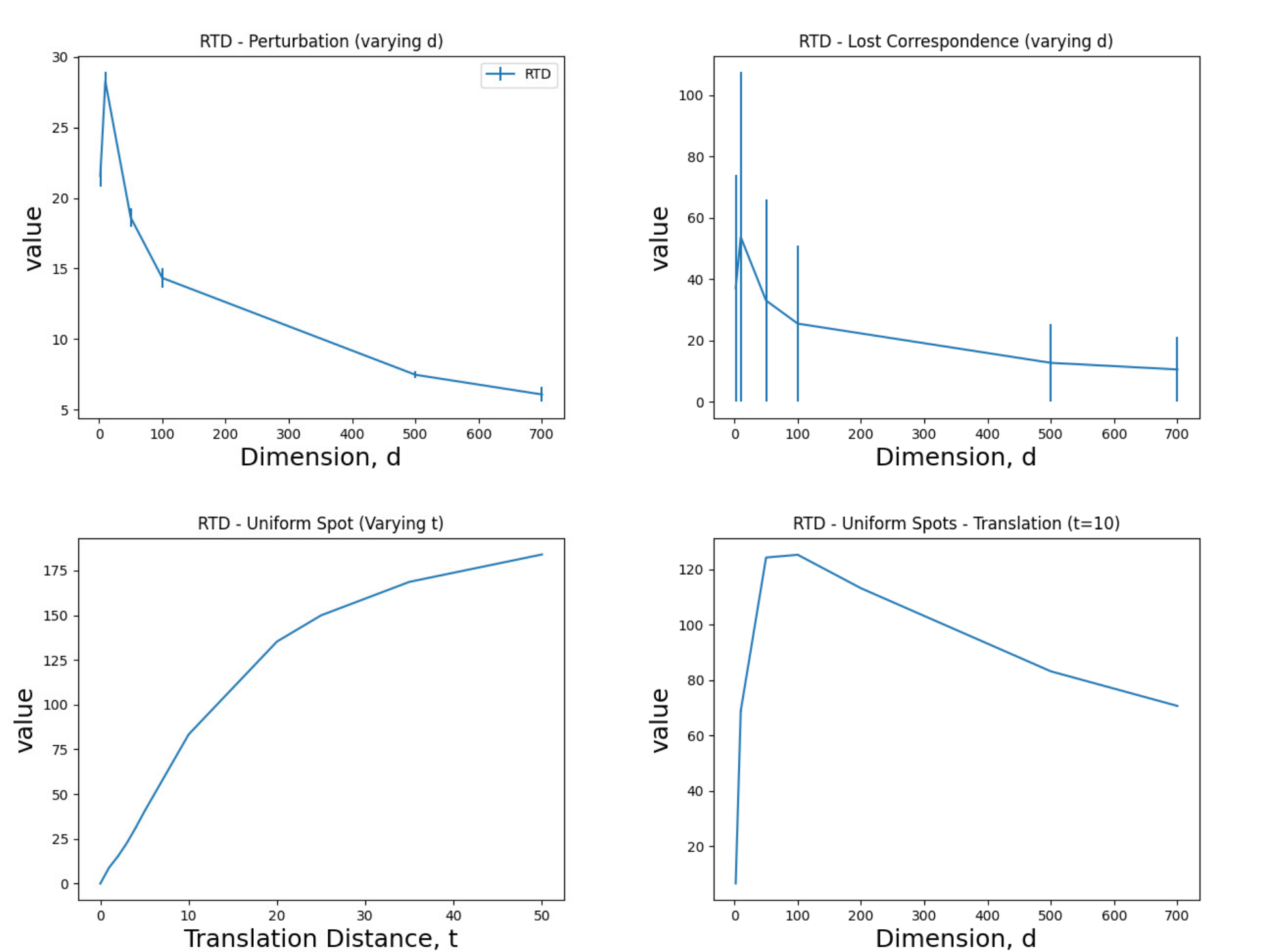}
    \caption{RTD values for a few experiments from Fig.~\ref{fig:gauss-fig}. For lost correspondence, RTD exhibits higher variance.}
    \label{fig:rtd_raw}
\end{figure}

\begin{figure}[t]
    \centering
    \includegraphics[width=0.8\linewidth]{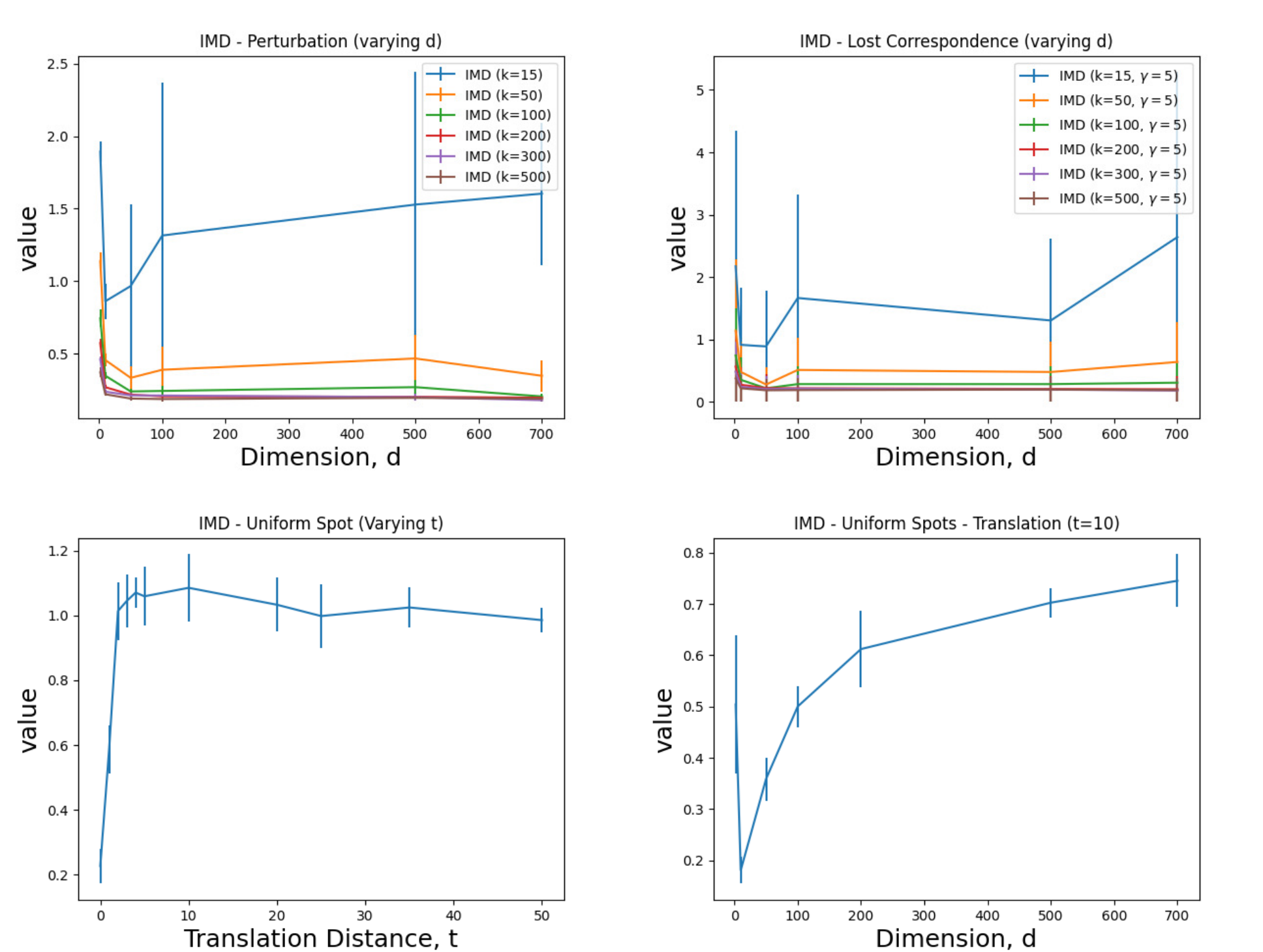}
    \caption{IMD values for a few experiments from Fig.~\ref{fig:gauss-fig}. These show much larger variance than other methods. However, as $k$ is increased, the method shows a more consistent behavior.}\label{fig:imd_raw}
\end{figure}

From Figs.~\ref{fig:all_dim_data}-\ref{fig:imd_raw} we show additional data for the experiment from Fig.~\ref{fig:gauss-fig}. Figure~\ref{fig:all_dim_data} shows the dependence on the nearest neighbor parameter $k$ to obtain a stable result. Overall, MKA is stable in the tested range of $k$, while other methods require a large value of $k$. Figure~\ref{fig:all_t_data} shows additional results for $t=50$ (in the main text, we only showed $t=10$).

In the main text, we scaled RTD and IMD values to $[0,1]$ using an exponential function so that they become easier to compare with MKA and CKA variants. Here we show (Figs.~\ref{fig:rtd_raw} and~\ref{fig:imd_raw}) the raw values of RTD and IMD for some of the experiments from Fig.~\ref{fig:gauss-fig}. In many cases, these algorithms do not exhibit consistent trends. Moreover, their raw values vary substantially across experiments.


\clearpage

\section{Implementation Details}\label{sec:implementaiton_Details}

We used our own implementation of CKA, kCKA, and MKA algorithms. For RTD, IMD, and SVCCA, we used~\cite{barannikov2021representation}'s implementation of the algorithm following examples from the corresponding GitHub repository\footnote{\url{https://github.com/IlyaTrofimov/RTD}}. Additionally, Figs.~\ref{fig:ranking_rings} and~\ref{fig:clusters_break} were also implemented by reusing code from the same repository.

The ReSi benchmark has been implemented from the publicly available repository. The full details of the benchmark are provided in Supplementary Section~\ref{sec:resi_details}.

The ResNet-50 networks have been trained using a standard training procedure (Adam~\citep{kingma2014adam} optimizer with learning rate 0.001, 50 epochs, batch size 128, with a step learning rate schedule at epochs 30 and 40 with gamma 0.1).

All experiments were conducted on a workstation equipped with two NVIDIA RTX 4090 GPUs (24 GB memory each), an AMD Ryzen Threadripper 7960X processor with 24 cores, and 256 GB of system RAM.

\section{Details of ReSi Benchmark}\label{sec:resi_details}

\subsection{Summary}
We use the representational similarity measures benchmark ReSi \citep{klabunde2024resi} to evaluate MKA and compare it with many other commonly used measures \footnote{\url{https://github.com/mklabunde/resi}}. We adopt the ReSi benchmark design, which grounds representational similarity either by prediction (tests 1–2) or by design (tests 3–6). In each test, we construct a controlled set of models and compare layer‑wise representations on held‑out data. ReSi provides the training protocols, datasets, and reference implementations for 24 baseline similarity measures; we add three new variants - MKA, CKA with RBF kernel, and CKA with RBF kernel and k-NN. The benchmark evaluates measures per test/dataset/model and reports rank‑ and decision‑based metrics accordingly.

Some measures, including PWCCA, Uniformity Difference, and Second-Order Cosine Similarity, are left blank in the result tables, and results for test 5 in the vision domain are also missing. These omissions arise from issues such as numerical instability, the occurrence of negative eigenvalues, prohibitively high runtime, or cases where the measures collapse to identical similarity values across comparisons.

\subsection{Datasets}

\subsubsection{Vision}
ImageNet-100 is a balanced subset of 100 classes sampled from the full ImageNet-1k dataset \citep{russakovsky2015imagenet}. The images are resized and center-cropped to 224×224 for both CNNs and ViTs. 

\subsubsection{Language}
MNLI \citep{williams2017broad} is a large-scale natural language inference dataset with three labels: entailment, contradiction, and neutral. It consists of premise–hypothesis pairs sampled from ten text genres. We fine-tune BERT and ALBERT on MNLI and evaluate representations exclusively on the validation-matched split. 

\subsubsection{Graphs}
For graph representation similarity tests, we use node classification datasets with fixed splits:
\paragraph{Cora}
A citation network of 2,708 machine learning publications categorized into 7 classes. Each node represents a paper, edges denote citations, and input features are 1,433-dimensional bag-of-words vectors. \citep{yang2016revisiting}
\paragraph{Flickr}
A social network dataset where the nodes represent users, edges represent follow relationships, and node features are 500-dimensional vectors derived from user metadata. The classification task has 7 labels. We subsample 10,000 test nodes for representation extraction. \citep{zeng2019graphsaint}
\paragraph{OGBN-Arxiv}
A large-scale citation network from the Open Graph Benchmark (OGB). Nodes represent 169k CS papers, edges are citation links, and each node has a 128-dimensional feature vector. The classification task involves 40 subject areas. We subsample 10,000 nodes from the test split for representation extraction. \citep{hu2020open}

\subsection{Models}
To ensure broad coverage of architectural families, we adopt representative models from vision, language, and graph domains. All models are trained or fine-tuned under standardized protocols following the ReSi benchmark, and their hidden representations are extracted in a consistent manner for similarity evaluation.

\subsubsection{Vision}
We employ three canonical CNN families and a transformer-based architecture family. These four families allow us to test whether similarity measures generalize across convolutional, residual, and attention-based architectures.
\paragraph{ResNets}
ResNet-18, ResNet-34, ResNet-101, trained from scratch on IN100 using cross-entropy loss and SGD with momentum. These models capture hierarchical convolutional features with residual connections. \citep{he2016deep}
\paragraph{VGGs}
VGG-11 and VGG-19 were trained from scratch under identical optimization schedules. Compared to ResNets, VGGs lack skip connections, providing a useful contrast in representational geometry. \citep{simonyan2014very}
\paragraph{Vision Transformers (ViTs)}
ViT-B/32 and ViT-L/32, initialized from ImageNet-21k pretraining and fine-tuned on IN100. Inputs are tokenized into 32×32 image patches with learnable positional embeddings. \citep{dosovitskiy2020image}

\subsubsection{Language}
We fine-tune two Transformer encoder models on MNLI, and both models are evaluated on the validation-matched split of MNLI.
\paragraph{BERT (base)}
Pre-trained BERT is fine-tuned with a linear learning rate schedule, 10\% warm-up, and maximum learning rate $5 \times 10^{-5}$. \citep{devlin2019bert}
\paragraph{ALBERT}
A parameter-reduced variant of BERT using factorized embeddings and cross-layer parameter sharing. Fine-tuning follows the same hyperparameter schedule as BERT. \citep{lan2019albert}

\subsubsection{Graph}
We use graph neural networks (GNNs) implemented in PyTorch Geometric, covering spectral, spatial, and attention-based designs.
\paragraph{Graph Convolutional Network (GCN)}
A spectral GNN where each layer propagates node features by normalized adjacency matrix multiplication. We train GCNs with two hidden layers for most tests, and extend to five hidden layers for the Layer Monotonicity test to ensure sufficient depth. \citep{kipf2016semi}
\paragraph{GraphSAGE}
A neighborhood-aggregation GNN that samples and aggregates neighbor features using mean aggregation. This model tests inductive generalization properties on large graphs such as Flickr and OGBN-Arxiv. \citep{hamilton2017inductive}
\paragraph{Graph Attention Network (GAT)}
A spatial GNN that computes attention coefficients over neighbors to weigh their contributions. We employ the standard configuration with 8 attention heads. \citep{velivckovic2017graph}
\paragraph{Position-aware GNN (P-GNN)}
A positional-encoding GNN that incorporates relative distance features. Due to computational constraints, P-GNN is evaluated only on Cora and excluded from augmentation tests because DropEdge perturbations are incompatible with its positional encodings. \citep{you2019position}
\subsubsection{Representation Extraction}
For all models across domains, we extract hidden representations in a standardized manner to ensure comparability of similarity measures. Unless otherwise required by a specific test (e.g., test 6: Layer Monotonicity), we always use the last hidden layer before the classifier head. For CNNs (ResNet, VGG), we take the post-global average pooling (GAP) feature vectors, and in the monotonicity test, we also extract intermediate convolutional blocks, with feature maps downsampled to a uniform 7×7 spatial resolution for memory control. For Vision Transformers, we use the [CLS] token from the final transformer block as the representation. For language models (BERT and ALBERT), we primarily use the final-layer [CLS] token embedding to represent each premise–hypothesis pair, while also including mean-pooled token embeddings as an alternative variant. For graph neural networks (GCN, GraphSAGE, GAT, P-GNN), we extract node embeddings from the last hidden layer, and in the monotonicity test, we additionally collect outputs from all intermediate layers. All representations are computed exclusively on held-out validation or test splits (IN100 validation set with 50 images per class, MNLI validation-matched set, and the test nodes of Cora/Flickr/OGBN-Arxiv) to prevent training leakage and to keep sample sizes fixed across similarity measures.

\subsection{Tests}

\subsubsection{Test 1 — Correlation to Accuracy Difference}
If two models differ in accuracy, their representations should differ accordingly. We train ten models per dataset, varying only random seeds, compute accuracies on the test split, and correlate pairwise representational similarity with the absolute accuracy difference.
\subsubsection{Test 2 — Correlation to Output Difference}
Models with similar accuracy can still produce different instance‑level predictions; we correlate representational similarity with (i) disagreement rate between hard labels and (ii) the mean Jensen–Shannon divergence (JSD) between probability vectors.
\subsubsection{Test 3 — Label Randomization}
Distinguish models trained with different degrees of label corruption. Groups are defined by randomization rate (e.g., 0\%, 25\%, 50\%, 75\%, 100\%), with five models per group. We then test if within‑group similarities exceed between‑group similarities.
\subsubsection{Test 4 — Shortcut Affinity}
Detect reliance on artificial shortcut features. We add synthetic label‑leaking features during training and form groups by shortcut “strength.” Each group consists of five independently trained models with different random seeds. A good similarity measure should assign higher similarity within groups of models trained on shortcuts of the same strength than across groups trained with different strengths.
\subsubsection{Test 5 — Augmentation}
Assess whether measures capture robustness to data augmentations. We train one “reference” group on standard data and additional groups with progressively stronger augmentation, but always evaluate on non‑augmented test data. Each group consists of five independently trained models with different random seeds. It is expected that models of the same group should have more similarity than those trained on differently augmented data.
\subsubsection{Test 6 — Layer Monotonicity}
Within a single model, nearby layers should be more similar than distant ones; we check whether similarity decreases with layer distance, and whether ordered pair constraints hold. We use the models from Tests 1–2 (for graphs, we increase the inner layers to five). We extract multiple intermediate layers and then compute (a) conformity to the ordinal constraints and (b) Spearman correlation between similarity and layer distance.

\subsection{Representational Similarity Measures}
\subsubsection{Baseline Measures (from ReSi)}
ReSi covers 24 measures spanning alignment/CCA‑type scores, RSM‑based distances, topology‑based divergences, neighborhood statistics, and simple statistics; we use their official implementations and hyperparameters.
\paragraph{CCA-based measures} ~ \\
PWCCA — Projection-Weighted Canonical Correlation Analysis \citep{morcos2018insights} \\
SVCCA — Singular Vector Canonical Correlation Analysis \citep{raghu2017svcca}
\paragraph{Alignment-based measures} ~ \\
AlignCos — Aligned Cosine Similarity \citep{hamilton2016diachronic} \\
AngShape — Orthogonal Angular Shape Metric \citep{williams2021generalized} \\
HardCorr — Hard Correlation Match \citep{li2015convergent} \\
LinReg — Linear Regression Alignment \citep{kornblith2019similarity} \\
OrthProc — Orthogonal Procrustes \citep{ding2021grounding} \\
PermProc — Permutation Procrustes \citep{williams2021generalized} \\
ProcDist — Procrustes Size-and-Shape Distance \citep{williams2021generalized} \\
SoftCorr — Soft Correlation Match \citep{li2015convergent}
\paragraph{RSM-based measures} ~ \\
CKA — Centered Kernel Alignment \citep{kornblith2019similarity} \\
DistCorr — Distance Correlation \citep{szekely2007measuring} \\
EOS — Eigenspace Overlap Score \citep{may2019downstream} \\
GULP — Generalized Unsupervised Linear Prediction \citep{boix2022gulp} \\
RSA — Representational Similarity Analysis \citep{kriegeskorte2008representational} \\
RSMDiff — RSM Norm Difference \citep{yin2018dimensionality}
\paragraph{Neighbor-based measures} ~ \\
2nd-Cos — Second-order Cosine Similarity \citep{hamilton2016cultural} \\
Jaccard — k-NN Jaccard Similarity \citep{wang2020towards} \\
RankSim — Rank Similarity \citep{wang2020towards}
\paragraph{Topology-based measures} ~ \\
IMD — Intrinsic Manifold Distance \cite{tsitsulin2019shape} \\
RTD — Representation Topology Divergence \cite{barannikov2021representation}
\paragraph{Statistic-based measures} ~ \\
ConcDiff — Concentricity Difference \citep{wang2020towards} \\
MagDiff — Magnitude Difference \citep{wang2020towards} \\
UnifDiff — Uniformity Difference \citep{wang2020understanding}

\subsubsection{Additional Measures}
In addition to the 24 baseline measures in the ReSi benchmark, we implemented three new kernel-based alignment variants (MKA, CKA with RBF kernel, and CKA with RBF kernel and k-NN).

\paragraph{Manifold Approximated Kernel Alignment (MKA)} 
In our implementation, we evaluate MKA under four neighborhood sizes, 
namely $k = 15, 50, 100, 200$. These values allow us to probe the trade-off between local geometry (small $k$) and more global manifold structure (large $k$).

\paragraph{CKA with RBF Kernel and $k$-Nearest Neighbors (kCKA)} 
In addition to the dense RBF kernel, we also evaluate a sparsified version that restricts non-zero entries to a fixed number of nearest neighbors. 
Given a representation set $X = \{x_1, \dots, x_N\}$, we compute pairwise Euclidean distances $d(x_i,x_j) = \|x_i - x_j\|_2$. For each point $x_i$, we retain only its $k$ nearest neighbors, denoted $\mathrm{KNN}(x_i, k)$. The sparsified RBF kernel matrix is then defined as
\begin{equation}
K_{ij} =
\begin{cases}
\exp \!\left( - \dfrac{d(x_i,x_j)}{2\sigma} \right), & \text{if } x_j \in \mathrm{KNN}(x_i,k), \\[8pt]
0, & \text{otherwise},
\end{cases}
\end{equation}
where the bandwidth parameter $\sigma$ is chosen as the median distance 
among all retained neighbor pairs.

The final CKA score between two representation sets $X$ and $Y$, with 
sparsified RBF kernels $K$ and $L$, is computed in the same way as standard CKA using the normalized HSIC formulation (for additional details, see Section~\ref{sec:cknna_kcka}). In our experiments, we set the neighborhood size to $k = 100$, so that each instance is only connected to its 100 nearest neighbors in the kernel matrix.

\clearpage
\section{ReSi Benchmark Scores}\label{sec:resi_scores}

\subsection{Vision Task}

\begin{table}[htbp]
\caption{Results of Test 1 (Correlation to Accuracy Difference) for the vision domain on ImageNet-100}
\label{tab:grounding-comparison}
\centering
\small
\renewcommand{\arraystretch}{1.2}
\setlength{\tabcolsep}{4pt}
\resizebox{\textwidth}{!}{
\begin{tabular}{l|ccccccc}
\toprule
\multicolumn{1}{c|}{\textbf{Test}} &
\multicolumn{7}{c}{\textbf{Accuracy Correlation}} \\
\multicolumn{1}{c|}{\textbf{Dataset}} &
\multicolumn{7}{c}{\textbf{IN100}} \\
\multicolumn{1}{c|}{\textbf{Architecture}} &
RNet18 & RNet34 & RNet101 & VGG11 & VGG19 & ViT B32 & ViT L32 \\
\midrule
CKA & \bf 0.33 & -0.09 & 0.08 & 0.02 & -0.22 & -0.23 & 0.02 \\
CKA ($\delta=0.45$) & 0.29 & -0.06 & 0.01 & 0.00 & -0.10 & -0.28 & 0.09 \\
CKA ($\delta=0.2$) & 0.02 & 0.19 & -0.08 & -0.14 & -0.23 & -0.15 & 0.02 \\
kCKA ($k=100$) & 0.00 & -0.01 & \bf 0.15 & 0.09 & \bf 0.58 & -0.16 & \bf 0.10 \\
SVCCA & 0.29 & \bf 0.27 & 0.00 & -0.04 & -0.30 & -0.01 & -0.17 \\
\bottomrule
\end{tabular}
}
\end{table}

\begin{table}[htbp]
\caption{Results of Test 2 (Correlation to Output Difference) for the vision domain on ImageNet-100}
\label{tab:grounding-comparison}
\centering
\small
\renewcommand{\arraystretch}{1.2}
\setlength{\tabcolsep}{4pt}
\resizebox{\textwidth}{!}{%
\begin{tabular}{l|ccccccc|ccccccc}
\toprule
\multicolumn{1}{c|}{\textbf{Test}} &
\multicolumn{7}{c|}{\textbf{JSD Correlation}} &
\multicolumn{7}{c}{\textbf{Disagreement Correlation}} \\
\multicolumn{1}{c|}{\textbf{Dataset}} &
\multicolumn{7}{c|}{\textbf{IN100}} &
\multicolumn{7}{c}{\textbf{IN100}} \\
\multicolumn{1}{c|}{\textbf{Architecture}} &
RNet18 & RNet34 & RNet101 & VGG11 & VGG19 & ViT B32 & ViT L32 &
RNet18 & RNet34 & RNet101 & VGG11 & VGG19 & ViT B32 & ViT L32 \\
\midrule
CKA & \bf 0.26 & 0.02 & 0.30 & -0.09 & 0.04 & 0.02 & -0.12 & \bf 0.36 & 0.00 & 0.29 & -0.01 & -0.25 & 0.00 & -0.05 \\
CKA ($\delta=0.45$) & 0.20 & -0.13 & 0.30 & -0.08 & 0.19 & 0.03 & -0.21 & 0.33 & -0.14 & 0.35 & -0.00 & -0.21 & 0.04 & -0.16 \\
CKA ($\delta=0.2$) & 0.07 & -0.03 & -0.26 & 0.02 & -0.13 & 0.32 & -0.61 & 0.35 & -0.18 & 0.08 & \bf 0.39 & \bf 0.13 & 0.35 & -0.48 \\
kCKA ($k=100$) & -0.34 & \bf 0.35 & \bf 0.43 & -0.02 &\bf 0.24 & 0.32 & -0.32 & -0.56 & \bf 0.45 & 0.18 & -0.18 & -0.17 & 0.22 & -0.17 \\
SVCCA & 0.21 & -0.00 & 0.25 & -0.11 & 0.16 & 0.05 & 0.18 & 0.39 & 0.07 & 0.15 & 0.03 & 0.01 & -0.06 & 0.07 \\
\bottomrule
\end{tabular}%
}

\end{table}

\begin{table}[htbp]
\caption{Results of Test 3 (Label Randomization) for the vision domain on ImageNet-100}
\label{tab:grounding-comparison}
\centering
\small
\renewcommand{\arraystretch}{1.2}
\setlength{\tabcolsep}{4pt}
\resizebox{\textwidth}{!}{%
\begin{tabular}{l|ccccccc|ccccccc}
\toprule
\multicolumn{1}{c|}{\textbf{Evaluation}} &
\multicolumn{7}{c|}{\textbf{AUPRC}} &
\multicolumn{7}{c}{\textbf{Conformity Rate}} \\
\multicolumn{1}{c|}{\textbf{Dataset}} &
\multicolumn{7}{c|}{\textbf{IN100}} &
\multicolumn{7}{c}{\textbf{IN100}} \\
\multicolumn{1}{c|}{\textbf{Architecture}} &
RNet18 & RNet34 & RNet101 & VGG11 & VGG19 & ViT B32 & ViT L32 &
RNet18 & RNet34 & RNet101 & VGG11 & VGG19 & ViT B32 & ViT L32\\
\midrule
CKA
        & \bf 1.00 & \bf 1.00 & \bf 1.00 & \bf 1.00 & 0.77     & 0.52     & 0.81     & \bf 1.00 & \bf 1.00 & \bf 1.00 & \bf 1.00 & 0.77     & 0.65 & 0.83 \\
CKA ($\delta=0.45$)
        & \bf 1.00 & \bf 1.00 & \bf 1.00 & \bf 1.00 & 0.77     & 0.78     & 0.86     & \bf 1.00 & \bf 1.00 & \bf 1.00 & \bf 1.00 & 0.75     & 0.81 & 0.87 \\
CKA ($\delta=0.2$)
        & 0.65     & 0.70     & 0.82     & 0.80     & 0.76     & 0.77     & \bf 0.97 & 0.71     & 0.74     & 0.87     & 0.84     & 0.73     & 0.75 & \bf 0.99 \\

kCKA ($k=100$)
        & \bf 1.00 & \bf 1.00 & \bf 1.00 & 0.88     & 0.73     & 0.57     & 0.90     & \bf 1.00 & \bf 1.00 & \bf 1.00 & 0.95     & 0.70     & 0.73 & 0.97\\
SVCCA
        & 1.00     & 0.94     & 0.95     & 0.90     & 0.78     & 0.52     & 0.58     & \bf 1.00 & 0.97     & 0.96     & 0.89     & 0.89     & 0.74 & 0.77\\
\bottomrule
\end{tabular}%
}

\end{table}

\begin{table}[htbp]
\centering
\caption{Results of Test 4 (Shortcut Affinity) for the vision domain on ImageNet-100}
\label{tab:grounding-comparison}
\small
\renewcommand{\arraystretch}{1.2}
\setlength{\tabcolsep}{4pt}
\resizebox{\textwidth}{!}{%
\begin{tabular}{l|ccccccc|ccccccc}
\toprule
\multicolumn{1}{c|}{\textbf{Evaluation}} &
\multicolumn{7}{c|}{\textbf{AUPRC}} &
\multicolumn{7}{c}{\textbf{Conformity Rate}} \\
\multicolumn{1}{c|}{\textbf{Dataset}} &
\multicolumn{7}{c|}{\textbf{IN100}} &
\multicolumn{7}{c}{\textbf{IN100}} \\
\multicolumn{1}{c|}{\textbf{Architecture}} &
RNet18 & RNet34 & RNet101 & VGG11 & VGG19 & ViT B32 & ViT L32 &
RNet18 & RNet34 & RNet101 & VGG11 & VGG19 & ViT B32 & ViT L32 \\
\midrule
CKA 
        & \bf 1.00 & \bf 1.00 & \bf 1.00 & \bf 1.00 & \bf 1.00 & \bf 1.00 & 0.86 & \bf 1.00 & \bf 1.00 & \bf 1.00 & \bf 1.00 & \bf 1.00 & \bf 1.00 & 0.96 \\
CKA ($\delta=0.45$) 
        & \bf 1.00 & \bf 1.00 & \bf 1.00 & \bf 1.00 & \bf 1.00 & \bf 1.00 & 0.90 & \bf 1.00 & \bf 1.00 & \bf 1.00 & \bf 1.00 & \bf 1.00 & \bf 1.00 & 0.97 \\
CKA  ($\delta=0.2$)
        & 0.77 & 0.62 & 0.66 & 0.80 & 0.74 & \bf 1.00 & 0.79 & 0.94 & 0.85 & 0.88 & 0.93 & 0.94 & \bf 1.00 & 0.93 \\
kCKA ($k=100$)
        & 0.94 & 0.92 & 0.98 & \bf 1.00 & \bf 1.00 & \bf 1.00 & \bf 0.91 & 0.99 & 0.98 & \bf 1.00 & \bf 1.00 & \bf 1.00 & \bf 1.00 & 0.97 \\
SVCCA 
        & 0.55 & 0.68 & 0.51 & 0.68 & 0.29 & 0.60 & 0.28 & 0.81 & 0.84 & 0.81 & 0.91 & 0.62 & 0.82 & 0.57 \\
\bottomrule
\end{tabular}%
}
\end{table}

\begin{table}[htbp]
\caption{Results of Test 6 (Layer Monotonicity) for the vision domain on ImageNet-100}
\label{tab:grounding-comparison}
\centering
\small
\renewcommand{\arraystretch}{1.2}
\setlength{\tabcolsep}{4pt}
\resizebox{\textwidth}{!}{%
\begin{tabular}{l|ccccccc|ccccccc}
\toprule
\multicolumn{1}{c|}{\textbf{Evaluation}} &
\multicolumn{7}{c|}{\textbf{Spearman}} &
\multicolumn{7}{c}{\textbf{Conformity Rate}} \\
\multicolumn{1}{c|}{\textbf{Dataset}} &
\multicolumn{7}{c|}{\textbf{IN100}} &
\multicolumn{7}{c}{\textbf{IN100}} \\
\multicolumn{1}{c|}{\textbf{Architecture}} &
RNet18 & RNet34 & RNet101 & VGG11 & VGG19 & ViT B32 & ViT L32 &
RNet18 & RNet34 & RNet101 & VGG11 & VGG19 & ViT B32 & ViT L32 \\
\midrule
CKA &  0.97 &  0.79 &  0.97  &  0.88 &  0.93 & \bf 1.00 & \bf 1.00 & 0.95 & 0.98 &  0.99  & 0.90 & 0.92 & \bf 1.00 & \bf 1.00 \\
CKA ($\delta=0.45$) &  0.90 &  0.73 &  0.91  &  0.97 &  0.94 & \bf 1.00 & \bf 1.00 & 0.94 & 0.95 &  0.97  & 0.95 & 0.93 & \bf 1.00 & \bf 1.00 \\
CKA ($\delta=0.2$) &  0.97 &  0.89 &  0.84  &  \bf 1.00 &  0.84 & 0.95 & \bf 1.00 & 0.96 & 0.96 &  0.96  & \bf 1.00 & 0.94 & 0.93 & \bf 1.00 \\
kCKA ($k=100$) &  \bf 1.00 &  \bf 1.00 & \bf  1.00 &  \bf 1.00 & \bf 1.00 & \bf 1.00 & \bf 1.00 & \bf 1.00 & \bf 1.00 & \bf 1.00 & \bf 1.00 & \bf 1.00 & \bf 1.00 & \bf 1.00 \\
SVCCA &  0.20 &  0.27 &  0.43 &  0.42 &  0.40 & 0.87 & 0.61 & 0.72 & 0.58 & 0.72 & 0.75 & 0.69 & 0.86 & 0.78 \\
\bottomrule
\end{tabular}%
}

\end{table}

\clearpage

\subsection{NLP Task}
\begin{table}[htbp]
\caption{Results of Test 1 (Correlation to Accuracy Difference) on MNLI}
\label{tab:grounding-comparison}
\centering
\small
\renewcommand{\arraystretch}{1.2}
\setlength{\tabcolsep}{4pt}
\begin{tabular}{l|cc}
\toprule
\multicolumn{1}{c|}{\textbf{Representation}} &
\multicolumn{2}{c}{\textbf{CLS Token}} \\
\multicolumn{1}{c|}{\textbf{Test}} &
\multicolumn{2}{c}{\textbf{Accuracy Correlation}} \\
\multicolumn{1}{c|}{\textbf{Architecture}} &
BERT & ALBERT \\
\midrule
CKA & \bf 0.11 & \bf 0.26 \\
CKA ($\delta=0.45$) & 0.08 & 0.09 \\
CKA ($\delta=0.2$) & -0.17 & -0.24 \\
kCKA ($k=100) $& 0.03 & -0.24 \\
SVCCA & 0.32 & -0.00 \\
RTD & 0.11 & -0.23 \\
IMD & -0.26 & -0.08 \\
MKA ($k=100$) & -0.17 & -0.26 \\
\midrule
\midrule
CKA (linear) & 0.18 & 0.17 \\
MKA ($k=15$) & -0.16 & -0.24 \\
MKA ($k=50$) & -0.16 & -0.26 \\
MKA ($k=200$) & -0.16 & -0.27 \\
\midrule
AlignedCosineSimilarity & 0.25 & 0.00 \\
ConcentricityDifference & -0.00 & -0.07 \\
DistanceCorrelation & 0.15 & 0.25 \\
EigenspaceOverlapScore & 0.03 & -0.10 \\
Gulp & 0.06 & -0.15 \\
HardCorrelationMatch & 0.04 & 0.21 \\
JaccardSimilarity & -0.21 & -0.25 \\
LinearRegression & 0.20 & 0.04 \\
MagnitudeDifference & 0.22 & -0.06 \\
OrthogonalAngularShapeMetricCentered & 0.28 & 0.12 \\
OrthogonalProcrustesCenteredAndNormalized & 0.27 & 0.12 \\
PWCCA & -0.61 & -0.27 \\
PermutationProcrustes & 0.09 & -0.02 \\
ProcrustesSizeAndShapeDistance & 0.28 & -0.04 \\
RSA & 0.00 & 0.18 \\
RSMNormDifference & 0.30 & -0.15 \\
RankSimilarity & -0.09 & -0.27 \\
SecondOrderCosineSimilarity & -0.26 & -0.25 \\
SoftCorrelationMatch & 0.11 & 0.18 \\
UniformityDifference & 0.14 & -0.16 \\
\bottomrule
\end{tabular}%
\end{table}

\begin{table}[htbp]
\caption{Results of Test 2 (Correlation to Output Difference) on MNLI}
\label{tab:grounding-comparison}
\centering
\small
\renewcommand{\arraystretch}{1.2}
\setlength{\tabcolsep}{4pt}
\resizebox{\textwidth}{!}{%
\begin{tabular}{l|cccc|cccc}
\toprule
\multicolumn{1}{c|}{\textbf{Representation}} &
\multicolumn{4}{c|}{\textbf{CLS Token}} &
\multicolumn{4}{c}{\textbf{Mean-pooled Token}} \\
\multicolumn{1}{c|}{\textbf{Test}} &
\multicolumn{2}{c}{\textbf{JSD Correlation}} &
\multicolumn{2}{c|}{\textbf{Disagreement Correlation}} &
\multicolumn{2}{c}{\textbf{JSD Correlation}} &
\multicolumn{2}{c}{\textbf{Disagreement Correlation}} \\
\multicolumn{1}{c|}{\textbf{Architecture}} &
BERT & ALBERT & BERT & ALBERT & BERT & ALBERT & BERT & ALBERT \\
\midrule
CKA
        & 0.36     & \bf 0.29 & 0.15     & \bf 0.57 & 0.51     & 0.06     & 0.38     & 0.18 \\
CKA ($\delta=0.45$)
        & 0.31     & 0.05     & \bf 0.50 & 0.22     & \bf 0.53 & 0.02     & \bf 0.49 & 0.16 \\
CKA ($\delta=0.2$)
        & -0.25    & -0.11    & 0.27     & -0.04    & 0.13     & -0.06    & -0.25    & 0.05 \\
kCKA ($k=100$)
        & 0.27     & 0.03     & 0.33     & -0.00    & 0.45     & 0.05     & 0.41     & 0.10 \\
SVCCA 
        & \bf 0.47 & 0.12     & 0.00     & 0.33     & 0.46     & \bf 0.12 & 0.13     & 0.14 \\
\bottomrule
\end{tabular}%
}
\end{table}

\begin{table}[htbp]
\caption{Results of Test 3 (Label Randomization) on MNLI}
\label{tab:grounding-comparison}
\centering
\small
\renewcommand{\arraystretch}{1.2}
\setlength{\tabcolsep}{4pt}
\resizebox{\textwidth}{!}{%
\begin{tabular}{l|cccc|cccc}
\toprule
\multicolumn{1}{c|}{\textbf{Representation}} &
\multicolumn{4}{c|}{\textbf{CLS Token}} &
\multicolumn{4}{c}{\textbf{Mean-pooled Token}} \\
\multicolumn{1}{c|}{\textbf{Evaluation}} &
\multicolumn{2}{c}{\textbf{AUPRC}} &
\multicolumn{2}{c|}{\textbf{Conformity Rate}} &
\multicolumn{2}{c}{\textbf{AUPRC}} &
\multicolumn{2}{c}{\textbf{Conformity Rate}} \\
\multicolumn{1}{c|}{\textbf{Architecture}} &
BERT & ALBERT & BERT & ALBERT & BERT & ALBERT & BERT & ALBERT \\
\midrule
CKA            & \bf 0.75 & \bf 0.80 & 0.89 & \bf 0.93 & \bf 0.66 & 0.45     & 0.86     & 0.71 \\
CKA ($\delta=0.45$)          & 0.74     & 0.79     & 0.89 & \bf 0.93 & 0.58     & 0.45     & 0.81     & 0.68 \\
CKA ($\delta=0.2$)           & 0.56     & 0.54     & 0.82     & 0.75     & 0.45     & 0.31     & 0.66     & 0.53 \\
kCKA ($k=100$)        & 0.69     & 0.46     & 0.83     & 0.68     & 0.50     & 0.43     & 0.73     & 0.59 \\
SVCCA                  & 0.69     & 0.69     & 0.84     & 0.90     & \bf 0.66 & 0.46     & \bf 0.87 & 0.76 \\
\bottomrule
\end{tabular}%
}

\end{table}

\begin{table}[htbp]
\caption{Results of Test 4 (Shortcut Affinity) on MNLI}
\label{tab:grounding-comparison}
\centering
\small
\renewcommand{\arraystretch}{1.2}
\setlength{\tabcolsep}{4pt}
\resizebox{\textwidth}{!}{%
\begin{tabular}{l|cccc|cccc}
\toprule
\multicolumn{1}{c|}{\textbf{Representation}} &
\multicolumn{4}{c|}{\textbf{CLS Token}} &
\multicolumn{4}{c}{\textbf{Mean-pooled Token}} \\
\multicolumn{1}{c|}{\textbf{Evaluation}} &
\multicolumn{2}{c}{\textbf{AUPRC}} &
\multicolumn{2}{c|}{\textbf{Conformity Rate}} &
\multicolumn{2}{c}{\textbf{AUPRC}} &
\multicolumn{2}{c}{\textbf{Conformity Rate}} \\
\multicolumn{1}{c|}{\textbf{Architecture}} &
BERT & ALBERT & BERT & ALBERT & BERT & ALBERT & BERT & ALBERT \\
\midrule
CKA            & 0.59     & 0.63     & \bf 0.88 & 0.67     & 0.55     & 0.56     & \bf 0.85 & 0.64 \\
CKA ($\delta=0.45$)          & 0.58     & \bf 0.67 & 0.87     & \bf 0.81 & 0.55     & 0.56     & 0.84     & 0.66 \\
CKA ($\delta=0.2$)           & 0.50     & 0.43     & 0.79     & 0.58     & 0.55     & 0.32     & 0.74     & 0.52 \\
kCKA ($k=100$)       & 0.57     & 0.61     & 0.83     & 0.65     & 0.56     & \bf 0.58 & 0.82     & 0.66 \\
SVCCA                  & 0.42     & 0.60     & 0.78     & 0.62     & 0.49     & 0.52     & 0.80     & 0.61 \\
\bottomrule
\end{tabular}%
}

\end{table}

\begin{table}[htbp]
\caption{Results of Test 5 (Augmentation) on MNLI}
\label{tab:grounding-comparison}
\centering
\small
\renewcommand{\arraystretch}{1.2}
\setlength{\tabcolsep}{4pt}
\resizebox{\textwidth}{!}{%
\begin{tabular}{l|cccc|cccc}
\toprule
\multicolumn{1}{c|}{\textbf{Representation}} &
\multicolumn{4}{c|}{\textbf{CLS Token}} &
\multicolumn{4}{c}{\textbf{Mean-pooled Token}} \\
\multicolumn{1}{c|}{\textbf{Evaluation}} &
\multicolumn{2}{c}{\textbf{AUPRC}} &
\multicolumn{2}{c|}{\textbf{Conformity Rate}} &
\multicolumn{2}{c}{\textbf{AUPRC}} &
\multicolumn{2}{c}{\textbf{Conformity Rate}} \\
\multicolumn{1}{c|}{\textbf{Architecture}} &
BERT & ALBERT & BERT & ALBERT & BERT & ALBERT & BERT & ALBERT \\
\midrule
CKA            & 0.44     & \bf 0.85 & \bf 0.84 & 0.84     & \bf 0.36 & \bf 0.74 & \bf 0.79 & \bf 0.78 \\
CKA ($\delta=0.45$)           & 0.45     & \bf 0.85 & 0.80     & \bf 0.86 & 0.34     & 0.72     & 0.77     & \bf 0.78 \\
CKA ($\delta=0.2$)           & 0.43     & 0.69     & 0.71     & 0.74     & 0.35     & 0.42     & 0.56     & 0.55 \\
kCKA ($k=100$)         & 0.37     & 0.76     & 0.71     & 0.78     & 0.35     & 0.63     & 0.73     & 0.73 \\
SVCCA                   & 0.43     & 0.60     & 0.77     & 0.66     & 0.31     & 0.63     & 0.70     & 0.71 \\
\bottomrule
\end{tabular}%
}

\end{table}

\begin{table}[htbp]
\caption{Results of Test 6 (Layer Monotonicity) on MNLI}
\label{tab:grounding-comparison}
\centering
\small
\renewcommand{\arraystretch}{1.2}
\setlength{\tabcolsep}{4pt}
\resizebox{\textwidth}{!}{%
\begin{tabular}{l|cccc|cccc}
\toprule
\multicolumn{1}{c|}{\textbf{Representation}} &
\multicolumn{4}{c|}{\textbf{CLS Token}} &
\multicolumn{4}{c}{\textbf{Mean-pooled Token}} \\
\multicolumn{1}{c|}{\textbf{Evaluation}} &
\multicolumn{2}{c}{\textbf{Spearman}} &
\multicolumn{2}{c|}{\textbf{Conformity Rate}} &
\multicolumn{2}{c}{\textbf{Spearman}} &
\multicolumn{2}{c}{\textbf{Conformity Rate}} \\
\multicolumn{1}{c|}{\textbf{Architecture}} &
BERT & ALBERT & BERT & ALBERT & BERT & ALBERT & BERT & ALBERT \\
\midrule
CKA           & \bf 0.99 & 0.98 & \bf 0.98 & 0.97 & \bf 1.00 & 0.99 & 0.99 & 0.99 \\
CKA ($\delta=0.45$)          & \bf 0.99 & 0.98 & \bf 0.98 & 0.97 & \bf 1.00 & \bf 1.00 & 0.99 & \bf 1.00 \\
CKA ($\delta=0.2$)           & 0.92 & 0.95 & 0.90 & 0.93 & 0.95 & \bf 1.00 & 0.96 & \bf 1.00 \\
kCKA ($k=100$)        & 0.91 & 0.90 & 0.90 & 0.89 & \bf 1.00 & \bf 1.00 & \bf 1.00 & \bf 1.00 \\
SVCCA                  & 0.91 & 0.77 & 0.91 & 0.85 & 0.78 & 0.83 & 0.85 & 0.87 \\
\bottomrule
\end{tabular}%
}

\end{table}

\clearpage
\subsection{Graph Task}

\begin{table}[htbp]
\caption{Results of Test 1 (Correlation to Accuracy Difference) for the graph domain}
\label{tab:grounding-comparison}
\centering
\small
\renewcommand{\arraystretch}{1.2}
\setlength{\tabcolsep}{4pt}
\resizebox{\textwidth}{!}{%
\begin{tabular}{l|cccccccccc}
\toprule
\multicolumn{1}{c|}{\textbf{Evaluation}} &
\multicolumn{10}{c}{\textbf{Spearman}} \\
\multicolumn{1}{c|}{\textbf{Dataset}} &
\multicolumn{4}{c}{\textbf{Cora}} &
\multicolumn{3}{|c|}{\textbf{Flickr}} &
\multicolumn{3}{c}{\textbf{OGBN-Arxiv}} \\
\multicolumn{1}{c|}{\textbf{Architecture}} &
GCN & SAGE & GAT & PGNN & GCN & SAGE & GAT & GCN & SAGE & GAT \\
\midrule
CKA   & 0.04     & -0.15    & 0.03     & -0.01 & 0.50 & \bf 0.41 & -0.14 & -0.07 & 0.14   & \bf-0.03 \\
CKA ($\delta=0.45$) & \bf 0.09 & -0.22    & 0.00     & -0.06 & \bf 0.52 & 0.17 & -0.16 & -0.12 & 0.18   & -0.13 \\
CKA ($\delta=0.2$)  & 0.01     & -0.29    & 0.07     & 0.03 & 0.43 & 0.14 & \bf 0.11 & -0.22 & 0.17     & -0.28 \\
kCKA ($k=100$)    & 0.07     & -0.25    & -0.01    & \bf 0.11 & 0.42 & -0.22 & -0.28 &\bf -0.06 & \bf 0.23   & -0.16 \\
SVCCA       & -0.03    & -0.12    & -0.16    & 0.08 & 0.01 & 0.01 & -0.18 & -0.27 & 0.09    & -0.10 \\
\bottomrule
\end{tabular}%
}
\end{table}

\begin{table}[htbp]
\caption{Results of Test 2 (Correlation to Output Difference) for the graph domain}
\label{tab:grounding-comparison}
\centering
\small
\renewcommand{\arraystretch}{1.2}
\setlength{\tabcolsep}{4pt}
\resizebox{\textwidth}{!}{%
\begin{tabular}{l|cccccccccc|cccccccccc}
\toprule
\multicolumn{1}{c|}{\textbf{Type}} &
\multicolumn{20}{c}{\textbf{Grounding by Prediction}} \\
\multicolumn{1}{c|}{\textbf{Test}} &
\multicolumn{10}{c|}{\textbf{JSD Correlation}} &
\multicolumn{10}{c}{\textbf{Disagreement Correlation}} \\
\multicolumn{1}{c|}{\textbf{Evaluation}} &
\multicolumn{10}{c|}{\textbf{Spearman}} &
\multicolumn{10}{c}{\textbf{Spearman}} \\
\multicolumn{1}{c|}{\textbf{Dataset}} &
\multicolumn{4}{c}{\textbf{Cora}} &
\multicolumn{3}{|c|}{\textbf{Flickr}} &
\multicolumn{3}{c|}{\textbf{OGBN-Arxiv}} &
\multicolumn{4}{c}{\textbf{Cora}} &
\multicolumn{3}{|c|}{\textbf{Flickr}} &
\multicolumn{3}{c}{\textbf{OGBN-Arxiv}} \\
\multicolumn{1}{c|}{\textbf{Architecture}} &
GCN & SAGE & GAT & PGNN & GCN & SAGE & GAT & GCN & SAGE & GAT & GCN & SAGE & GAT & PGNN & GCN & SAGE & GAT & GCN & SAGE & GAT \\
\midrule
CKA & 0.73 & 0.12 & 0.59 & -0.23 & 0.54 & \bf 0.35 & 0.12 & 0.18 & 0.19 & \bf 0.38 & 0.65 & -0.05 & \bf 0.44 & 0.01 & 0.26 & 0.32 & 0.02 & 0.13 & 0.14 & \bf 0.21 \\
CKA ($\delta=0.45$) & 0.73 & 0.30 & 0.52 & -0.08 & \bf 0.56 & 0.18 & 0.15 & 0.22 & 0.22 & 0.25 & 0.64 & -0.06 & 0.34 & 0.14 & \bf 0.29 & 0.15 & 0.01 & 0.21 & \bf 0.18 & 0.09 \\
CKA ($\delta=0.2$) & 0.61 & 0.49 & 0.52 & -0.04 & 0.41 & 0.25 & -0.42 & 0.16 & 0.24 & 0.11 & 0.51 & 0.17 & 0.41 & 0.03 & 0.14 & 0.22 & -0.43 & 0.15 & 0.17 & -0.01 \\
\bottomrule
\end{tabular}%
}
\end{table}

\begin{table}[htbp]
\caption{Results of Test 3 (Label Randomization) for the graph domain}
\label{tab:grounding-comparison}
\centering
\small
\renewcommand{\arraystretch}{1.2}
\setlength{\tabcolsep}{4pt}
\resizebox{\textwidth}{!}{%
\begin{tabular}{l|cccccccccc|cccccccccc}
\toprule
\multicolumn{1}{c|}{\textbf{Evaluation}} &
\multicolumn{10}{c|}{\textbf{AUPRC}} &
\multicolumn{10}{c}{\textbf{Conformity Rate}} \\
\multicolumn{1}{c|}{\textbf{Dataset}} &
\multicolumn{4}{c|}{\textbf{Cora}} &
\multicolumn{3}{c|}{\textbf{Flickr}} &
\multicolumn{3}{c|}{\textbf{OGBN-Arxiv}} &
\multicolumn{4}{c|}{\textbf{Cora}} &
\multicolumn{3}{c|}{\textbf{Flickr}} &
\multicolumn{3}{c}{\textbf{OGBN-Arxiv}} \\
\multicolumn{1}{c|}{\textbf{Architecture}} &
GCN & SAGE & GAT & PGNN & GCN & SAGE & GAT & GCN & SAGE & GAT & GCN & SAGE & GAT & PGNN & GCN & SAGE & GAT & GCN & SAGE & GAT \\
\midrule
CKA & 0.48 & 0.56 & 0.43 & 0.25 & 0.88 & 0.42 & 0.31 & \bf 1.00 & \bf 1.00 & \bf 1.00 & 0.72 & 0.78 & 0.54 & 0.64 & 0.96 & 0.50 & 0.61 & \bf 1.00 & \bf 1.00 & \bf 1.00 \\
CKA ($\delta=0.45$) & 0.50 & 0.50 & 0.52 & 0.26 & \bf 0.92 & 0.45 & \bf 0.38 & \bf 1.00 & 0.96 & \bf 1.00 & 0.75 & 0.77 & 0.76 & \bf 0.65 & \bf 0.98 & 0.66 & \bf 0.69 & \bf 1.00 & 0.98 & \bf 1.00 \\
CKA ($\delta=0.2$) & \bf 0.74 & 0.72 & 0.73 & 0.21 & 0.78 & 0.42 & 0.23 & \bf 1.00 & 0.90 & 0.99 & \bf 0.93 & 0.88 & 0.86 & 0.52 & 0.93 & 0.50 & 0.57 & \bf 1.00 & 0.98 & \bf 1.00 \\
kCKA ($k=100$) & 0.43 & 0.42 & 0.42 & 0.24 & 0.74 & 0.43 & 0.37 & \bf 1.00 & 0.74 & 0.92 & 0.59 & 0.52 & 0.52 & 0.58 & 0.88 & 0.61 & 0.66 & \bf 1.00 & 0.88 & 0.98 \\
SVCCA & 0.37 & 0.31 & 0.42 & 0.19 & 0.33 & \bf 0.80 & 0.27 & \bf 1.00 & \bf 1.00 & 0.93 & 0.63 & 0.56 & 0.59 & 0.39 & 0.69 & \bf 0.91 & 0.54 & \bf 1.00 & \bf 1.00 & 0.96 \\
\bottomrule
\end{tabular}%
}
\end{table}

\begin{table}[htbp]
\caption{Results of Test 4 (Shortcut Affinity) for the graph domain}
\label{tab:grounding-comparison}
\centering
\small
\renewcommand{\arraystretch}{1.2}
\setlength{\tabcolsep}{4pt}
\resizebox{\textwidth}{!}{%
\begin{tabular}{l|cccccccccc|cccccccccc}
\toprule
\multicolumn{1}{c|}{\textbf{Evaluation}} &
\multicolumn{10}{c|}{\textbf{AUPRC}} &
\multicolumn{10}{c}{\textbf{Conformity Rate}} \\
\multicolumn{1}{c|}{\textbf{Dataset}} &
\multicolumn{4}{c|}{\textbf{Cora}} &
\multicolumn{3}{c|}{\textbf{Flickr}} &
\multicolumn{3}{c|}{\textbf{OGBN-Arxiv}} &
\multicolumn{4}{c|}{\textbf{Cora}} &
\multicolumn{3}{c|}{\textbf{Flickr}} &
\multicolumn{3}{c}{\textbf{OGBN-Arxiv}} \\
\multicolumn{1}{c|}{\textbf{Architecture}} &
GCN & SAGE & GAT & PGNN & GCN & SAGE & GAT & GCN & SAGE & GAT & GCN & SAGE & GAT & PGNN & GCN & SAGE & GAT & GCN & SAGE & GAT \\
\midrule
CKA & 0.67 & 0.82 & 0.80 & 0.42 & 0.41 & \bf 1.00 & 0.37 & \bf 1.00 & \bf 1.00 & \bf 1.00 & 0.85 & 0.92 & 0.95 & 0.62 & 0.78 & \bf 1.00 & 0.75 & \bf 1.00 & \bf 1.00 & \bf 1.00 \\ 
CKA ($\delta=0.45$) & 0.73 & 0.81 & 0.82 & 0.42 & 0.75 & \bf 1.00 & 0.53 & \bf 1.00 & \bf 1.00 & \bf 1.00 & 0.91 & 0.92 & 0.96 & 0.55 & 0.94 & \bf 1.00 & 0.80 & \bf 1.00 & \bf 1.00 & \bf 1.00 \\ 
CKA ($\delta=0.2$) & \bf 0.79 & 0.85 & 0.78 & 0.42 & 0.97 & \bf 1.00 & 0.29 & \bf 1.00 & \bf 1.00 & \bf 1.00 & \bf 0.95 & 0.96 & 0.93 & 0.55 & 0.99 & \bf 1.00 & 0.60 & \bf 1.00 & \bf 1.00 & \bf 1.00 \\ 
kCKA ($k=100$) & 0.77 & 0.86 & \bf 0.84 & 0.42 & \bf 1.00 & \bf 1.00 & \bf 0.55 & \bf 1.00 & \bf 1.00 & \bf 1.00 & 0.94 & 0.95 & \bf 0.97 & 0.56 & \bf 1.00 & \bf 1.00 & \bf 0.84 & \bf 1.00 & \bf 1.00 & \bf 1.00 \\ 
SVCCA & 0.23 & 0.36 & 0.46 & 0.24 & 0.24 & 0.93 & 0.32 & \bf 1.00 & 0.97 & 0.83 & 0.46 & 0.60 & 0.69 & 0.53 & 0.57 & 0.97 & 0.66 & \bf 1.00 & 0.99 & 0.91 \\
\bottomrule
\end{tabular}%
}

\end{table}

\begin{table}[htbp]
\caption{Results of Test 5 (Augmentation) for the graph domain}
\label{tab:grounding-comparison}
\centering
\small
\renewcommand{\arraystretch}{1.2}
\setlength{\tabcolsep}{4pt}
\resizebox{\textwidth}{!}{%
\begin{tabular}{l|cccccccc|cccccccc}
\toprule
\multicolumn{1}{c|}{\textbf{Evaluation}} &
\multicolumn{8}{c|}{\textbf{AUPRC}} &
\multicolumn{8}{c}{\textbf{Conformity Rate}} \\
\multicolumn{1}{c|}{\textbf{Dataset}} &
\multicolumn{3}{c}{\textbf{Cora}} &
\multicolumn{3}{|c|}{\textbf{Flickr}} &
\multicolumn{2}{c|}{\textbf{OGBN-Arxiv}} &
\multicolumn{3}{c}{\textbf{Cora}} &
\multicolumn{3}{|c|}{\textbf{Flickr}} &
\multicolumn{2}{c}{\textbf{OGBN-Arxiv}} \\
\multicolumn{1}{c|}{\textbf{Architecture}} &
GCN & SAGE & GAT & GCN & SAGE & GAT & GCN & SAGE & GCN & SAGE & GAT & GCN & SAGE & GAT & GCN & SAGE \\
\midrule
CKA & 0.13 & 0.76 & 0.91 & 0.64 & 0.80 & 0.57 & \bf 1.00 & \bf 1.00 & 0.37 & 0.92 & 0.97 & 0.85 & 0.94 & 0.77 & \bf 1.00 & \bf 1.00 \\
CKA ($\delta=0.45$) & 0.13 & 0.80 & \bf 0.98 & 0.69 & 0.94 & 0.58 & \bf 1.00 & \bf 1.00 & 0.38 & 0.93 & \bf 0.99 & 0.86 & 0.99 & 0.80 & \bf 1.00 & \bf 1.00\\
CKA ($\delta=0.2$) & 0.13 & \bf 0.93 & 0.93 & \bf 0.75 & 0.42 & 0.55 & 1.00 & 0.99 & 0.38 & \bf 0.99 & 0.97 & 0.90 & 0.52 & \bf 0.84 & \bf 1.00 & \bf 1.00\\
kCKA ($k=100$) & 0.13 & 0.76 & 0.96 & 0.74 & 0.82 & 0.60 & \bf 1.00 & 0.96 & 0.37 & 0.92 & \bf 0.99 & 0.89 & 0.94 & 0.83 & \bf 1.00 & 0.99\\
SVCCA & \bf 0.14 & 0.29 & 0.34 & 0.64 & 0.67 & 0.49 & 0.89 & 0.81 & 0.39 & 0.55 & 0.58 & 0.84 & 0.86 & 0.68 & 0.89 & 0.85\\
\bottomrule
\end{tabular}%
}
\end{table}

\begin{table}[htbp]
\caption{Results of Test 6 (Layer Monotonicity) for the graph domain}
\label{tab:grounding-comparison}
\centering
\small
\renewcommand{\arraystretch}{1.2}
\setlength{\tabcolsep}{4pt}
\resizebox{\textwidth}{!}{%
\begin{tabular}{l|cccccccccc|cccccccccc}
\toprule
\multicolumn{1}{c|}{\textbf{Evaluation}} &
\multicolumn{10}{c|}{\textbf{Conformity Rate}} &
\multicolumn{10}{c}{\textbf{Spearman}} \\
\multicolumn{1}{c|}{\textbf{Dataset}} &
\multicolumn{4}{c|}{\textbf{Cora}} &
\multicolumn{3}{c|}{\textbf{Flickr}} &
\multicolumn{3}{c|}{\textbf{OGBN-Arxiv}} &
\multicolumn{4}{c|}{\textbf{Cora}} &
\multicolumn{3}{c|}{\textbf{Flickr}} &
\multicolumn{3}{c}{\textbf{OGBN-Arxiv}} \\
\multicolumn{1}{c|}{\textbf{Architecture}} &
GCN & SAGE & GAT & PGNN & GCN & SAGE & GAT & GCN & SAGE & GAT & GCN & SAGE & GAT & PGNN & GCN & SAGE & GAT & GCN & SAGE & GAT \\
\midrule
CKA & 0.99 & \bf 1.00 & \bf 1.00 & 0.99 & 0.81 & \bf 0.99 & 0.63 & 0.88 & 0.70 & 0.90 & 0.99 & \bf 1.00 & \bf 1.00 & 0.99 & 0.73 & \bf 0.99 & 0.48 & 0.86 & 0.58 & 0.91 \\
CKA ($\delta=0.45$) & \bf 1.00 & \bf 1.00 & \bf 1.00 & \bf 1.00 & 0.86 & \bf 0.99 & 0.80 & 0.84 & 0.80 & 0.93 & \bf 1.00 & \bf 1.00 & \bf 1.00 & \bf 1.00 & 0.79 & \bf 0.99 & 0.66 & 0.75 & 0.73 & 0.93 \\
CKA ($\delta=0.2$) & \bf 1.00 & \bf 1.00 & \bf 1.00 & \bf 1.00 & 0.97 & \bf 0.99 & 0.85 & 0.91 & 0.96 & \bf 1.00 & \bf 1.00 & \bf 1.00 & \bf 1.00 & \bf 1.00 & 0.98 & \bf 0.99 & 0.64 & 0.84 & 0.91 & \bf 1.00 \\
kCKA ($k=100$) & \bf 1.00 & \bf 1.00 & \bf 1.00 & 0.91 & 0.95 & 0.95 & 0.96 & 0.88 & 0.98 & \bf 1.00 & \bf 1.00 & \bf 1.00 & \bf 1.00 & 0.91 & 0.90 & 0.97 & 0.95 & 0.85 & 0.99 & \bf 1.00 \\
SVCCA & 0.41 & 0.78 & 0.61 & \bf 1.00 & 0.75 & 0.69 & 0.57 & 0.49 & 0.41 & 0.56 & 0.63 & 0.80 & 0.81 & \bf 1.00 & 0.53 & 0.44 & 0.47 & -0.17 & -0.10 & 0.33 \\
\bottomrule
\end{tabular}%
}

\end{table}

\end{document}